\documentclass[sigconf]{acmart}

\usepackage{multirow}
\usepackage{multicol}
\usepackage{makecell}
\usepackage{utfsym}
\usepackage{hyperref}
\usepackage{graphicx}
\usepackage{subfig}
\usepackage{array}
\usepackage{color}
\usepackage{amsmath}

\renewcommand\footnotetextcopyrightpermission[1]{}
\AtBeginDocument{%
  }

\setcopyright{acmcopyright}
\copyrightyear{2018}
\acmYear{2018}
\acmDOI{XXXXXXX.XXXXXXX}

\acmPrice{15.00}
\acmISBN{978-1-4503-XXXX-X/18/06}

\begin{document}

\title{INF³: Implicit Neural Feature Fusion Function for Multispectral and Hyperspectral Image Fusion}

\author{ShangQi Deng}
\authornote{Both authors contributed equally to this research.}
\email{shangqideng0124@gmail}
\author{RuoCheng Wu}
\authornotemark[1]
\affiliation{%
  \institution{School of Mathematical Sciences, University of Electronic Science and Technology of China}
  \city{Chengdu}
  \country{China}
  \postcode{611731}
}

\author{Liang-Jian Deng}
\authornote{Corresponding author}
\email{liangjian.deng@uestc.edu.cn}
\orcid{}
\affiliation{%
	\institution{School of Mathematical Sciences, University of Electronic Science and Technology of China}
	\city{Chengdu}
	\country{China}
	\postcode{611731}
}

\author{Ran Ran}
\email{RanRan@std.uestc.edu.cn}
\orcid{}
\affiliation{%
	\institution{School of Mathematical Sciences, University of Electronic Science and Technology of China}
	\city{Chengdu}
	\country{China}
	\postcode{611731}
}

\author{Gemine Vivone}
\email{gemine.vivone@gmail.com}
\orcid{}
\affiliation{%
	\institution{Institute of Methodologies for Environmental Analysis (CNR-IMAA), \\ National Biodiversity Future Center}
	\city{Tito}
	\country{Italy}
	\postcode{I-85050}
}

\renewcommand{\shortauthors}{Trovato et al.}

\begin{abstract}
	Multispectral and Hyperspectral Image Fusion (MHIF) is a practical task that aims to fuse a high-resolution multispectral image (HR-MSI) and a low-resolution hyperspectral image (LR-HSI) of the same scene to obtain a high-resolution hyperspectral image (HR-HSI). Benefiting from powerful inductive bias capability, CNN-based methods have achieved great success in the MHIF task. However, they lack certain interpretability and require convolution structures be stacked to enhance performance. Recently, Implicit Neural Representation (INR) has achieved good performance and interpretability in 2D tasks due to its ability to locally interpolate samples and utilize multimodal content such as pixels and coordinates. Although INR-based approaches show promise, they require extra construction of high-frequency information (\emph{e.g.,} positional encoding). In this paper, inspired by previous work of MHIF task, we realize that HR-MSI could serve as a high-frequency detail auxiliary input, leading us to propose a novel INR-based hyperspectral fusion function named Implicit Neural Feature Fusion Function (INF³). As an elaborate structure, it solves the MHIF task and addresses deficiencies in the INR-based approaches. Specifically, our INF³ designs a Dual High-Frequency Fusion (DHFF) structure that obtains high-frequency information twice from HR-MSI and LR-HSI, then subtly fuses them with coordinate information. Moreover, the proposed INF³ incorporates a parameter-free method named INR with cosine similarity (INR-CS) that uses cosine similarity to generate local weights through feature vectors. Based on INF³, we construct an Implicit Neural Fusion Network (INFN) that achieves state-of-the-art performance for MHIF tasks of two public datasets, \emph{i.e.,} CAVE and Harvard. The code will soon be made available on GitHub.
	
\end{abstract}

\keywords{Implict Neural Representation (INR), Multispectral and Hyperspectral Image Fusion (MHIF)}
\maketitle

\begin{figure}[t]
	\centering
	\includegraphics[width=8.5cm, height=6cm]{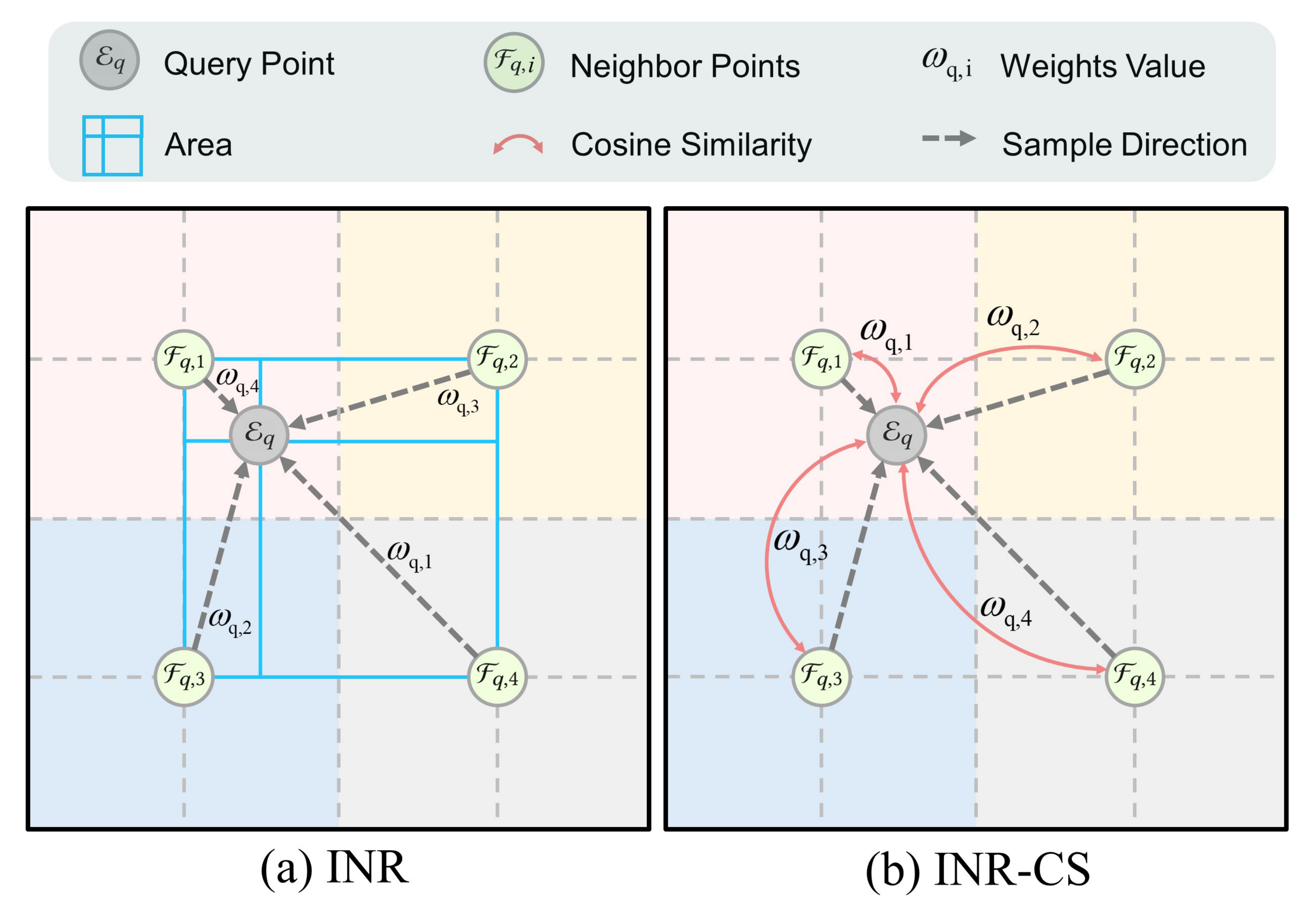}
	\caption{(a) The left figure shows the INR of generating weights based on area in LIIF~\cite{chen2021learning}. (b) The proposed INR with cosine similarity (INR-CS) method, which generates weights based on cosine similarity and takes into account the pixel values, is depicted in the right figure. Our method generates more reasonable weights by taking into account the correlation of points in the feature space, in contrast to LIIF-generated weights that consider only relative positions in the coordinate space.}
	\vspace{-5mm}
\end{figure}
\section{Introduction}
Hyperspectral imaging involves capturing a scene in various contiguous spectral bands. Compared to traditional single or few-band images (such as those with RGB channels), hyperspectral (HS) images provide finer information about real observations and thus better characterize image scenes. As a result, HSIs have found wide application in different areas of computer vision and have improved the accuracy of several tasks, such as object recognition, classification, tracking, and segmentation~\cite{fauvel2012advances,van2010tracking,tarabalka2009segmentation,uzair2013hyperspectral}. However, practical optical sensor systems face limitations in incident energy, necessitating tradeoffs between spatial resolution and spectral refinement. In particular, hyperspectral (HS) images with more than 100 bands often have a relatively low spatial resolution, while multispectral (MS) images with a limited number of bands have a relatively high spatial resolution. Therefore, exploring the fusion of a high spatial resolution multispectral image (HR-MSI) and a low spatial resolution hyperspectral image (LR-HSI) of the same scenario into a high spatial resolution hyperspectral image (HR-HSI) has attracted increasing attention. The aim is to obtain as rich and precise HR and HS data as possible.

In recent times, the CNN-based method has achieved considerable success due to its remarkable ability to extract advanced features when applied to multispectral and hyperspectral image fusion. Researchers have demonstrated that the two-stream fusion network designed for HR-MSI and MR-HSI is bounded by the two-stream fusion network for them. To maintain both spatial and spectral information, existing work attempts to design attention modules that produce high-quality spatial details. However, most existing networks are based on a generic CNN framework, which lacks interpretability for MHIF tasks.

Motivated by recent advancements in Implicit Neural Representation (INR) for 3D object/scene representation~\cite{mildenhall2021nerf,jiang2020local,park2019deepsdf} and image super-resolution~\cite{chen2021learning,zhang2022implicit,tang2021joint}, we propose to re-examine the fusion process from a different perspective. INR involves mapping continuous spatial coordinates to signals in a domain through an implicit function. In order to obtain prior information from different scenes and integrate it with the implicit function, an existing encoder is typically employed to extract the latent code from the scene/imagery. For 2D tasks, the implicit function usually takes a weighted average of a fixed number of neighboring latent codes to ensure output value continuity. However, due to the lack of sufficient prior information across neighboring coordinates, the weights of such implicit interpolation are commonly dependent on area~\cite{chen2021learning} or network parameters~\cite{tang2021joint}, which limit performance or interpretability. Thus we generate fusion weights using parameter-free cosine similarity solving of the latent code. Additionally, the MLP-ReLU structure used by INR has inherent high-frequency information bias~\cite{rahaman2019spectral} that is not easily eliminated during training. Therefore, we propose aligning HR-MSI and LR-HSI images to extract high-frequency information in a multiscale and multimodal manner. Finally, we integrate the learning framework of weight generation and image fusion into a unified implicit function, called the implicit neural feature fusion function (INF³) representation.

The contribution of this paper is listed as follows:
\begin{itemize}
	\item We propose an Implicit Neural Feature Fusion Function (INF³), which is the first attempt that applied Implicit Neural Representation (INR) on Multispectral and Hyperspectral Image Fusion (MHIF) task. In the fusion stage, we only utilize an MLP layer, which reduces the burden brought by massive use of convolution.
	
	\item To enrich the network's input, our INF³ adopts the practice of Dual High-Frequency Fusion (DHFF) structure across three modalities which combines high-frequency spatial information at different resolutions. Concretely, this method allows the MLP layer in INF³ to access more high-frequency information for detail recovery.
	
	\item The proposed INR with cosine similarity (INR-CS) method utilizes cosine similarity to generate weights that makes better use of information inside the pixel rather than distance or area. The proposed method does not depend on any extra parameters or network structures. Instead, it generates parameters based on cosine similarity between feature vectors and fuses local information.
	
	\item Based upon INF³, we construct an Implicit Neural Fusion Network (INFN) using encoder-decoder architecture. The proposed INFN has achieved state-of-the-art performance on two public datasets, \emph{i.e.,} CAVE and Harvard. Specifically, the proposed decoder has a lightweight structure yet prevents overfitting of INR structures on MHIF tasks.
	
\end{itemize}

\section{Related work}
\subsection{CNNs in MHIF}
Recently, CNN-based techniques have shown significant success in multispectral and hyperspectral image fusion (MHIF) due to their capacity to learn high-level features from input data through end-to-end training. Among these methods, SSRNet~\cite{zhang2020ssr} uses three convolution modules\textemdash  fusion, spatial edge, and spectral edge\textemdash to restructure the image, with a loss function connected to the spatial and spectral edges ensuring training reliability. Similarly, ResTFNet~\cite{LIU20201} utilizes residual structures and a two-stream fusion network to learn input data from different modalities, inspired by the widespread application of ResNet~\cite{he2016deep} in image super-resolution. MHF-net~\cite{xie2020mhf}, on the other hand, was specifically designed for the MS/HS fusion task, incorporating a well-researched linear mapping that links the HR-HSI image to the HR-MSI and LR-HSI images, as well as clear interpretability. Meanwhile, MoG-DCN~\cite{dong2021model} builds a dedicated sub-network for approximating the degradation matrix and leverages DCN-based image regularization~\cite{dong2018denoising} for HISR, fully exploiting prior HSI knowledge. For simultaneous extraction of spatial and spectral information and production of high-quality details, HSRnet~\cite{hu2021hyperspectral} employs channel and spatial attention modules. To ensure bidirectional data consistency and improve accuracy in both spatial and spectral domains, DBIN~\cite{Wang_2019_ICCV} proposes a deep learning-based approach that optimizes the observation model and fusion procedures repeatedly and alternately during reconstruction. Finally, while CNN has a strong structure, INR-based approaches have shown tremendous potential for both 3D and 2D tasks.

\subsection{Implicit Neural Representation}
Recently, implicit representations of 3D objects, scenes and shapes have gained significant momentum in research. Traditional discrete explicit representations have been partly replaced by implicit neural representations (INR), which use parameterized MLPs to map coordinate information into signals (coor-MLP) in the target domain. For example, NeRF~\cite{mildenhall2021nerf} expanded the input 3D coordinate to a continuous 5D scene representation with a 2D viewing direction, resulting in better renderings of high-frequency scene content than explicit 3D representations such as voxel methods, point cloud, and mesh. DeepSDF~\cite{park2019deepsdf} takes a 3D coordinate and a categorical latent code as input and outputs the signed distance (SDF) at this coordinate to determine whether it is inside the target shape. Related works have enhanced INR's ability to model 3D surfaces and shapes~\cite{sitzmann2019scene,michalkiewicz2019implicit,chabra2020deep,jiang2020local}. This approach has also been extended to the 2D domain, for example, Local Implicit Image Function (LIIF)~\cite{chen2021learning} extracts a set of latent codes distributed in the LR domain to interpolate the HR target image. Based on LIIF, UltraSR~\cite{xu2021ultrasr} attempts to apply residual structure to the 2D INR process and add the multiple injection of coordinate information and residual structure. Furthermore, LTE~\cite{lee2022local} proposes a local texture estimator to characterize the image information into the Fourier domain and incorporate it with the coordinate information as input to the MLP. SIREN~\cite{sitzmann2020implicit} proposes an overall implicit neural representation framework to adapt the complex natural signals and their derivatives using a periodic activation function. CRM~\cite{shen2022high} performs image segmentation refinement using implicit neural representations. When applied to processing multimodal data, JIIF~\cite{tang2021joint} proposes using INR to reconstruct depth images in the HR domain by using LR domain RGB images guided with noisy low-resolution depth images. This work strongly inspired us to use INR to process multispectral and hyperspectral Image Fusion. However, previous work has demonstrated the limitations and biases of the MLP-ReLU structure in learning high-frequency information~\cite{rahaman2019spectral}. Therefore, we focus on designing strategies for the fusion process of different modes to improve the performance of high-frequency representation and add a decoder after the MLP layer to correct the bias.

\subsection{Motivation}
LR-HSI and HR-MSI provide abundant spectral and spatial information, respectively, making them a valuable resource for image analysis. However, fully utilizing the local content of these images and fusing information from different modalities, such as spatial, spectral, and coordinate, are challenging. To address this issue, we propose an implicit neural fusion network (INFN) that relies on the implicit neural representation (INR) of the image. The INR-based approaches have demonstrated exceptional performance in arbitrary-scaled image super-resolution tasks~\cite{chen2021learning}, frequently employing a multilayer perceptron (MLP) as the fusion component. However, MLPs tend to acquire low-frequency information, necessitating additional input of high-frequency data, such as position or frequency encoding~\cite{xu2021ultrasr,song2023ope}. To overcome this limitation, we introduce the implicit neural feature fusion function (INF³). Inspired by the multiscale injection branch of SSconv~\cite{wang2021ssconv}, in INF³ we inject detailed high-frequency information in dual scales, specifically using MLPs to learn high-frequency data for MHIF task. Additionally, we address the challenge of identifying feature vectors that are close in distance but different in angle by proposing that our INF³ utilizes cosine similarity between feature vectors to compute coefficients. In detail, we utilize full-size and reduced-size HR-MSI to generate interpolated weights, eliminating the need for network learning or additional parameters. As a result, our fusion framework has demonstrated state-of-the-art performance on two publicly available datasets.

\section{Methodology}
In this section, we present our INF³ representation designed for the MHIF task. We first introduce the overall architecture of our implicit neural fusion network (INFN) in Sec.~\ref{section:3.1}. Subsequently, we review recent implicit neural representations (INR) for 2D tasks in Sec.~\ref{section:3.2}. Finally, we describe the design of INF³ in Sec.~\ref{section:3.3} for the fusion process.

\begin{figure*}[t]
	\scriptsize
	\setlength{\tabcolsep}{0.7pt}
	\centering
	\includegraphics[width=17.5cm, height=7.3cm]{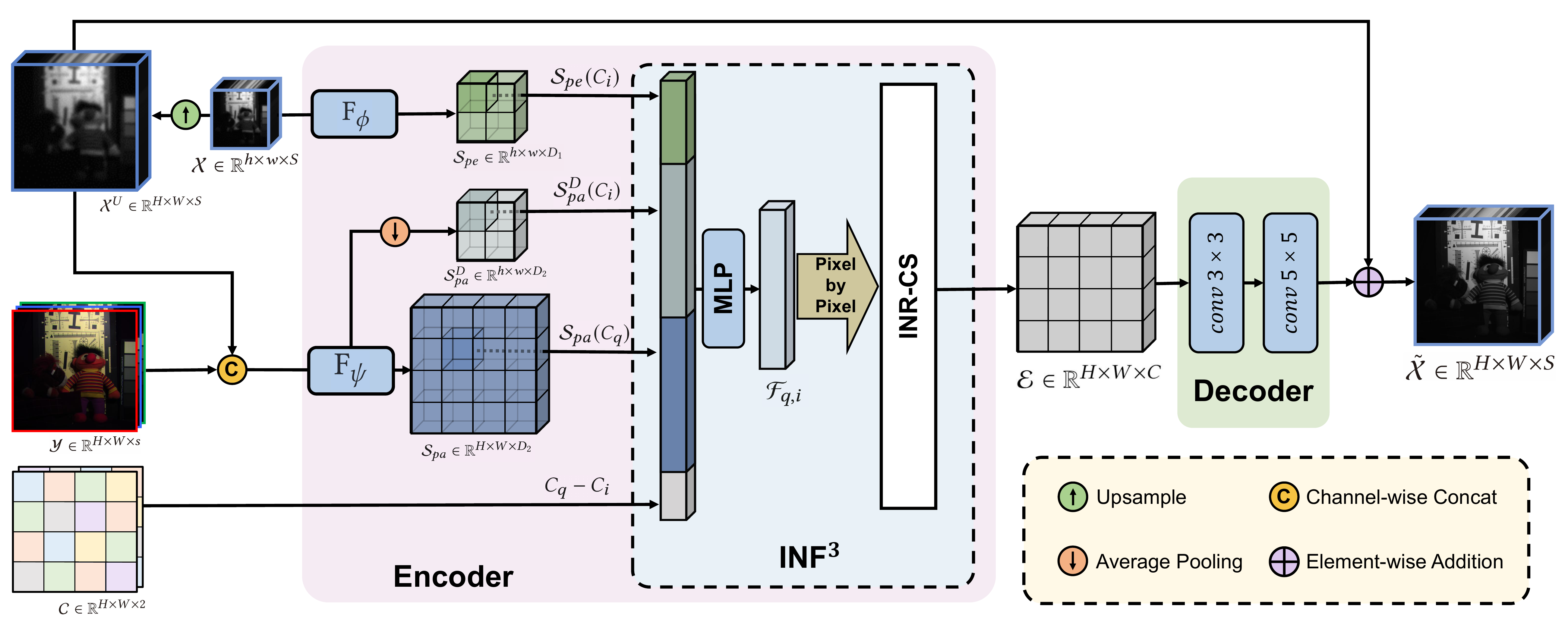}
	\caption{The overall architecture of the proposed INFN, which consisted of two segments: encoder and decoder. Specifically, we input three modal information such as LR-HSI $\mathcal{X}$, HR-HSI $\mathcal{Y}$ and coordinate $\mathcal{C}$ into the encoder, and subsequently put the encodered result into the decoder and add it with up-sampled LR-HSI $\mathcal{X}^{U}$ to get the final output $\tilde{\mathcal{X}}$. The INR-CS is described with detail in Sec.~\ref{section:3.2} and Sec.~\ref{section:3.3}.}
	\label{network}
	\vspace{-4mm}
\end{figure*}
\subsection{The Overall Architecture}
\label{section:3.1}
As shown in Fig.~\ref{network}, the INFN is generally divided into two segments: encoder and decoder. In practice, it is evident that directly applying an INR-based approach to address MHIF tasks often leads to overfitting. To overcome this challenge and ensure network stability during training, we have opted for an encoder-decoder architecture. Supplementary materials will include relevant ablation experiments. Specifically, the encoder stage can be formulated as follows:
\begin{equation}  
	\mathcal{E}={\rm Encoder}\left(\mathcal{X},\mathcal{Y},\mathcal{C}\right),\\
\end{equation}
where $\mathcal{E}\in \mathbb{R}^{H\times W\times D}$ represents the fusion result of the encoder, $\mathcal{X} \in \mathbb{R}^{h\times w\times S}$ denotes the LR-HSI, $\mathcal{Y} \in \mathbb{R}^{H\times W\times s}$ denotes the HR-MSI, and $\mathcal{C} \in \mathbb{R}^{H \times W \times 2}$ is the normalized 2D coordinate map in the high resolution (HR) domain. In detail, we propose to represent a pixel by its center position and scale the coordinate map of $H\times W$ into the square grid of size $[-1,1]\times [-1,1]$ to make it convenient to share the coordinates in both the HR and LR domains. The normalization process in HR domain can be formulated as:
\begin{equation}
	\mathcal{C}\left(i,j\right)=\left[-1+\frac{2i+1}{H},-1+\frac{2j+1}{W} \right],
\end{equation}
where $i\in \left[0,H-1\right]$, $j\in \left[0,W-1\right]$. To deal with information about different modes, \emph{i.e.,} LR-HSI and HR-MSI, we utilize function ${\rm F}_\psi$ and function ${\rm F}_\phi$ to extract the spatial and spectral information, respectively. The process of spectral function can be formulated as follows:
\begin{equation}
	\mathcal{S}_{pe}={\rm F}_{\phi} \left(\mathcal{X}\right),
	\label{spe}
\end{equation}
where $\mathcal{S}_{pe}\in \mathbb{R}^{h\times w\times D_1}$ is the feature map of spectral modality and $\phi$ is learnable parameters of spectral function. $\mathcal{D}_1$ is the number of output channels of the spectral function. To extract information from spatial modality, we propose to concatenate bicubic interpolated LR-HSI $\mathcal{X}^U \in \mathbb{R}^{H\times W\times S}$ with the HR-MSI $\mathcal{Y} \in \mathbb{R}^{H\times W\times s}$, thus inputting it into the spatial function $f_\phi$ for extracting. In special, this process can be expressed by the formula:
\begin{equation}
	\mathcal{S}_{pa}={\rm F}_\psi({\rm Cat}(\mathcal{X}^U,\mathcal{Y})),
	\label{spa}
\end{equation}
where $\mathcal{S}_{pa}\in \mathbb{R}^{H\times W\times D_2}$ is the feature map of spatial modality and $\psi$ is learnable parameters of spatial function. $\mathcal{D}_2$ is the number of output channels of the spatial function. In addition, ${\rm Cat}(\cdot)$ means the concatenation operation in channel dimension. We view the INF³ framework as the key of encoder, which can be formulated as:
\begin{equation}
	\mathcal{E} = \mathrm{INF^3}(\mathcal{S}_{pe},\mathcal{S}_{pa}, \mathcal{C}).
\end{equation}
For the decoding process, we work on the encoding output $\mathcal{E} \in \mathbb{R}^{H\times W \times C}$ to generate the decoding result $\mathcal{D} \in \mathbb{R}^{H\times W \times S}$ via a two-layer convolution structure. The parameters of the decoder are shared by all training patches. In general, the neural network tends to predict frequencies located near a low frequency region. Yet, past work has proved that a long skip connection in local implicit representation enriches high-frequency components in residuals and stabilizes convergence~\cite{kim2016accurate}. Thus, we add the bicubic interpolation LR-HSI $\mathcal{X}^U$ as a long skip connection to ameliorate the above problem. Thus, the final signal take the form: 
\begin{equation}
	\tilde{\mathcal{X}} = \mathrm{Decoder}(\mathcal{E}) + \mathcal{X}^U.
\end{equation}

\subsection{Implicit Neural Representation}
\label{section:3.2}
In this session, we will introduce the implicit neural representation (INR) from the perspective of interpolation method. Given a low-resolution image $x \in \mathbb{R}^{h\times w\times 3}$ and the corresponding high-resolution (HR) interpolated image $\hat{x} \in \mathbb{R}^{H\times W\times 3}$ as an example, the INR up-sampling process at position $C_q$ can be expressed as:
\begin{equation}
	\hat{x}(C_q)=\sum_{i\in\mathcal{N}_q}w_{q,i}v_{q,i},
\end{equation}
where $C_q\in \mathbb{R}^2$ is the normalized coordinate of the query pixel in the HR domain, $\mathcal{N}_q \in \mathbb{R}^4$ is the coordinate of neighbor pixels for $C_q$ in the LR domain, $w_{q,i} \in \mathbb{R}$ is the interpolation weight of $v_{q,i} \in \mathbb{R}^{1\times 1\times 3}$, and $v_{q,i}$ is the vector of $x$. The interpolation weights are usually normalized so that $\sum_{i\in\mathcal{N}_q}w_{q,i}=1$. Previous work usually proposes to set $\mathcal{N}_q$ to the pixels at the four nearest centers of $C_q$ in the LR domain. The calculation of the interpolation weights varies from articles to articles, and the simplest formulation of area weight interpolation used by LIIF~\cite{chen2021learning} is as follows:
\begin{equation}
	w_{q,i}=\frac{A_i}{A},
\end{equation}
where $A_i$ is the partial area diagonally opposite to the $i$ corner pixel, $A = \sum_{i\in\mathcal{N}_q}A_i$ is the total area serving as the denominator. In detail, the LIIF fuses LR pixel information with HR relative coordinate information through MLP to generate the interpolation value $v_{q,i}$, which takes the following form: 
\begin{equation}
	v_{q,i}=\mathrm{MLP}_{\Theta}(x(C_i),C_q-C_i),
\end{equation}
where $\mathrm{MLP}_{\Theta}(\cdot)$ is an MLP layer with learnable parameters $\Theta$ that takes a local feature vector $x(C_q)$ in the LR domain and a relative coordinate $C_q-C_i$ as inputs. From the above equations, the interpolated vector can be represented by a set of local feature vectors in the LR domain, which stores the low-resolution information of the local region. In general, INR-based methods implement up-sampling by querying $x(C_q)$ with the relative query coordinate $C_q-C_i$ in the arbitrary super-resolution task.

\begin{figure*}[!h]
	\scriptsize
	\setlength{\tabcolsep}{0.9pt}
	\centering
	\subfloat[GT]{
		\begin{minipage}[b]{0.19\linewidth}
			\includegraphics[width=1.35in,height=1.35in]{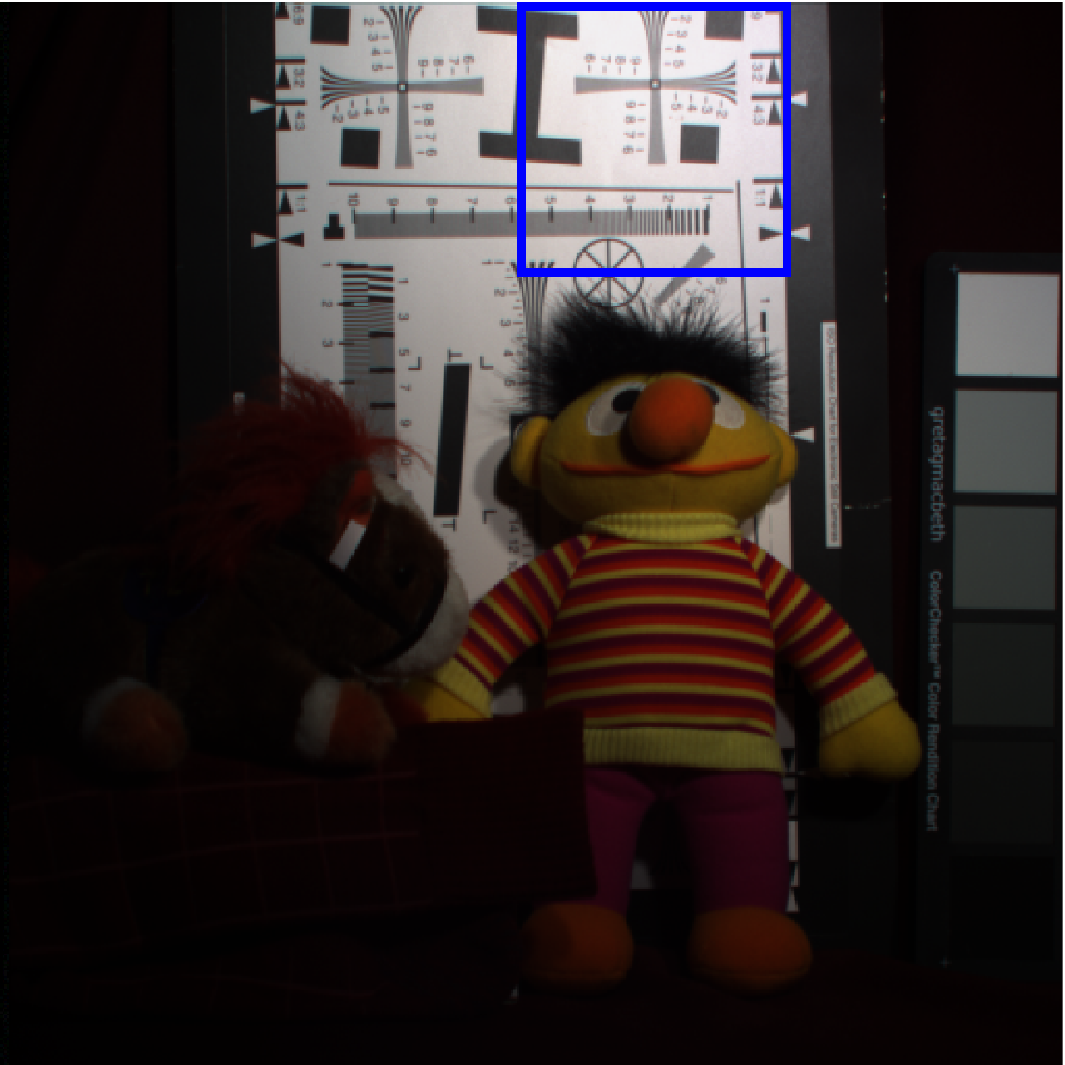}\vspace{0.2mm} \\
			\includegraphics[width=1.35in,height=1.35in]{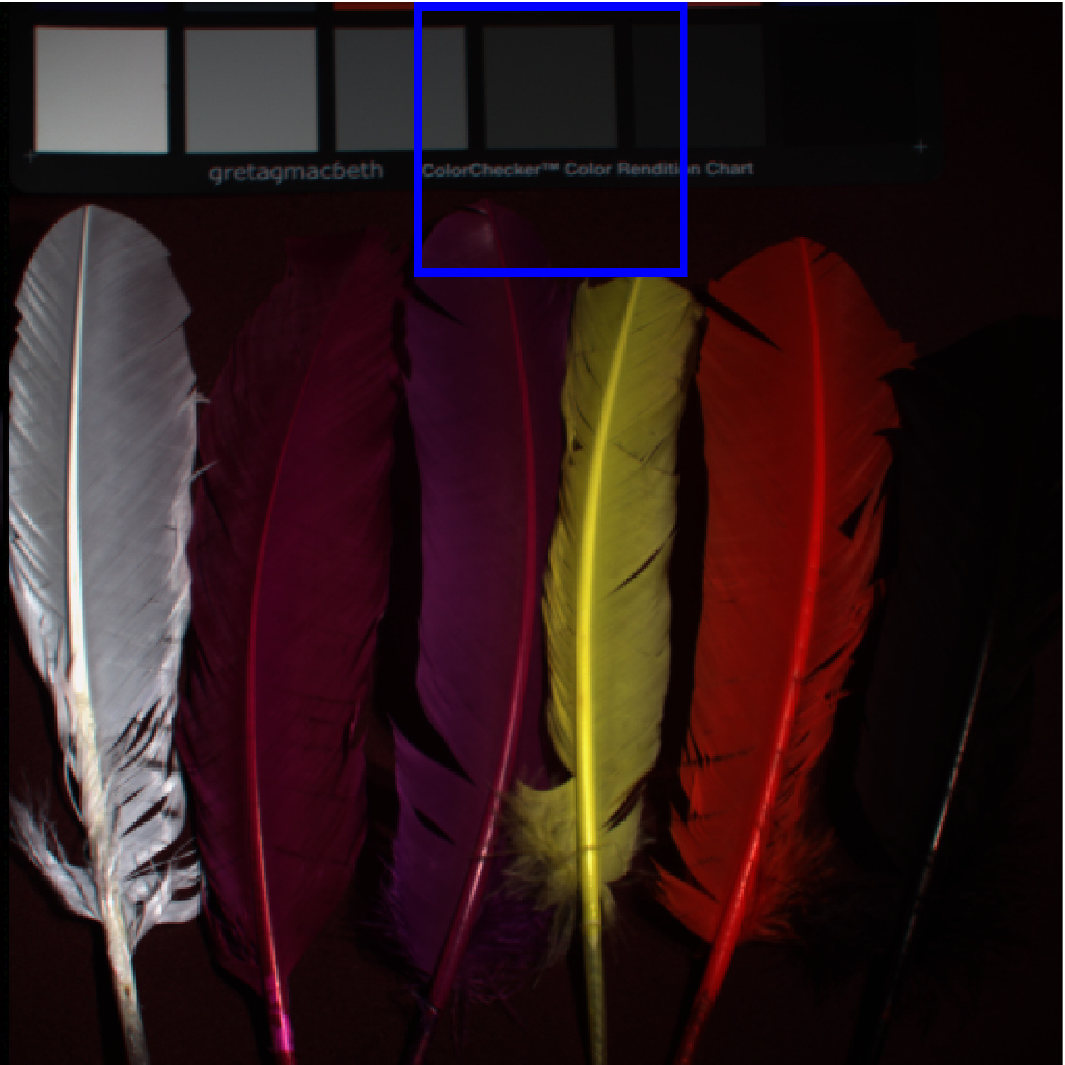}
		\end{minipage}\hspace{0.4mm}}
	\subfloat[Ours]{
		\begin{minipage}[b]{0.095\linewidth}
			\includegraphics[width=0.655in,height=0.657in]{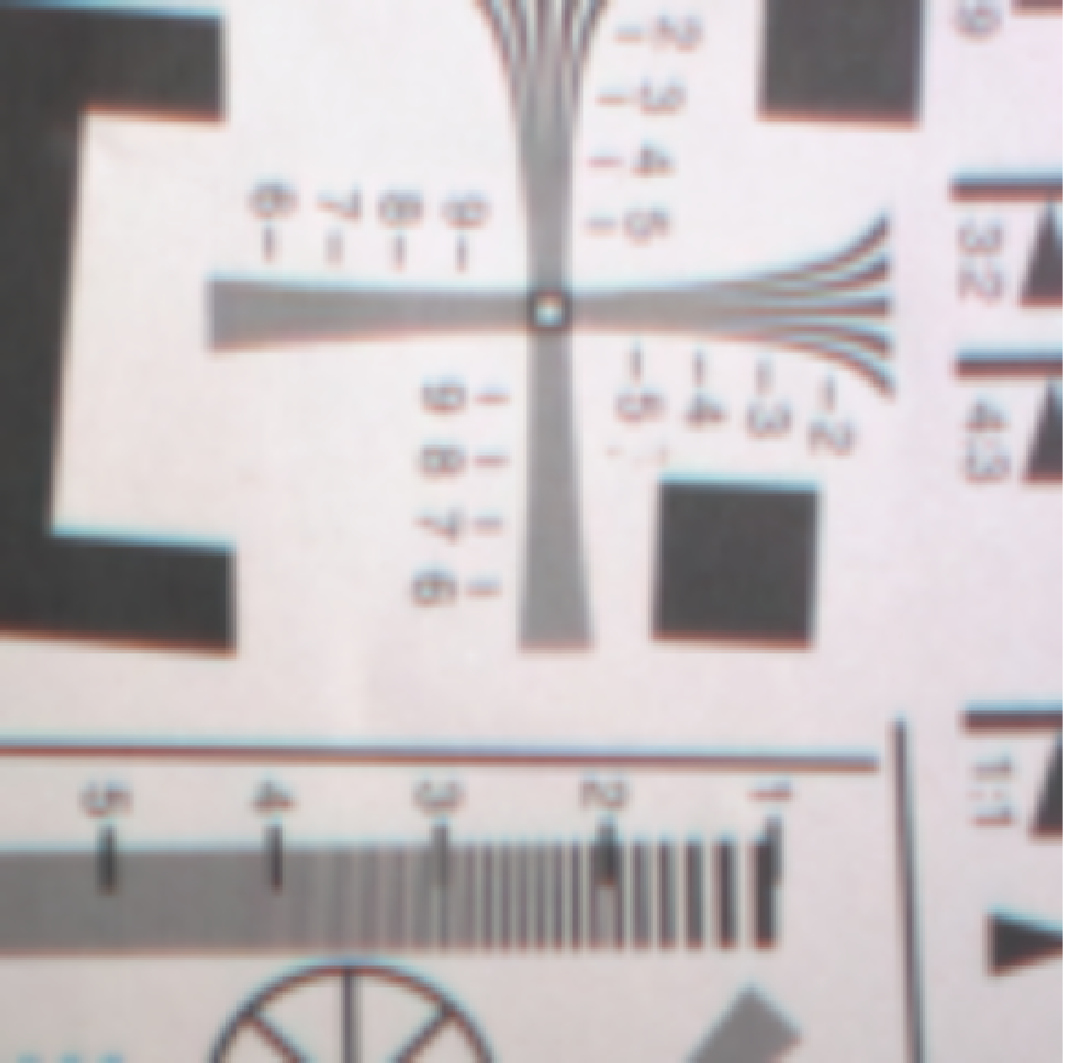}\vspace{0.4mm} \\
			\includegraphics[width=0.655in,height=0.657in]{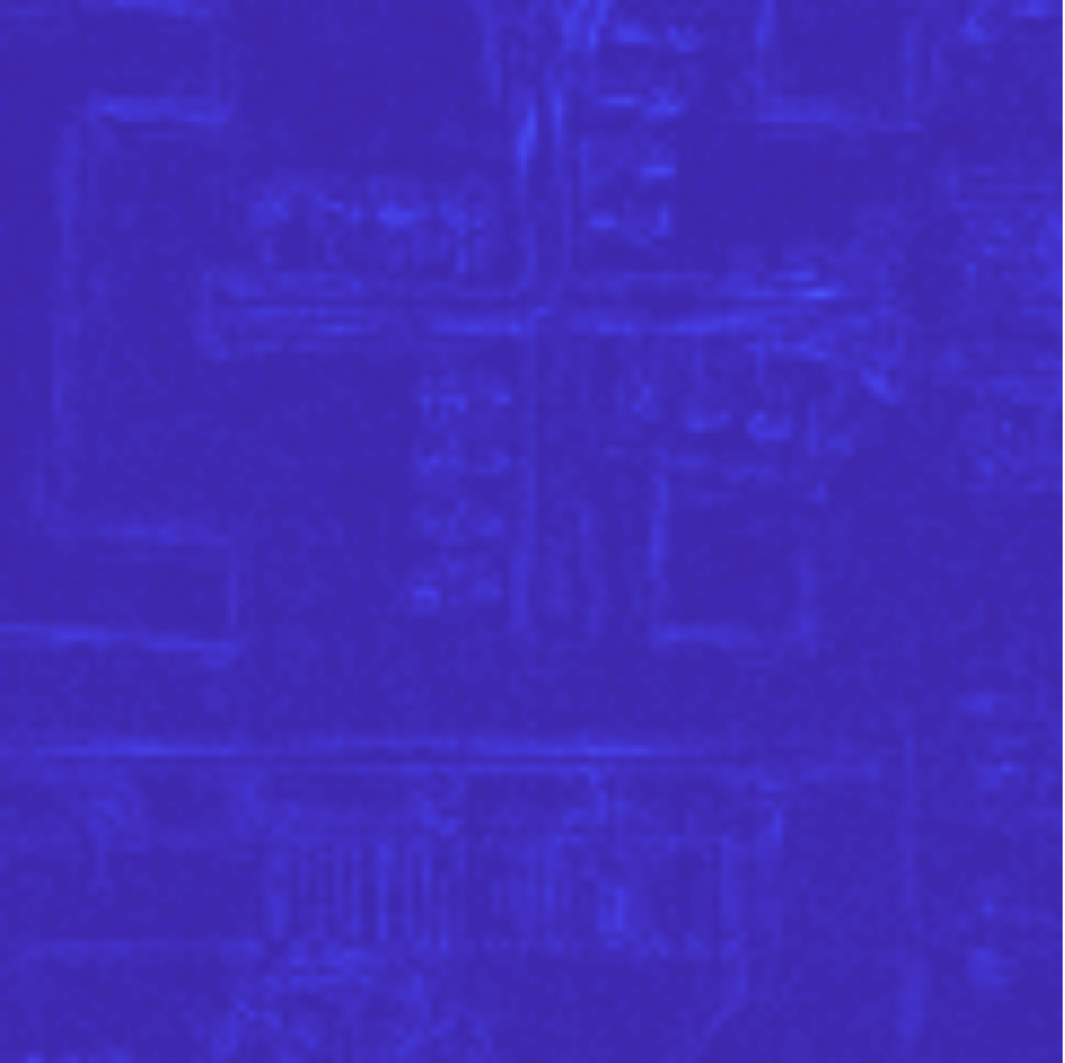}\vspace{0.4mm} \\
			\includegraphics[width=0.655in,height=0.657in]{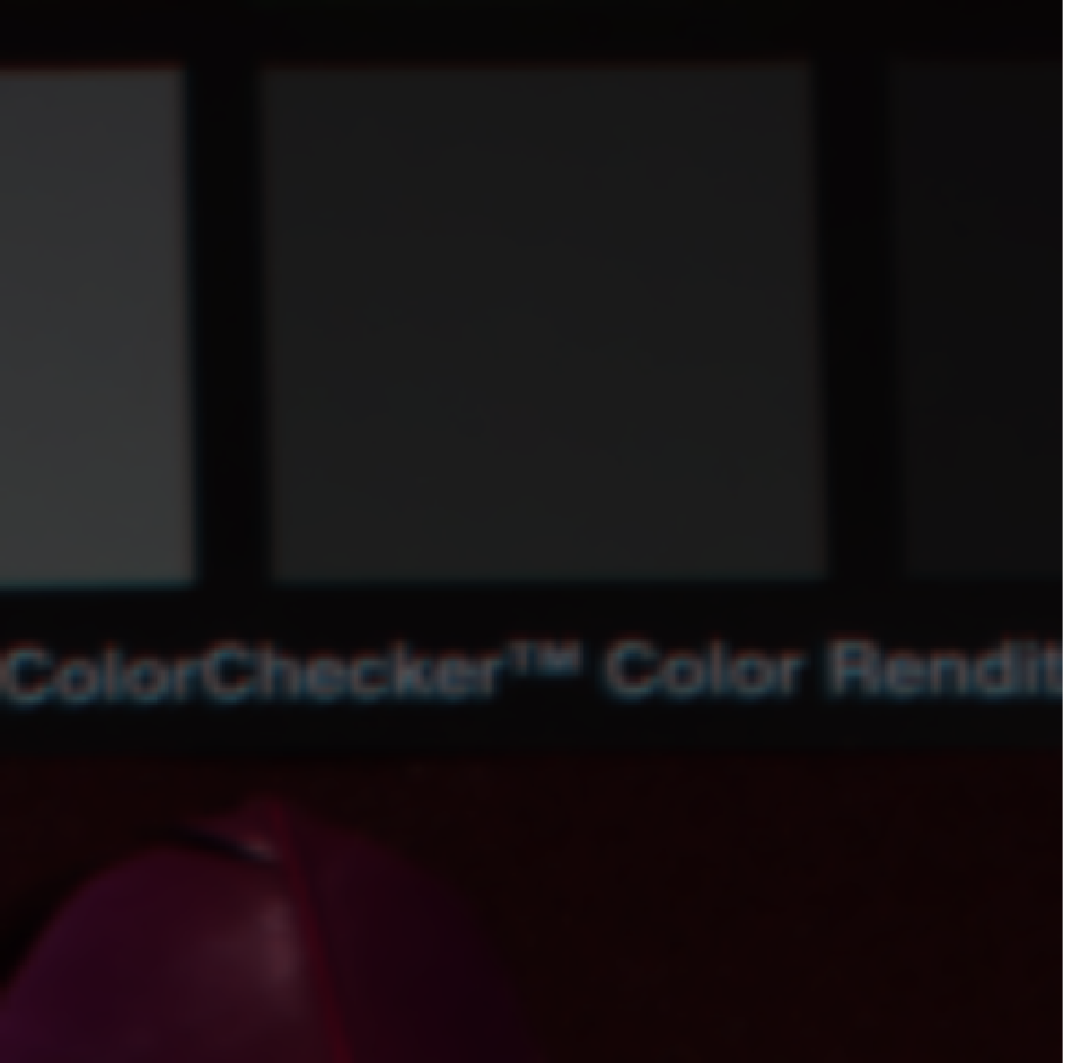}\vspace{0.4mm} \\
			\includegraphics[width=0.655in,height=0.657in]{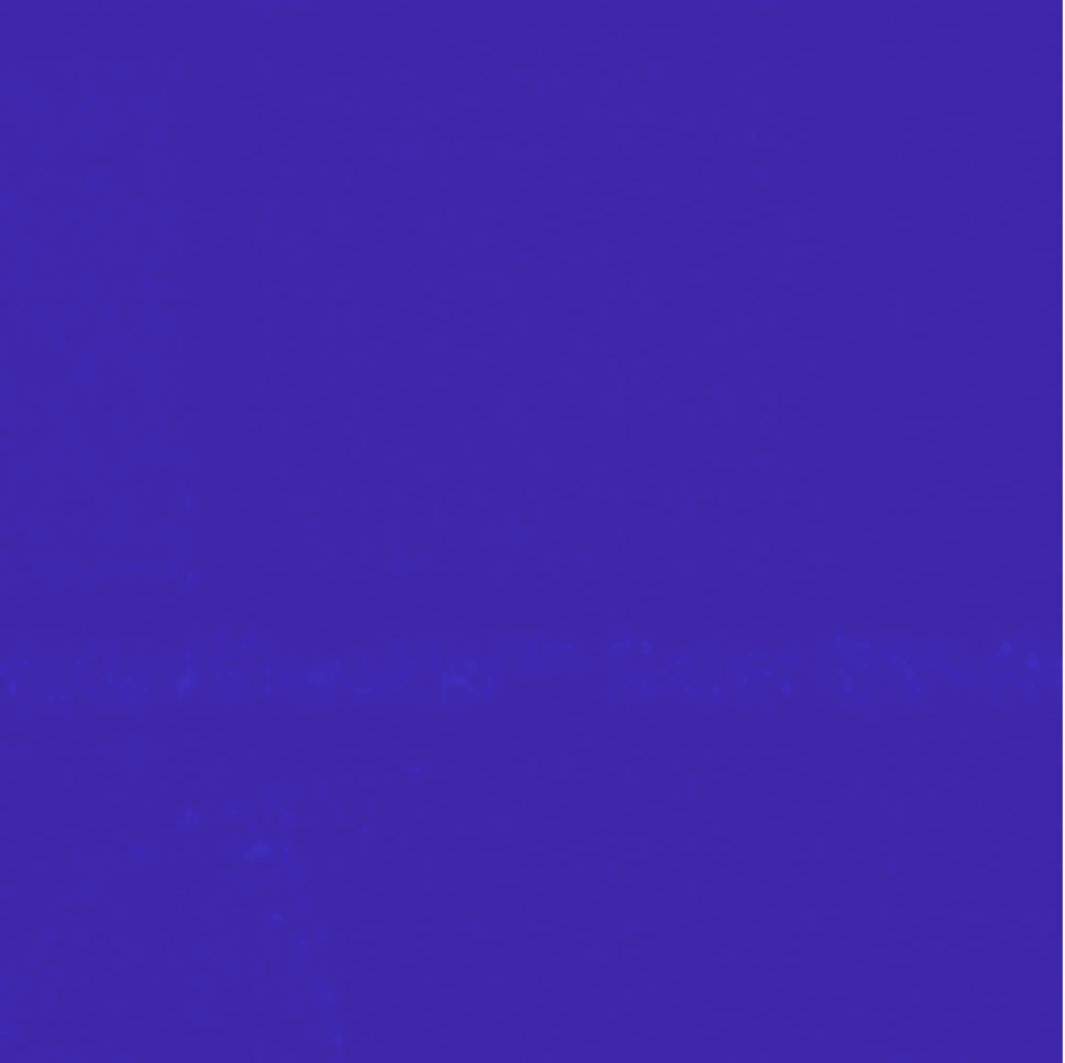} 
		\end{minipage}\hspace{-0.3mm}}
	\subfloat[DHIF]{
		\begin{minipage}[b]{0.095\linewidth}
			\includegraphics[width=0.655in,height=0.657in]{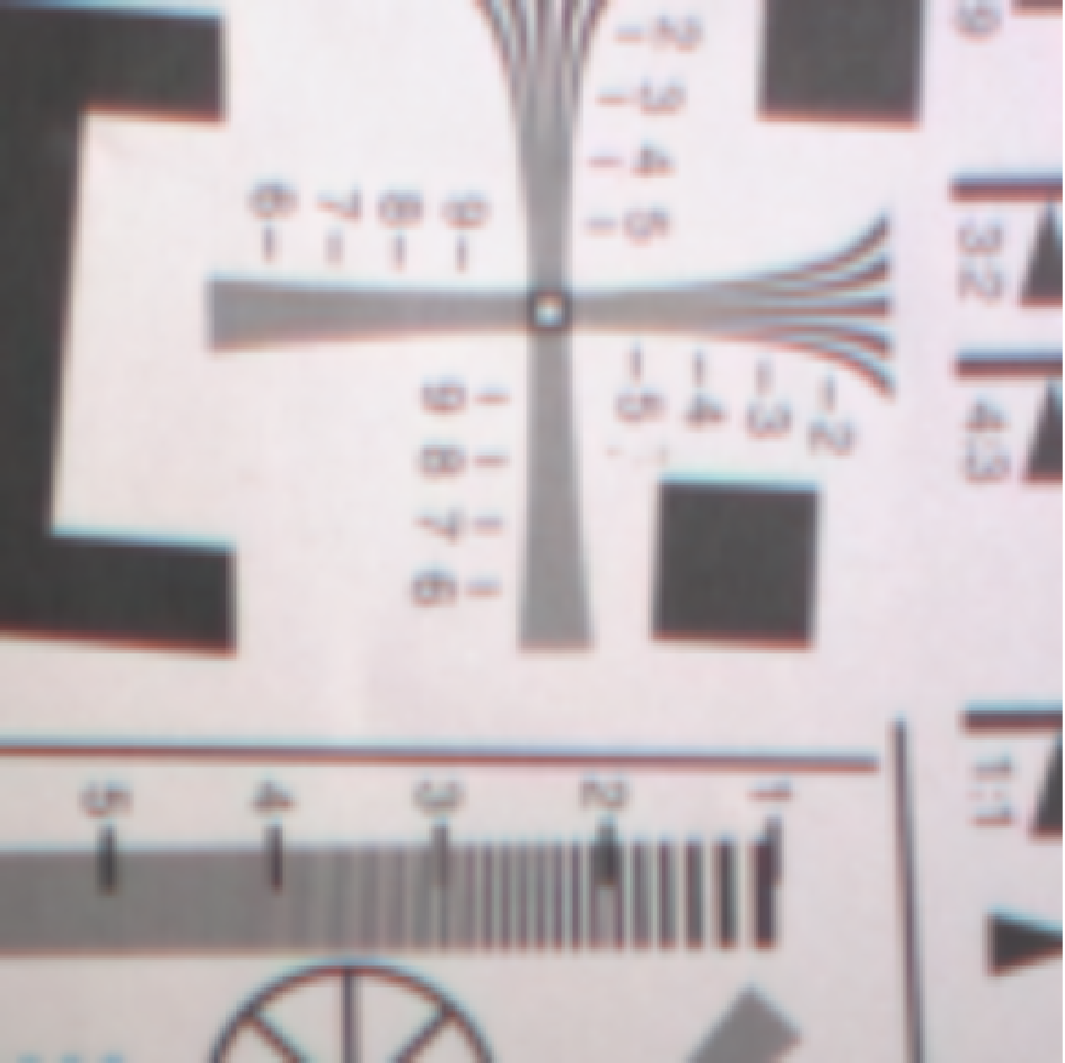}\vspace{0.4mm} \\
			\includegraphics[width=0.655in,height=0.657in]{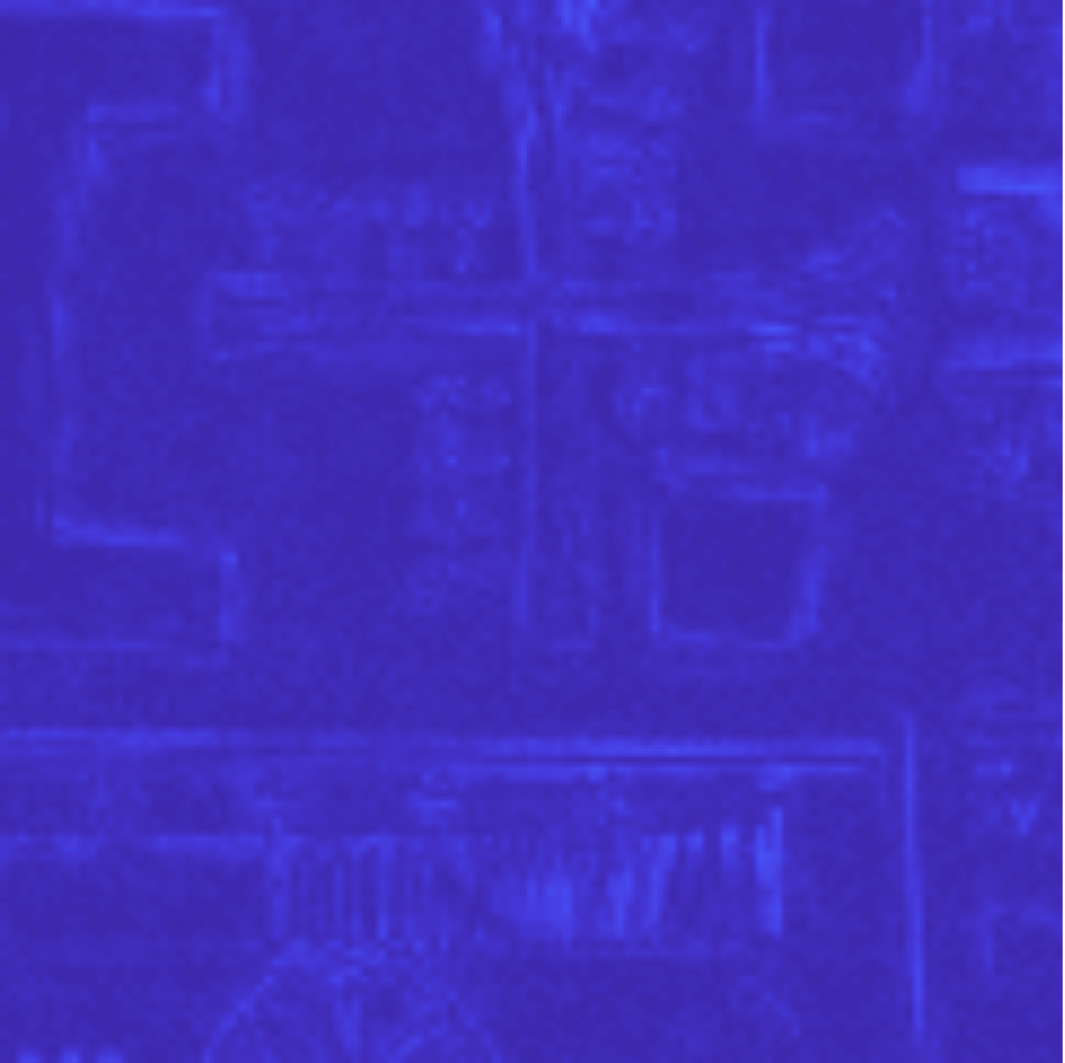}\vspace{0.4mm} \\
			\includegraphics[width=0.655in,height=0.657in]{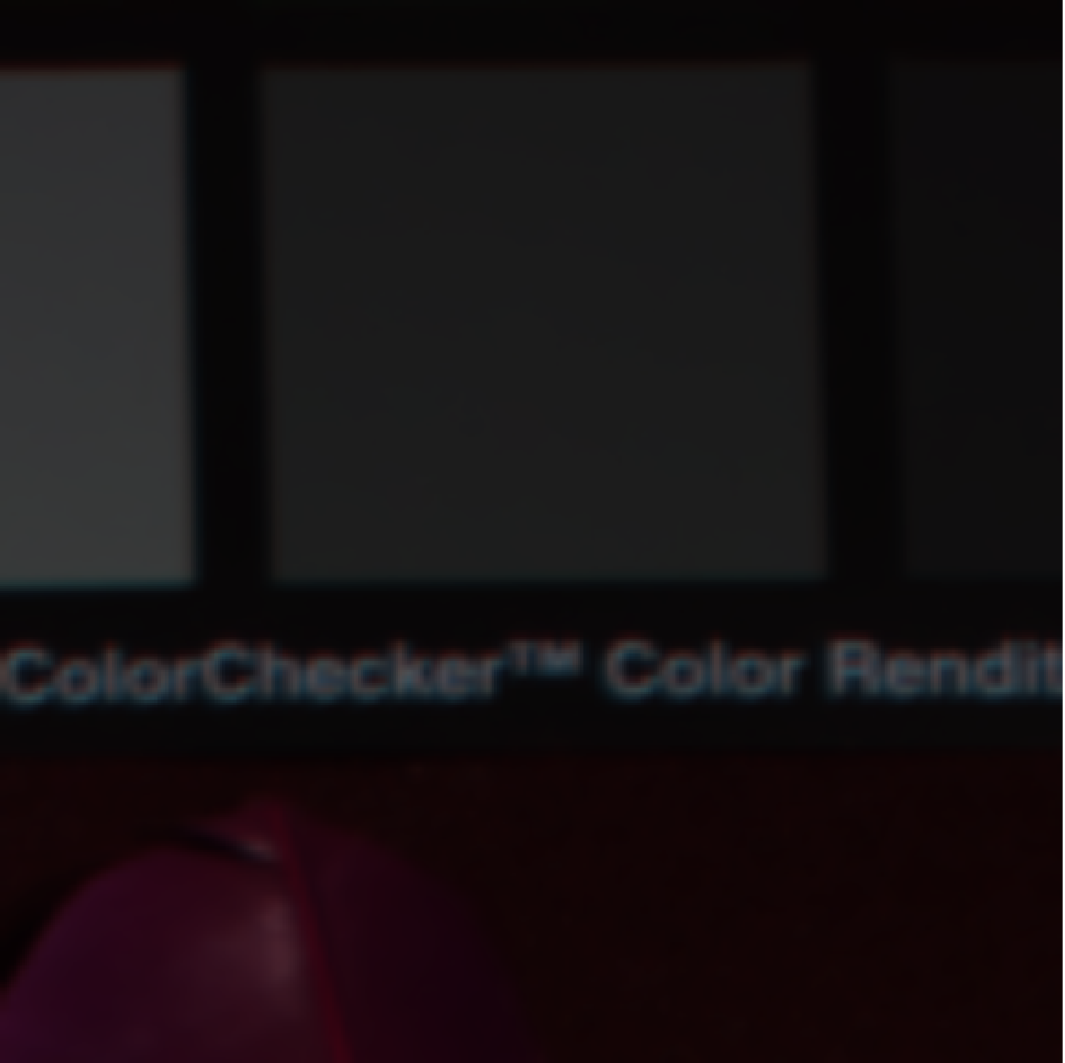}\vspace{0.4mm} \\
			\includegraphics[width=0.655in,height=0.657in]{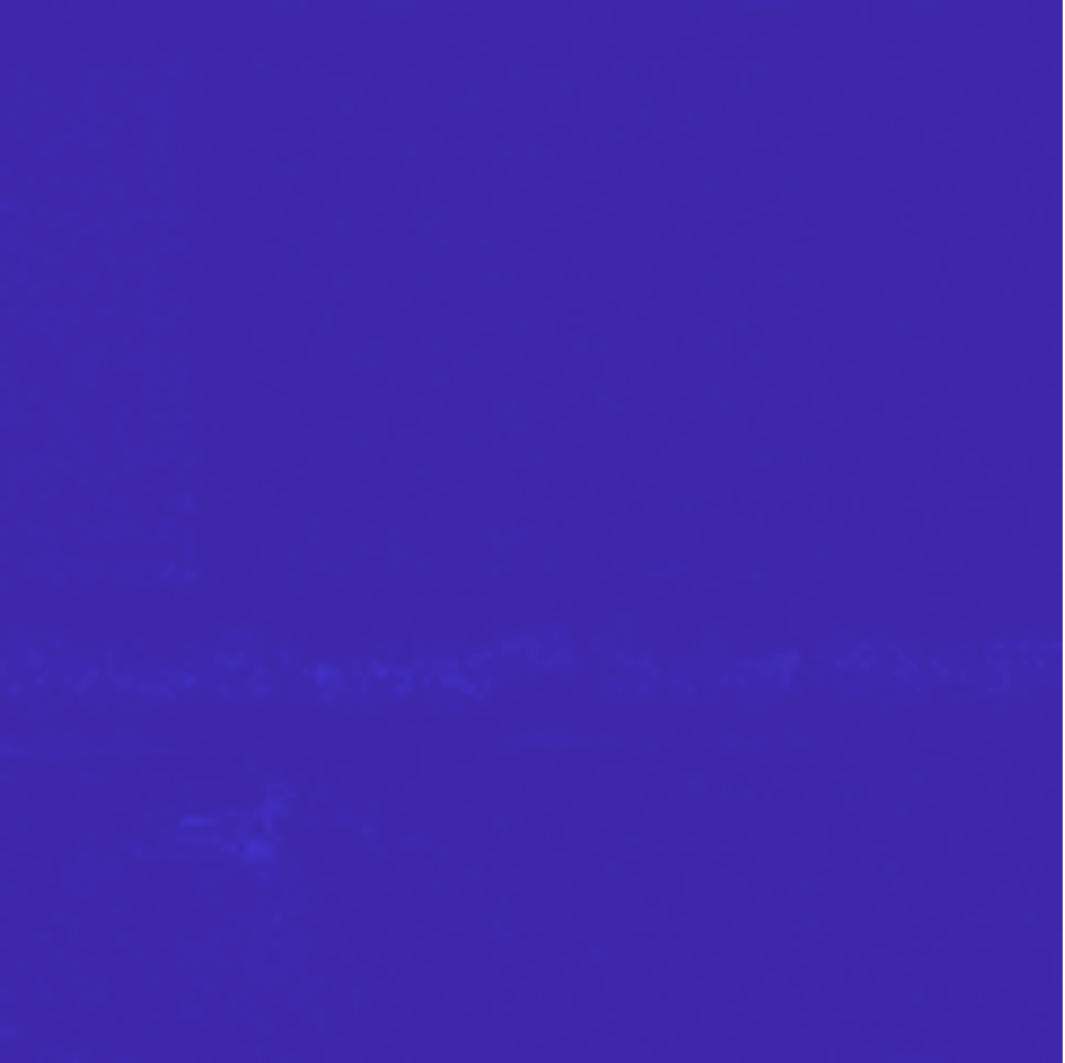}
		\end{minipage}\hspace{-0.3mm}}
	\subfloat[Fusformer]{
		\begin{minipage}[b]{0.095\linewidth}
			\includegraphics[width=0.655in,height=0.657in]{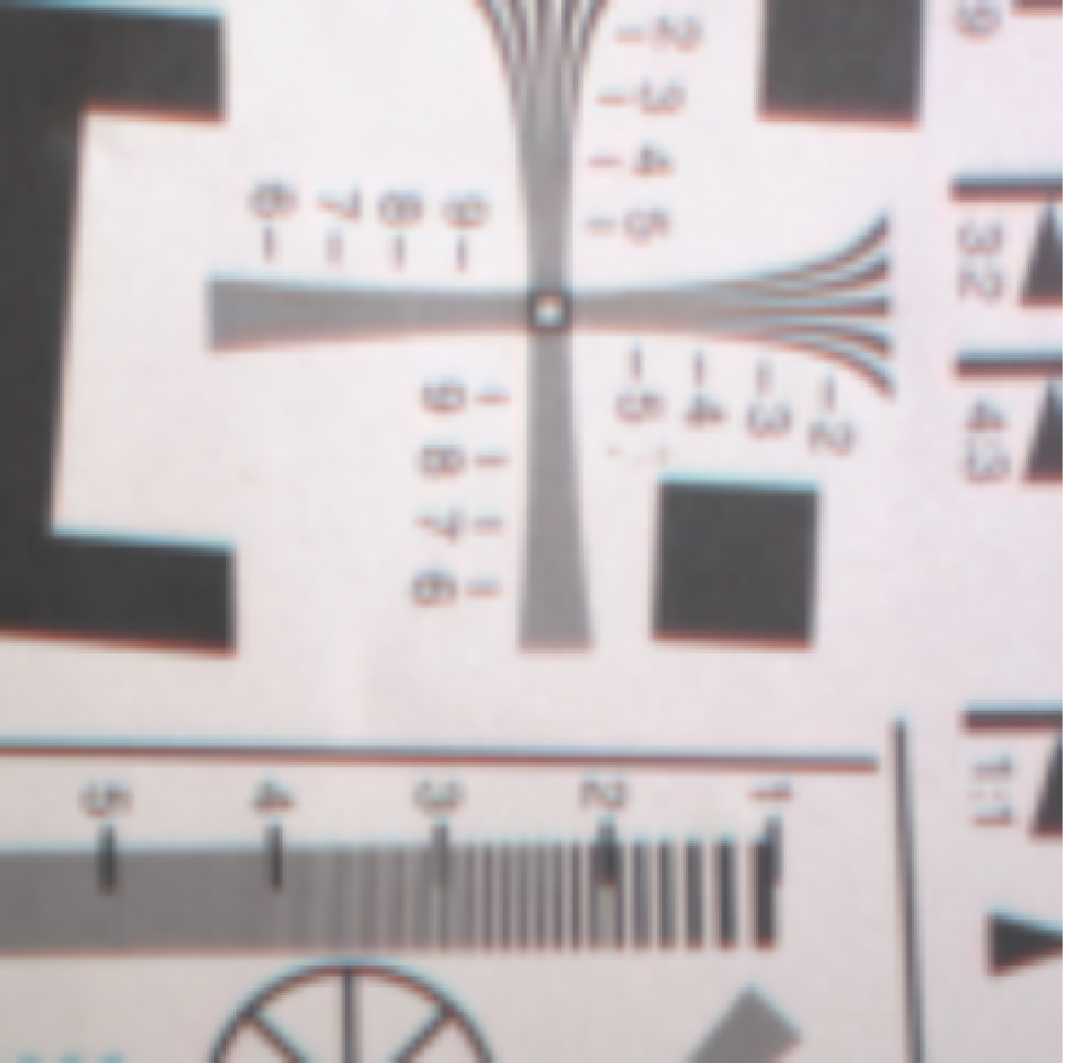}\vspace{0.4mm} \\
			\includegraphics[width=0.655in,height=0.657in]{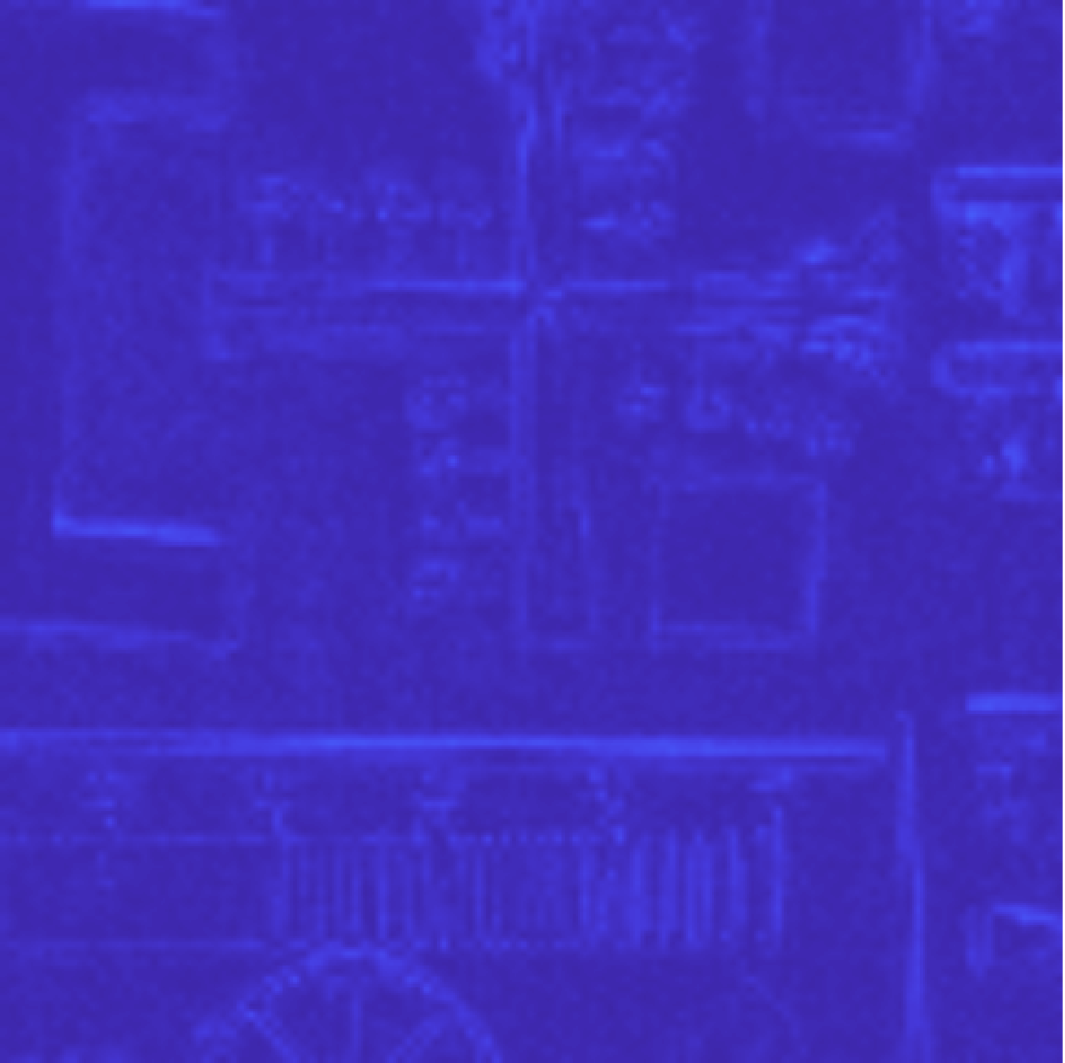}\vspace{0.4mm} \\
			\includegraphics[width=0.655in,height=0.657in]{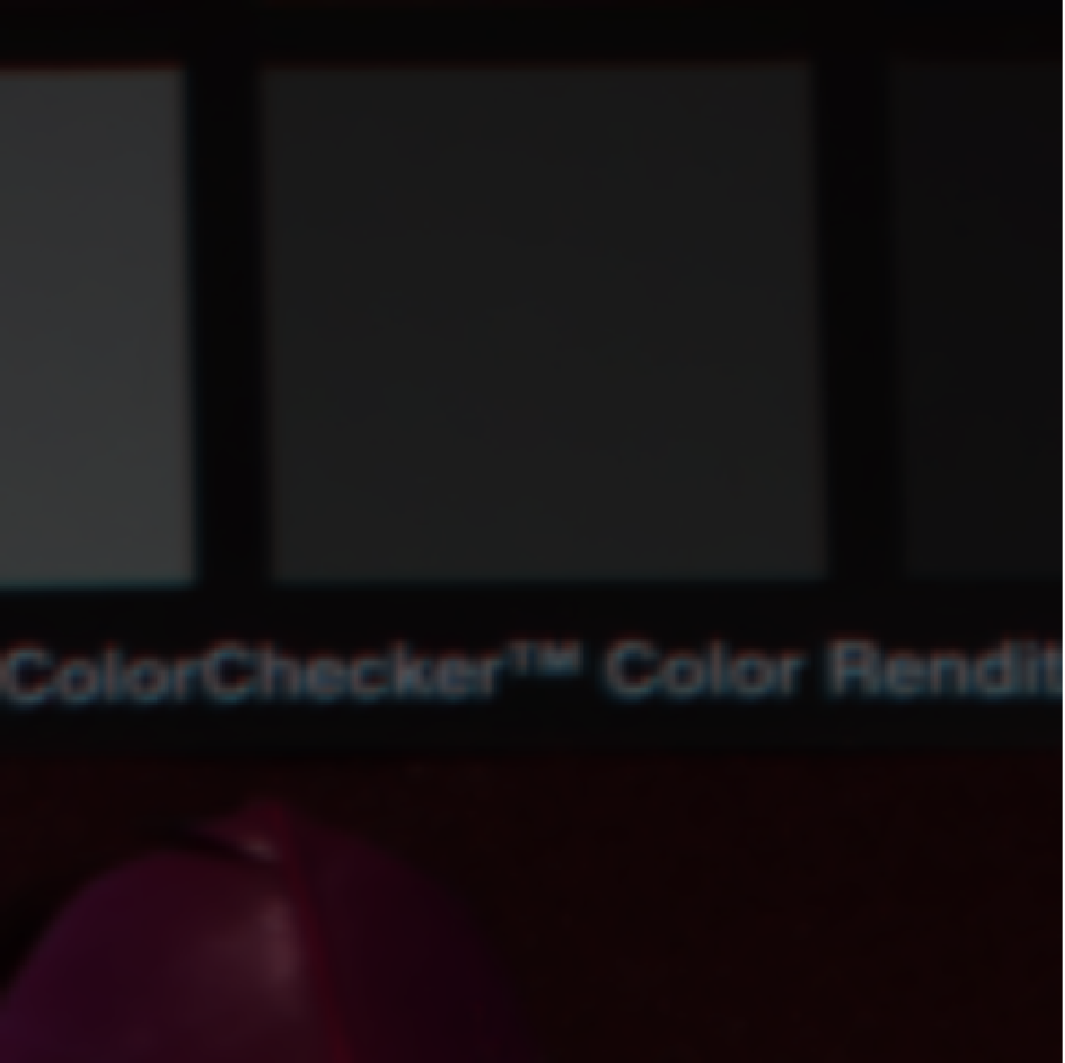}\vspace{0.4mm} \\
			\includegraphics[width=0.655in,height=0.657in]{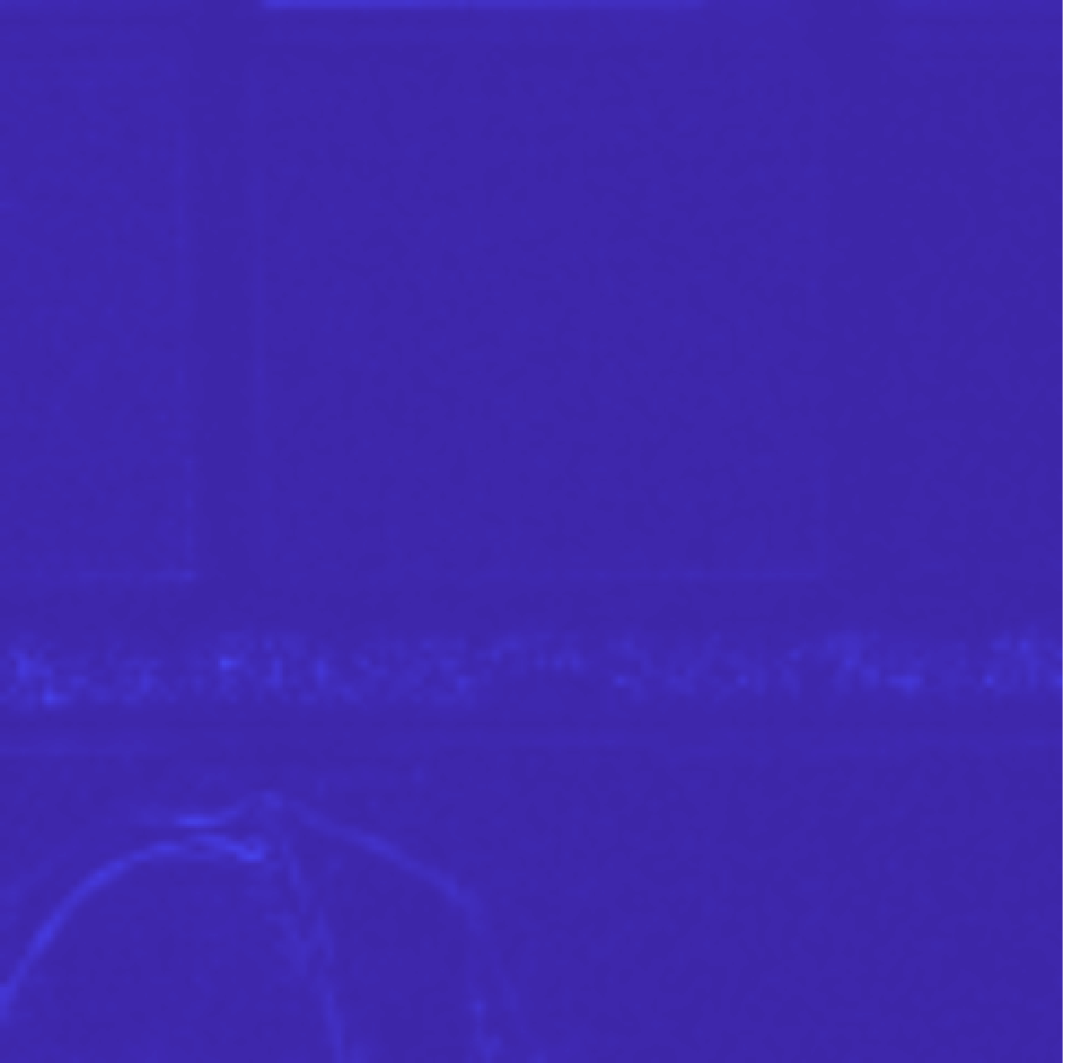}
		\end{minipage}\hspace{-0.3mm}}
	\subfloat[MoG-DCN]{
		\begin{minipage}[b]{0.095\linewidth}
			\includegraphics[width=0.655in,height=0.657in]{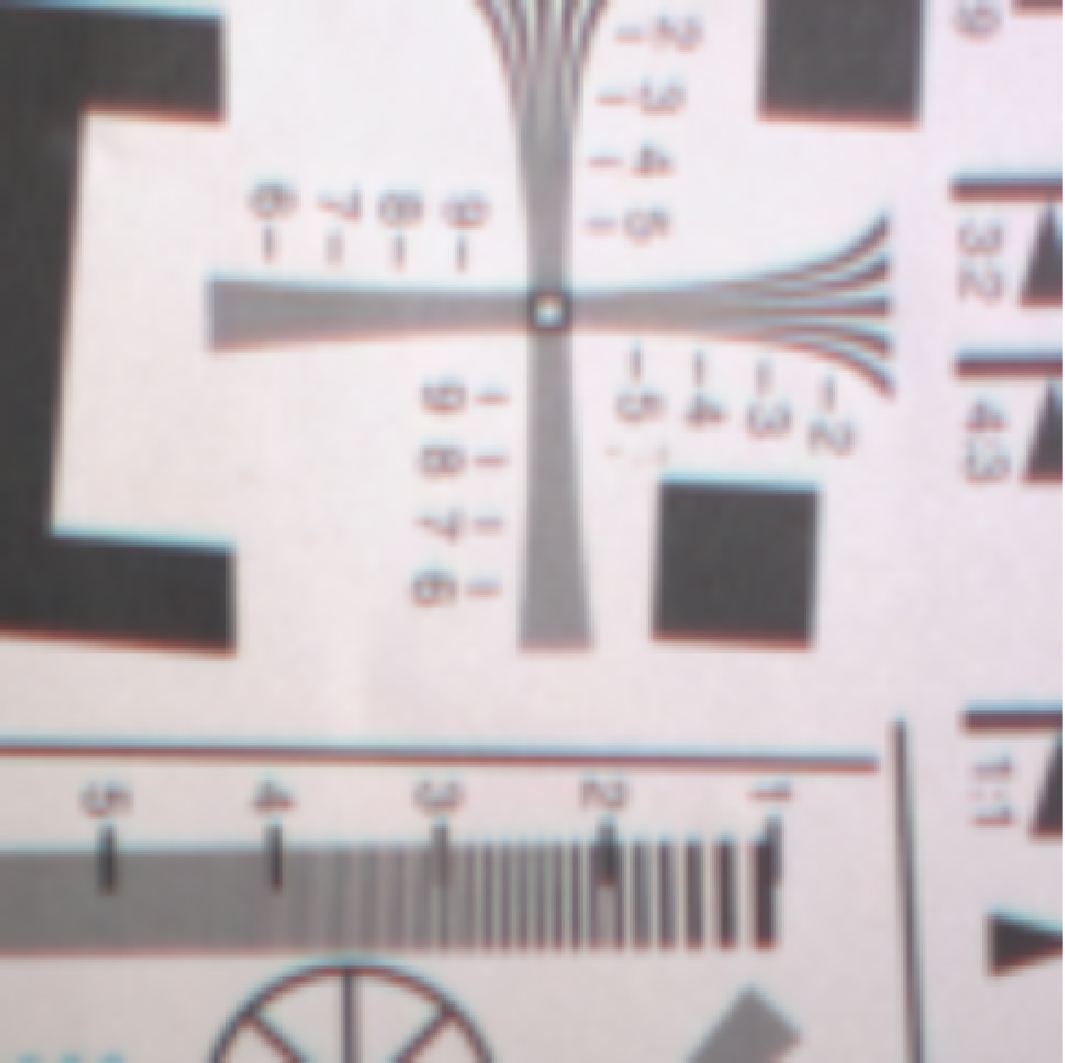}\vspace{0.4mm} \\
			\includegraphics[width=0.655in,height=0.657in]{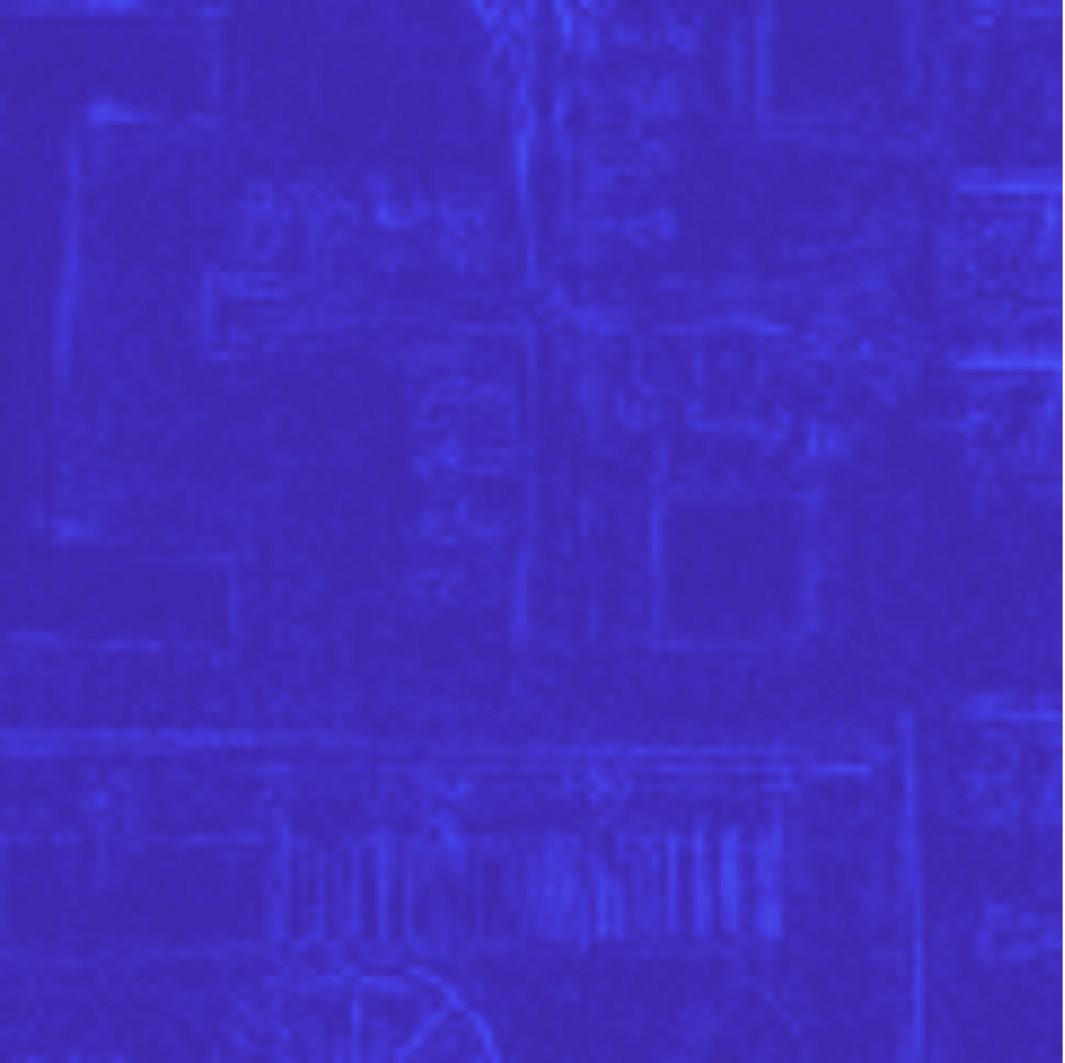}\vspace{0.4mm} \\
			\includegraphics[width=0.655in,height=0.657in]{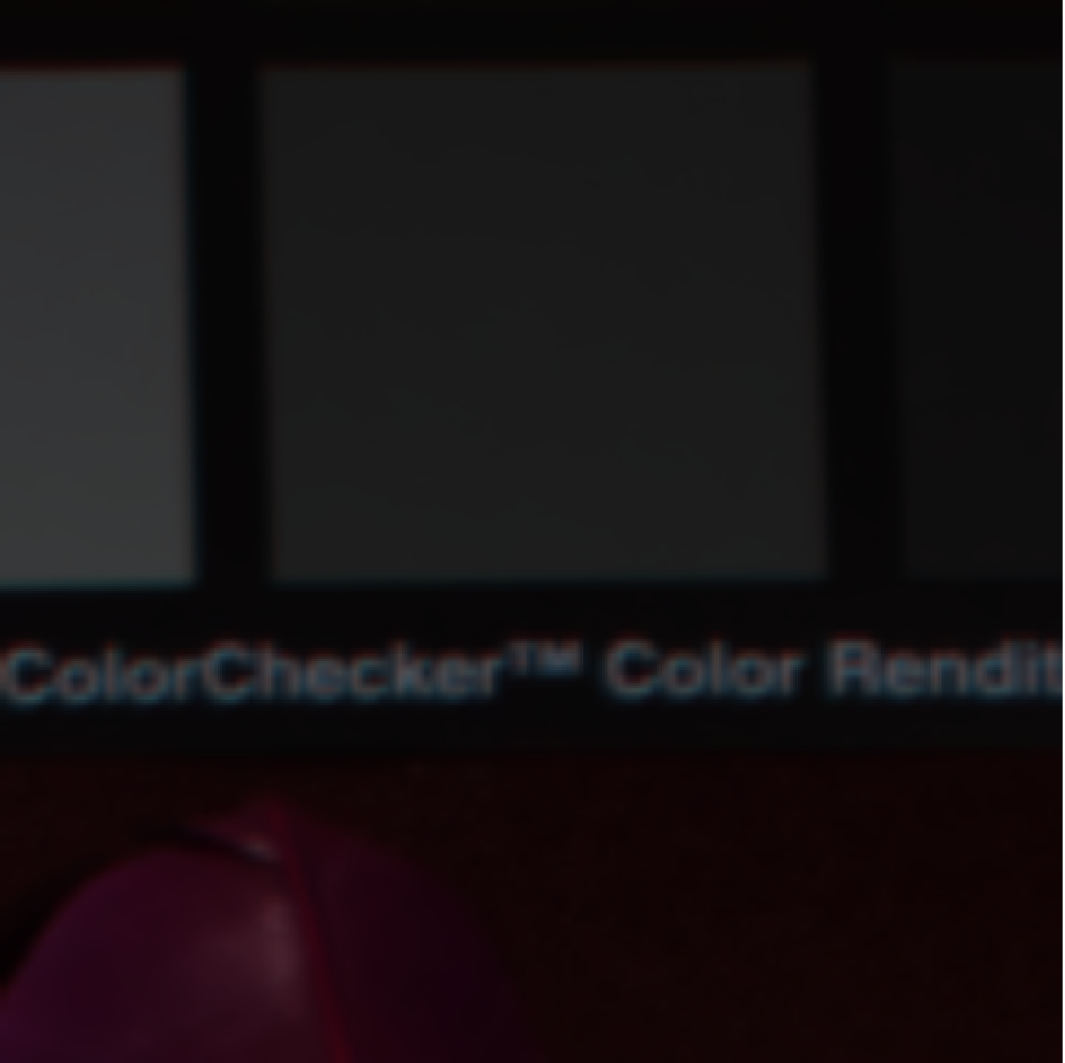}\vspace{0.4mm} \\
			\includegraphics[width=0.655in,height=0.657in]{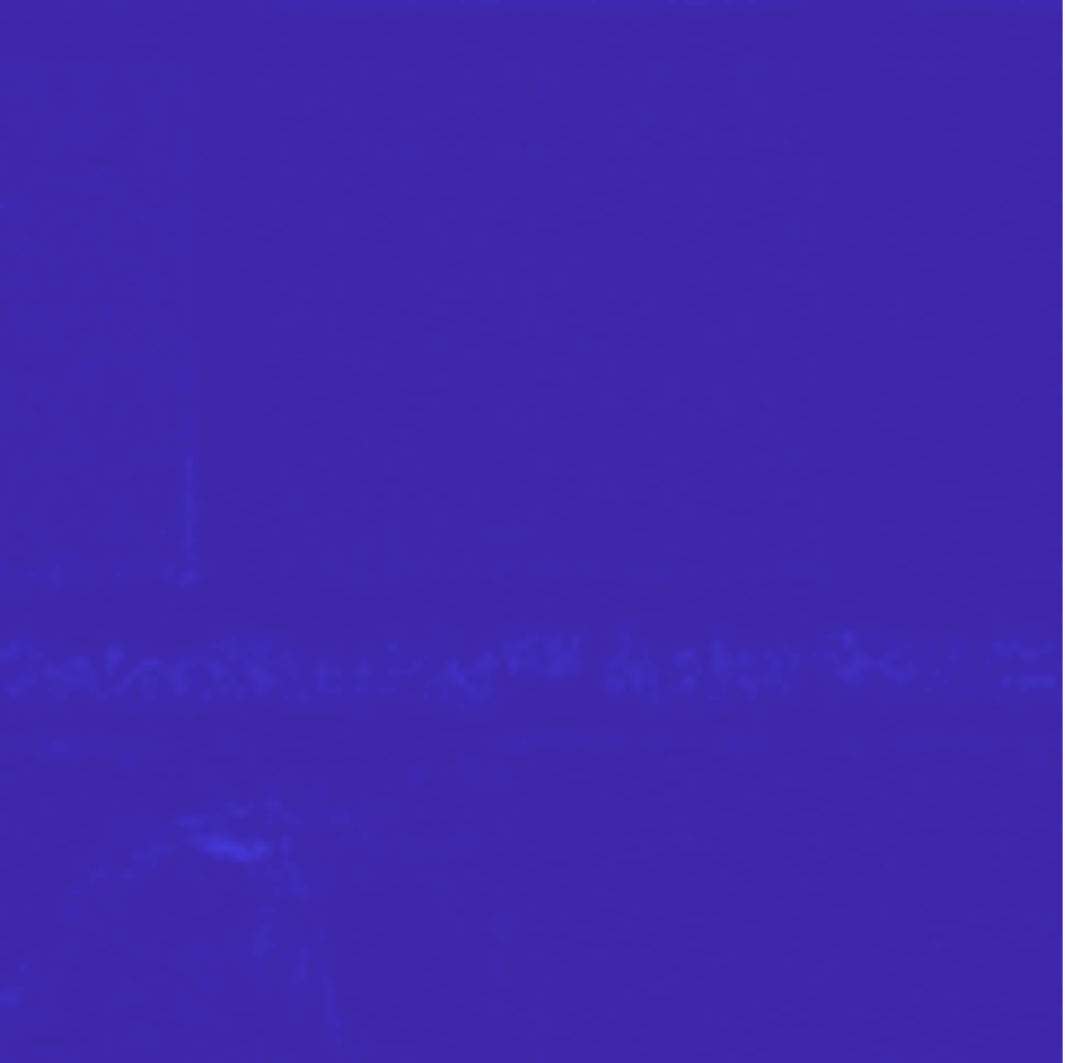}
		\end{minipage}\hspace{-0.3mm}}  
	\subfloat[HSRNet]{
		\begin{minipage}[b]{0.095\linewidth}
			\includegraphics[width=0.655in,height=0.657in]{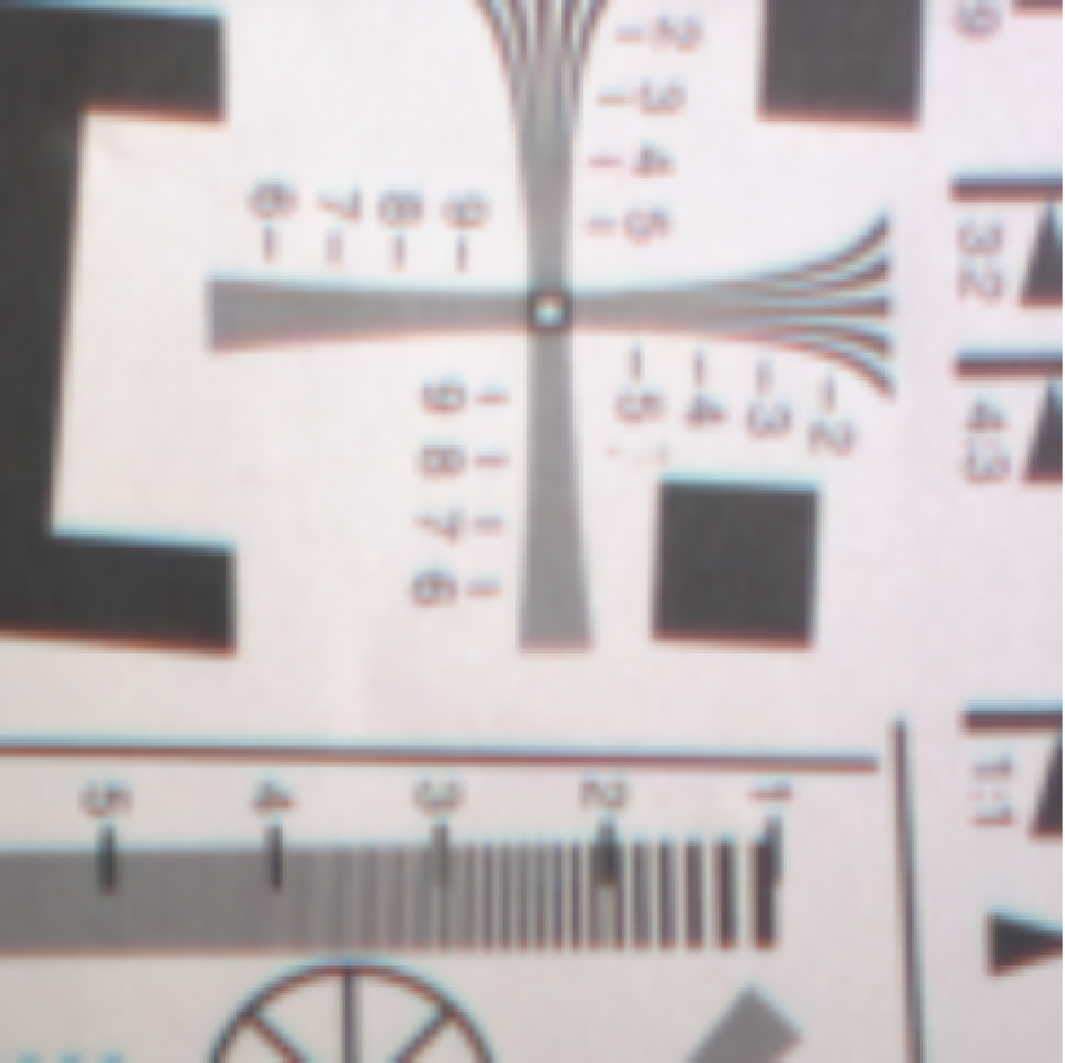}\vspace{0.4mm} \\
			\includegraphics[width=0.655in,height=0.657in]{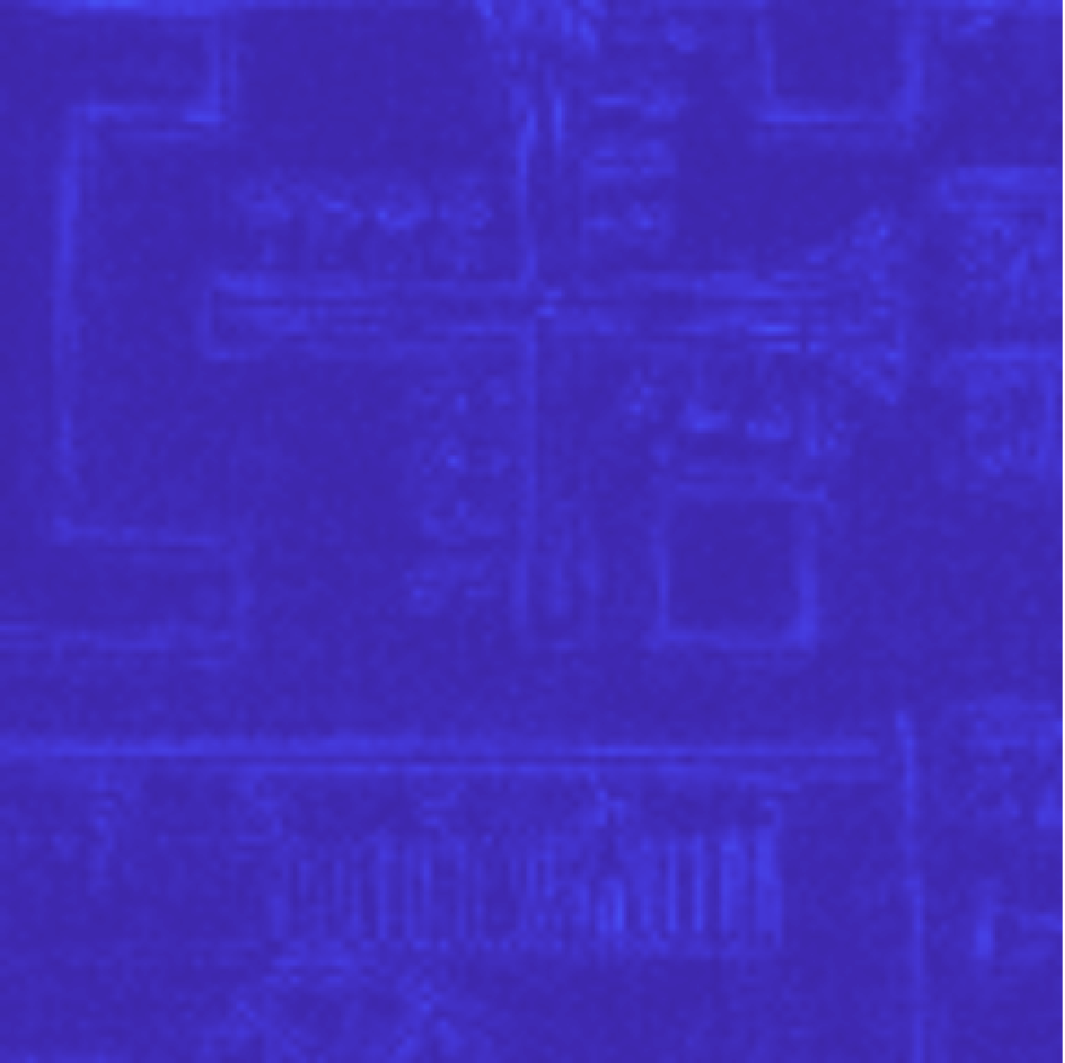}\vspace{0.4mm} \\
			\includegraphics[width=0.655in,height=0.657in]{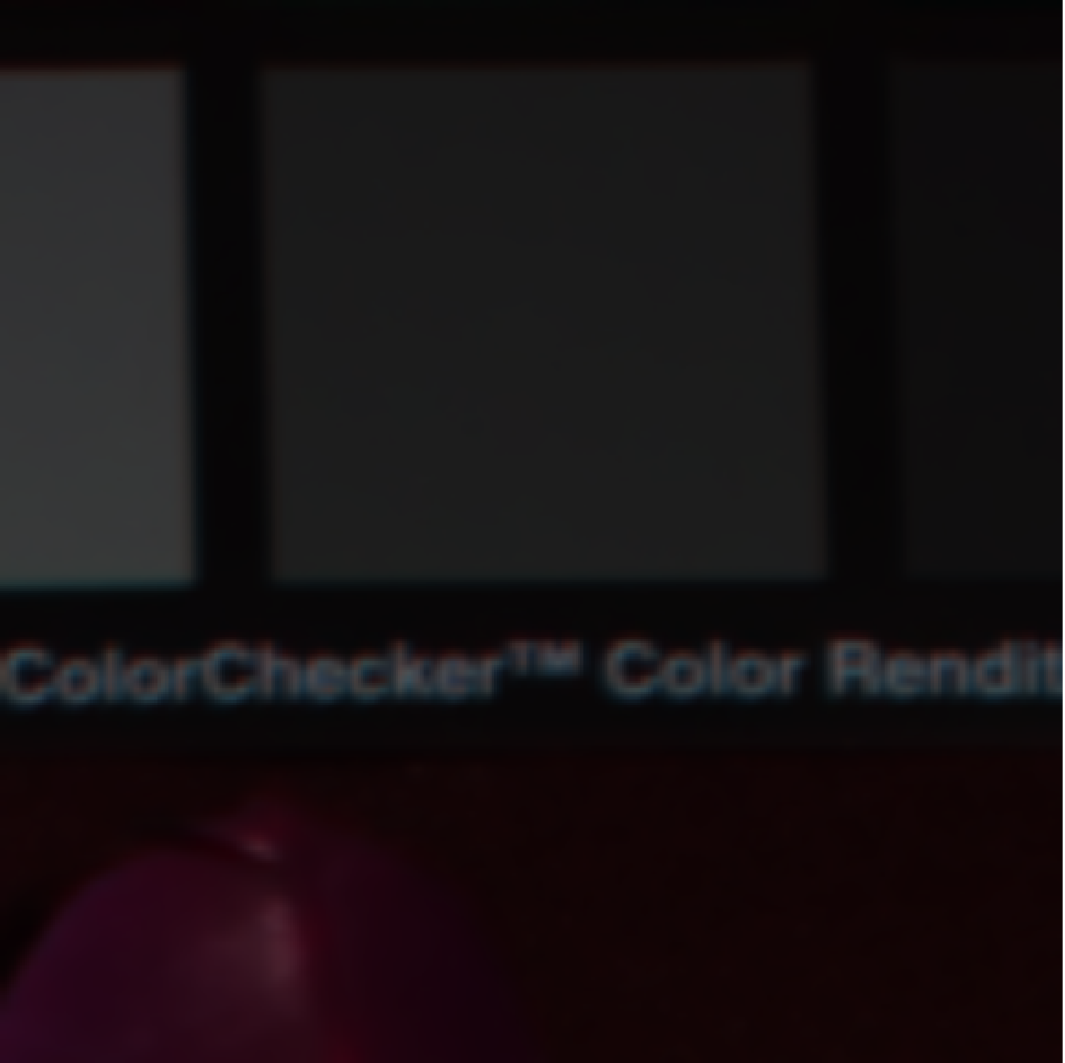}\vspace{0.4mm} \\
			\includegraphics[width=0.655in,height=0.657in]{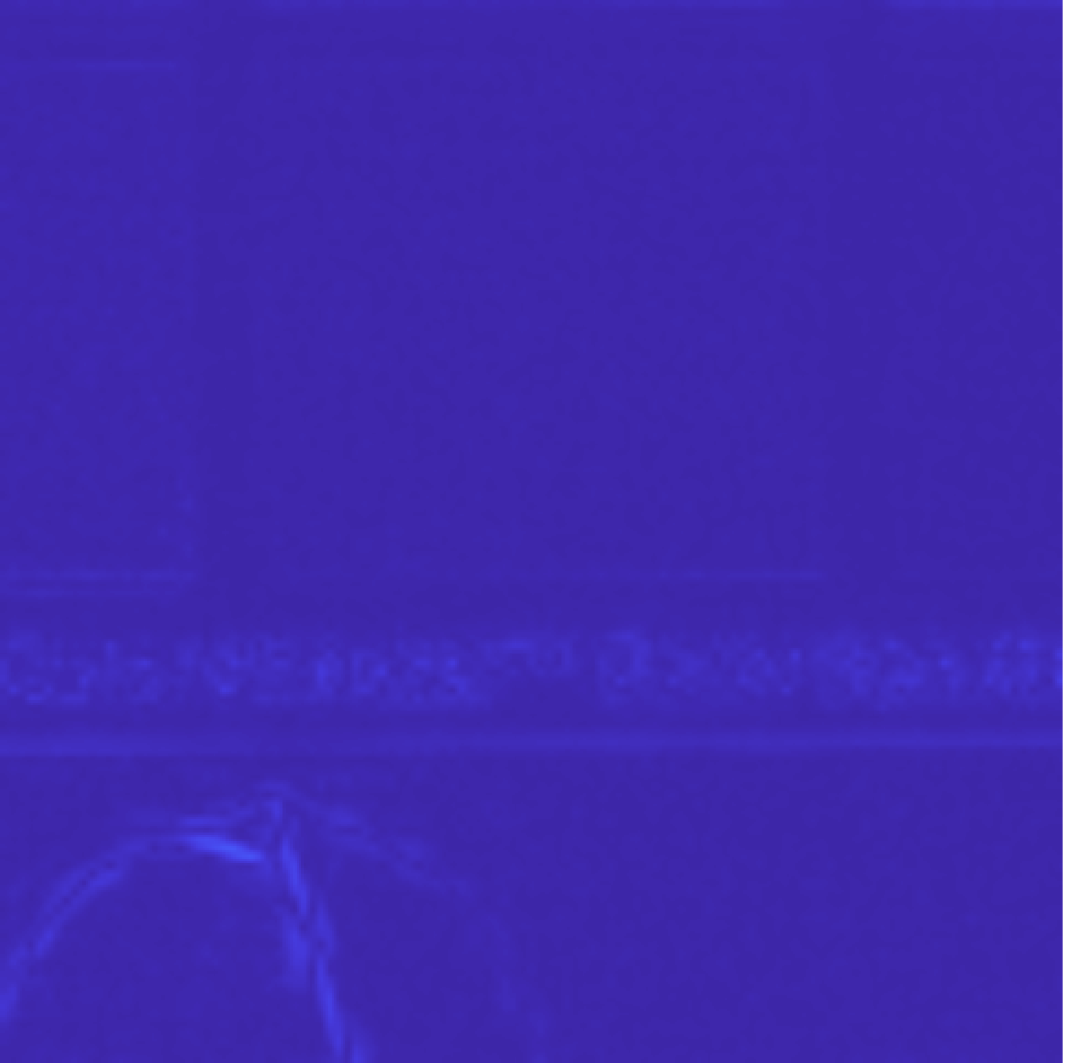}
		\end{minipage}\hspace{-0.3mm}}
	\subfloat[SSRNet]{
		\begin{minipage}[b]{0.095\linewidth}
			\includegraphics[width=0.655in,height=0.657in]{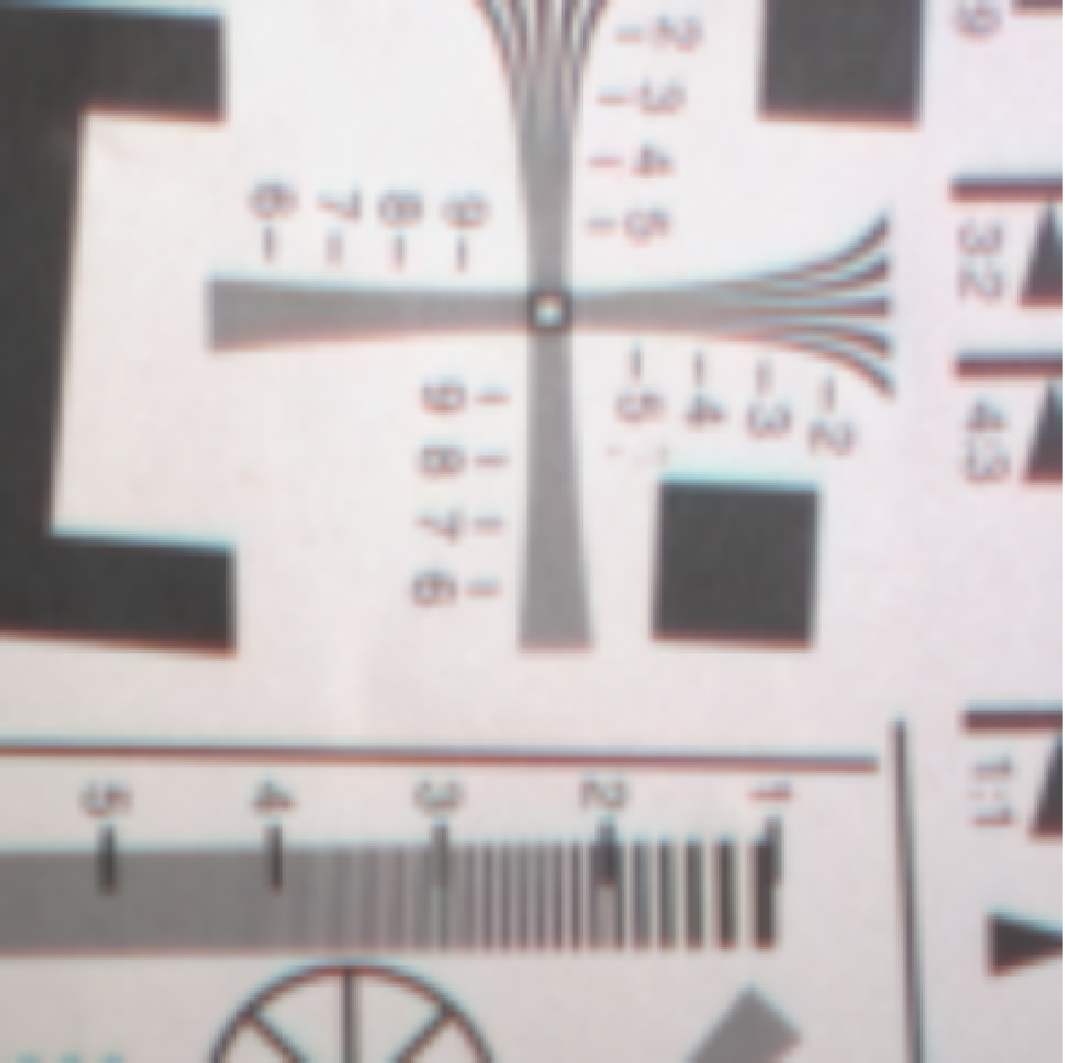}\vspace{0.4mm} \\
			\includegraphics[width=0.655in,height=0.657in]{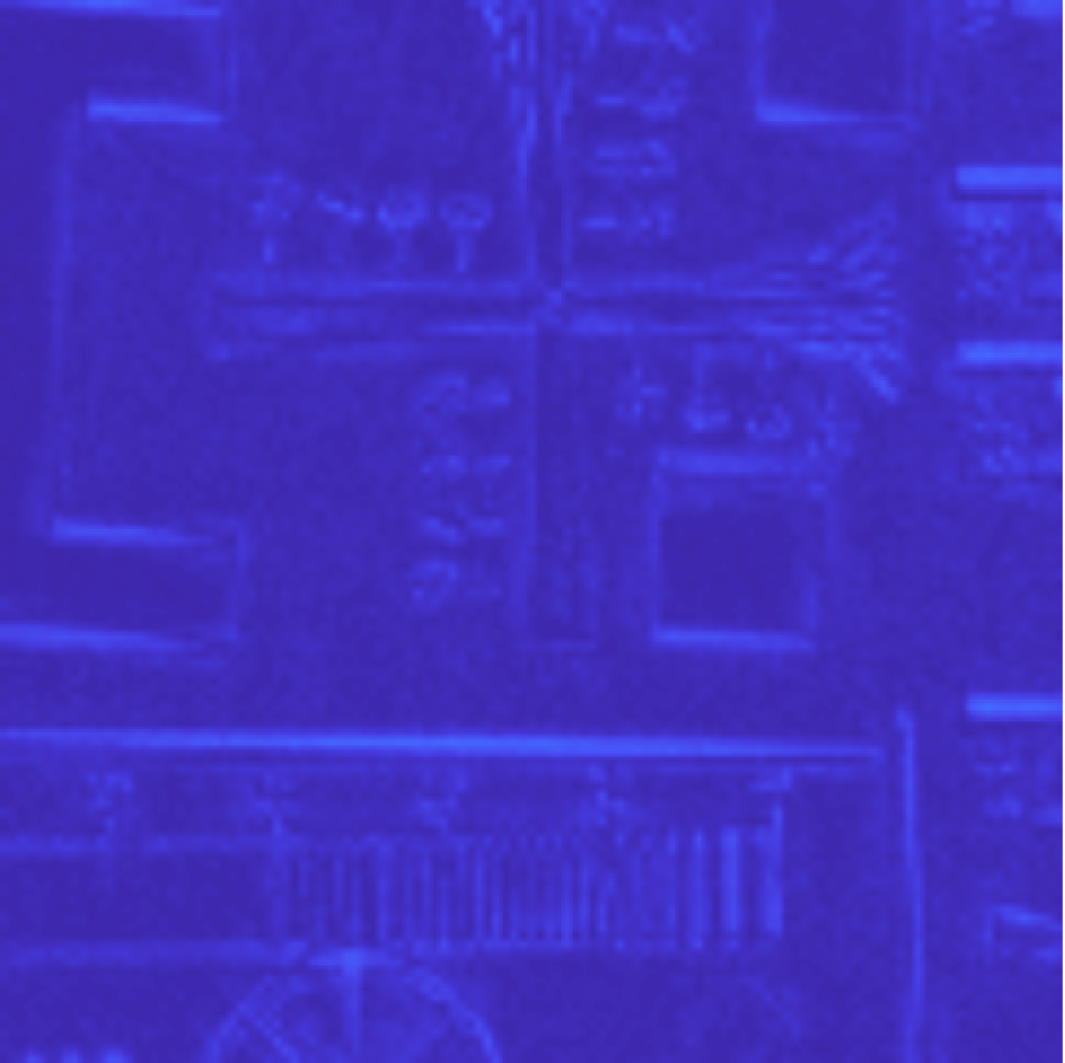}\vspace{0.4mm} \\
			\includegraphics[width=0.655in,height=0.657in]{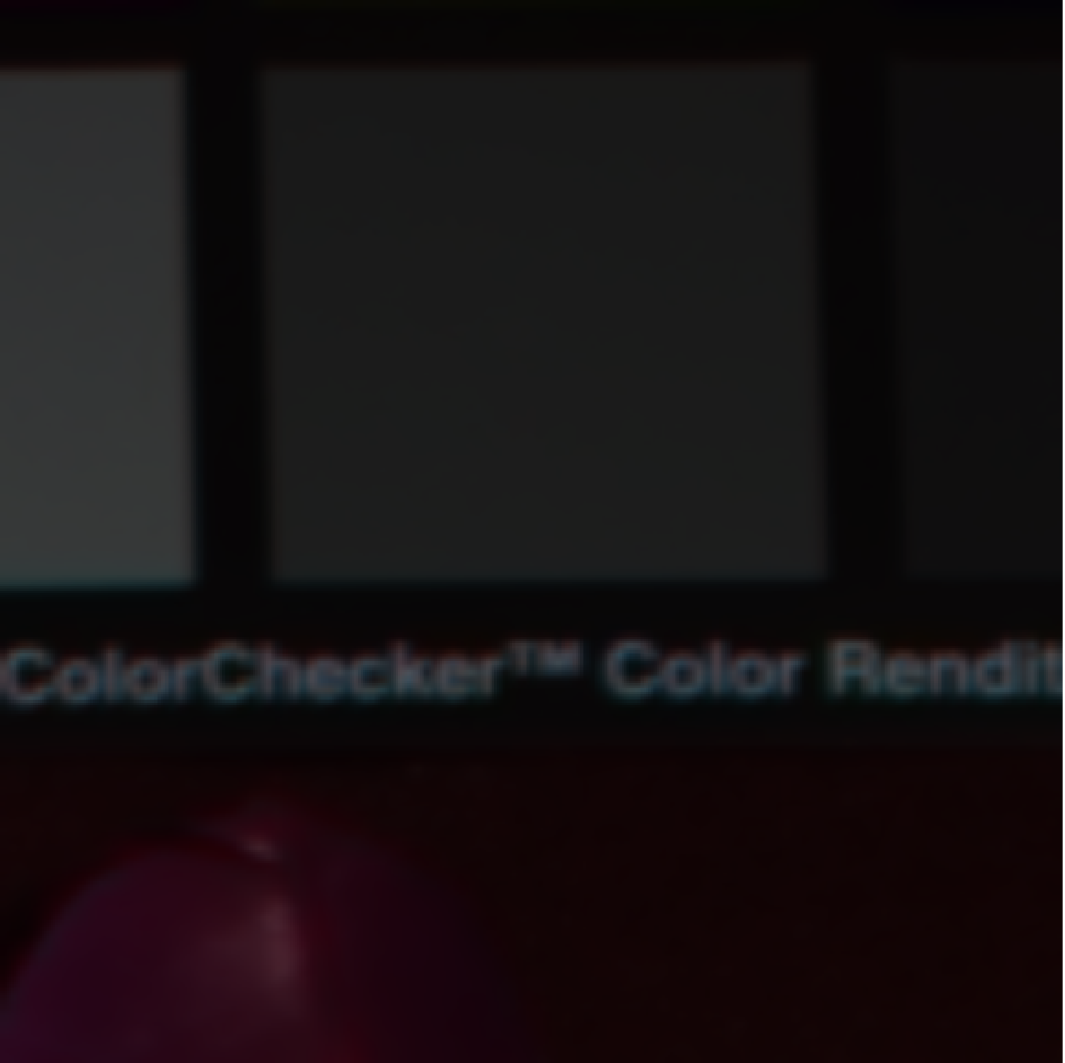}\vspace{0.4mm} \\
			\includegraphics[width=0.655in,height=0.657in]{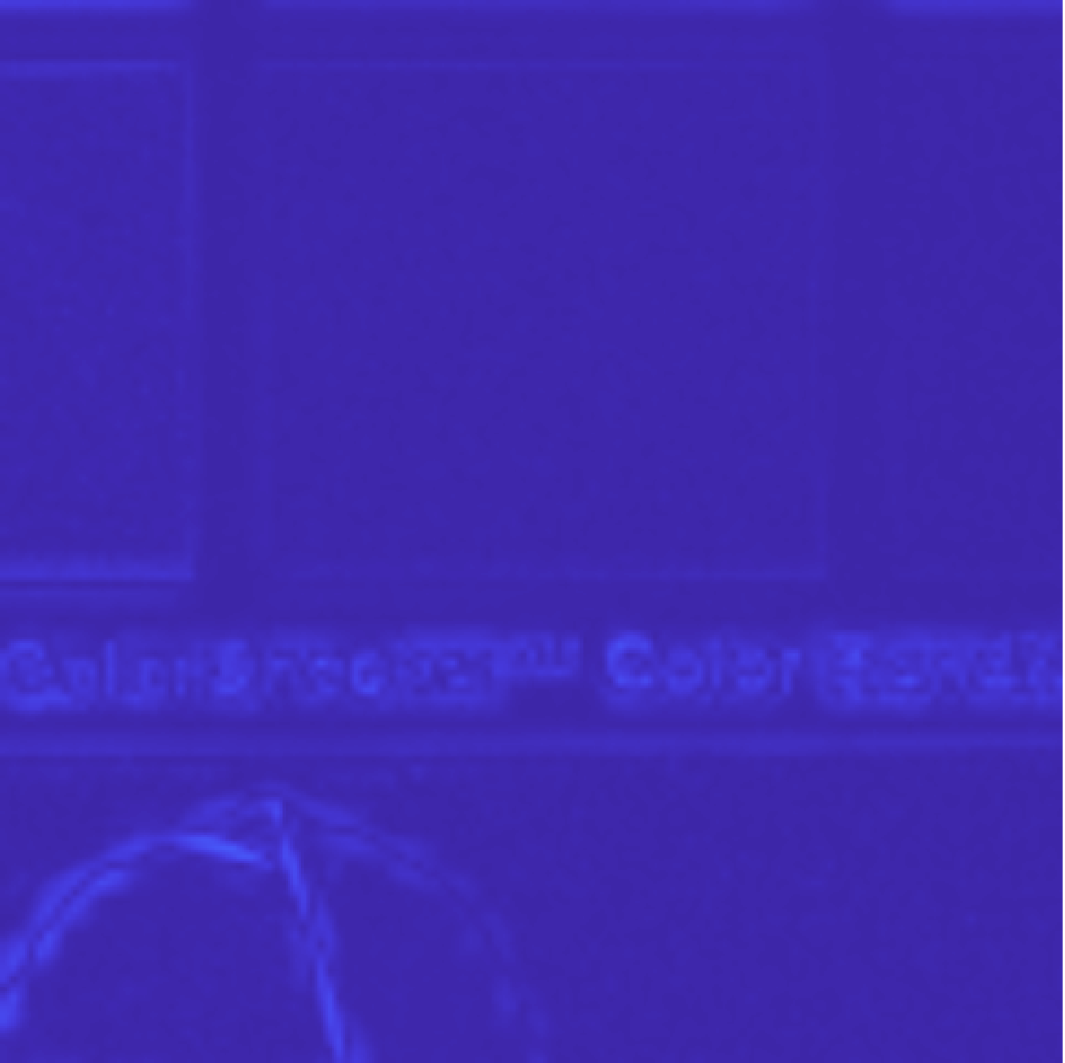}
		\end{minipage}\hspace{-0.3mm}}    
	\subfloat[DBIN]{
		\begin{minipage}[b]{0.095\linewidth}
			\includegraphics[width=0.655in,height=0.657in]{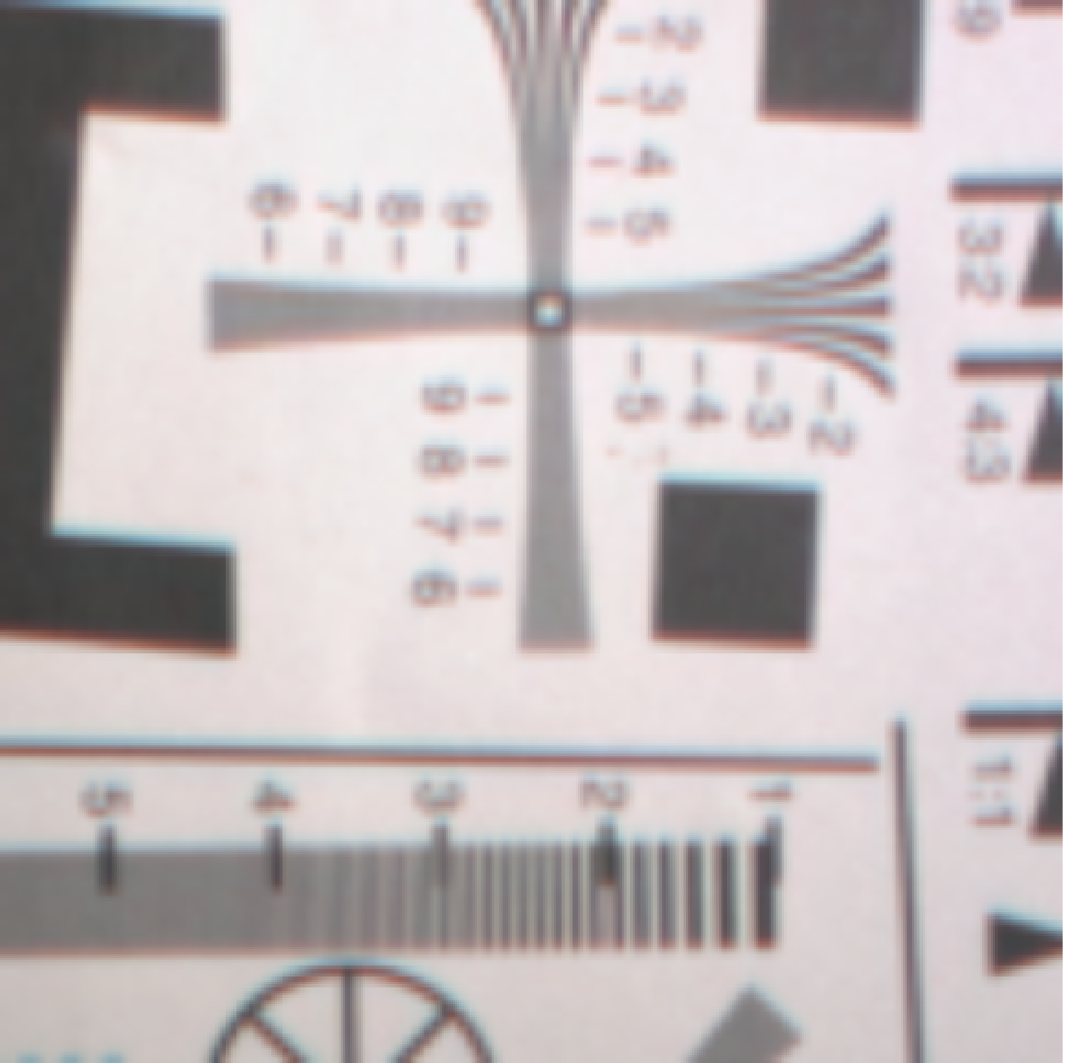}\vspace{0.4mm} \\
			\includegraphics[width=0.655in,height=0.657in]{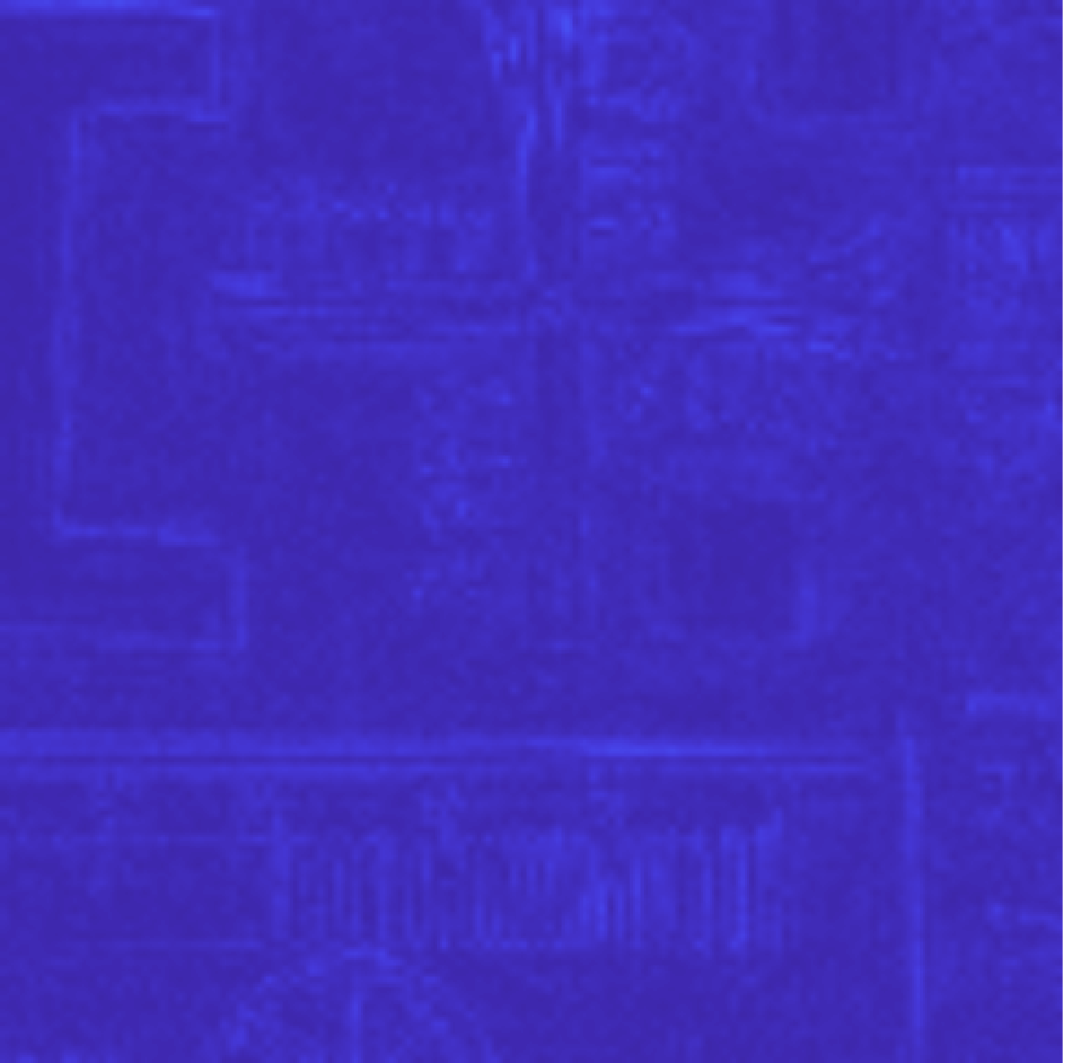}\vspace{0.4mm} \\
			\includegraphics[width=0.655in,height=0.657in]{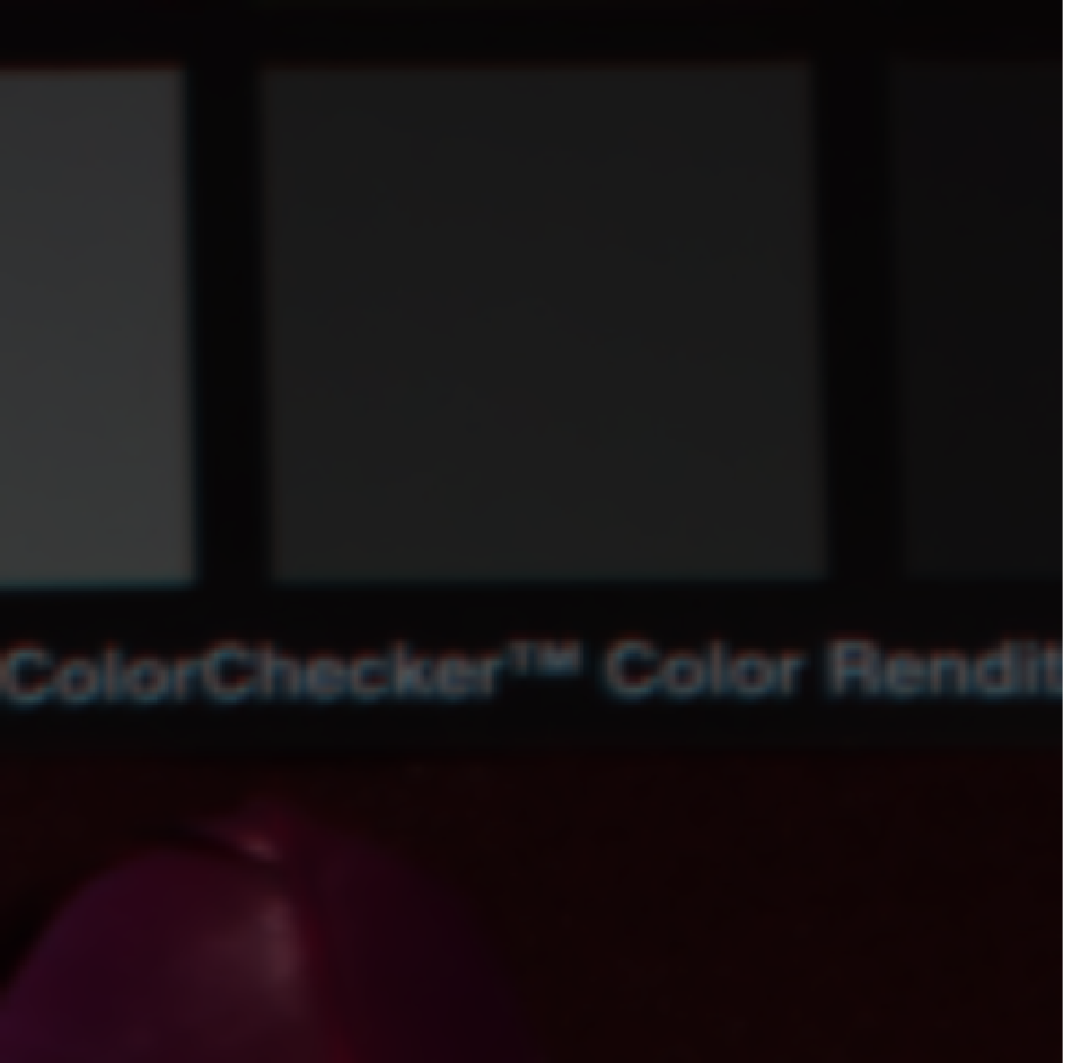}\vspace{0.4mm} \\
			\includegraphics[width=0.655in,height=0.657in]{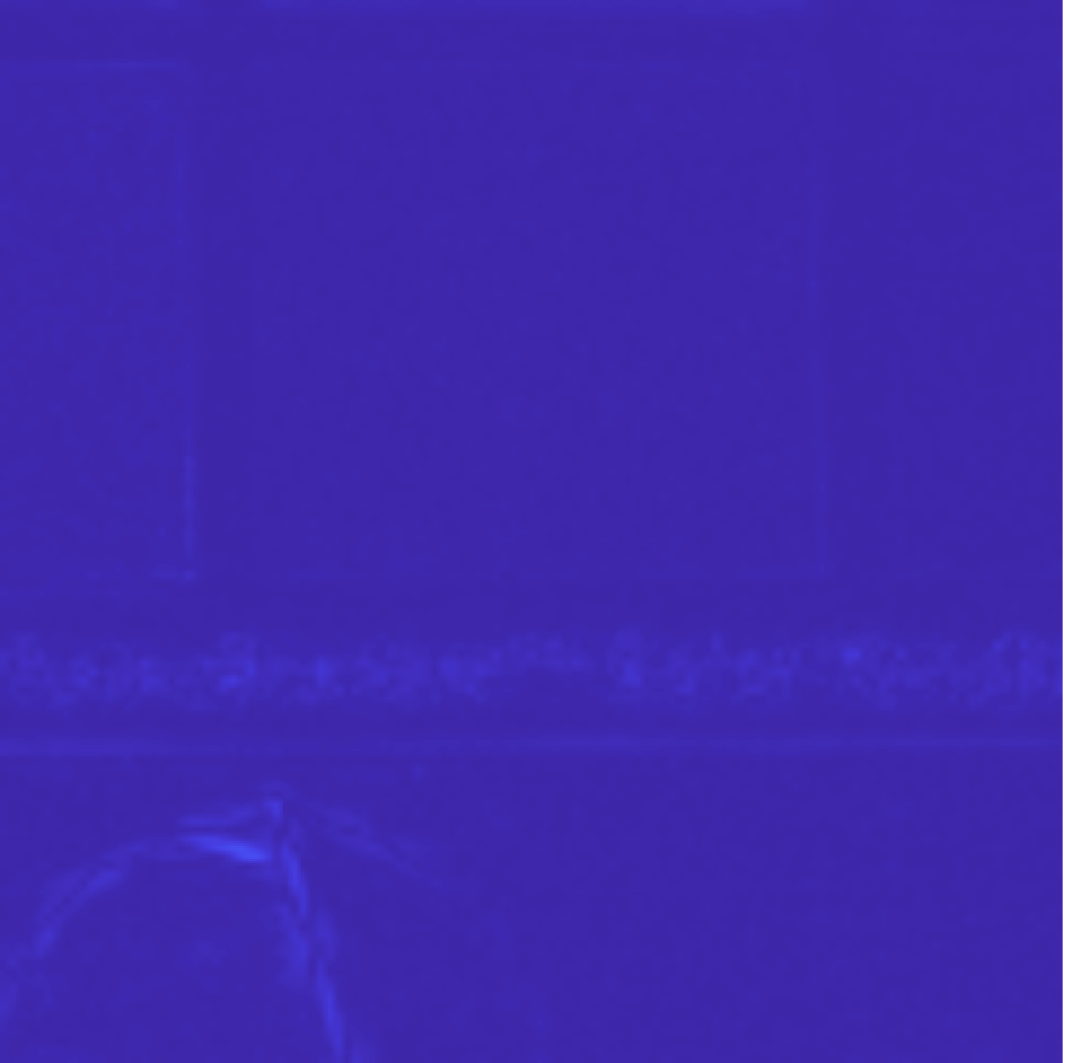}
		\end{minipage}\hspace{-0.3mm}}
	\subfloat[ResTFNet]{
		\begin{minipage}[b]{0.095\linewidth}
			\includegraphics[width=0.655in,height=0.657in]{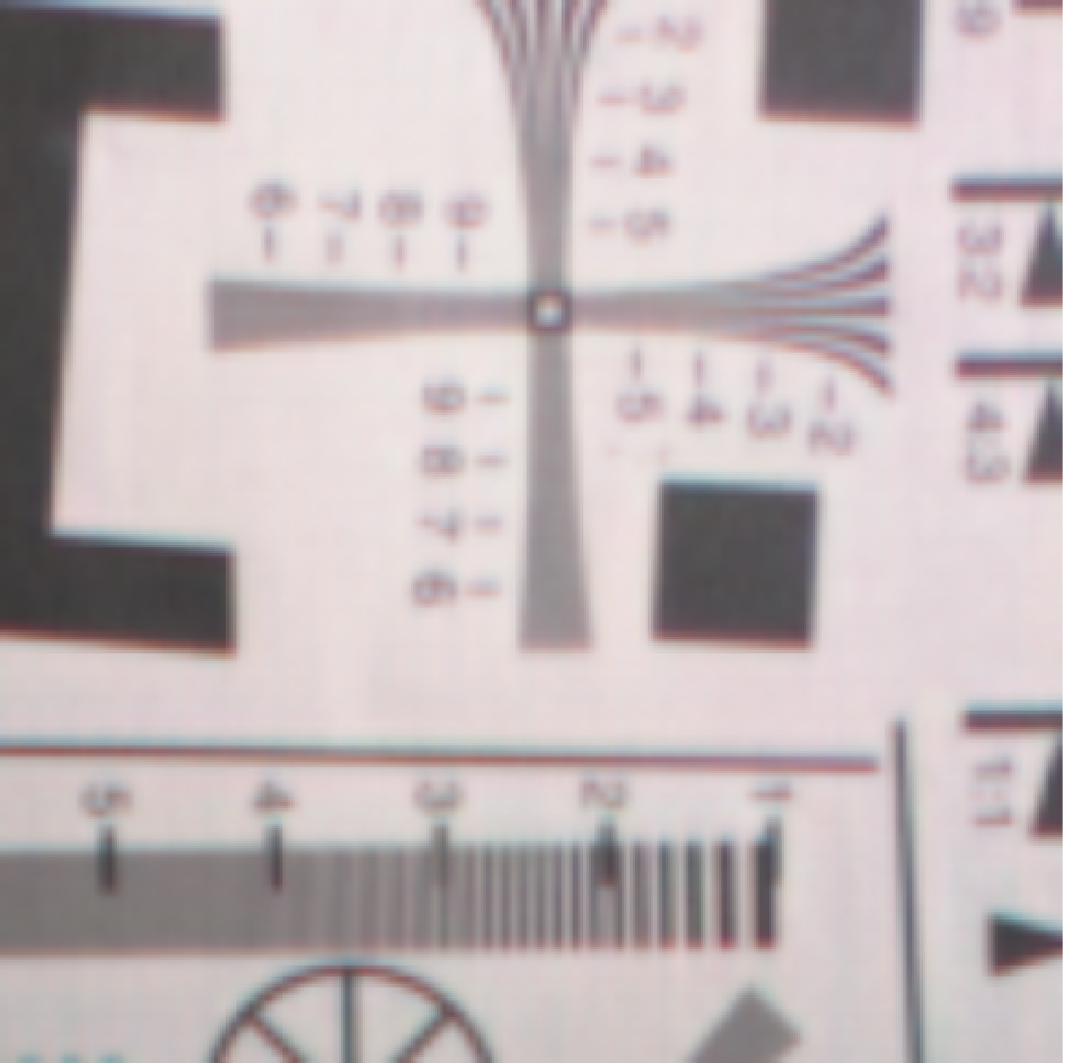}\vspace{0.4mm} \\
			\includegraphics[width=0.655in,height=0.657in]{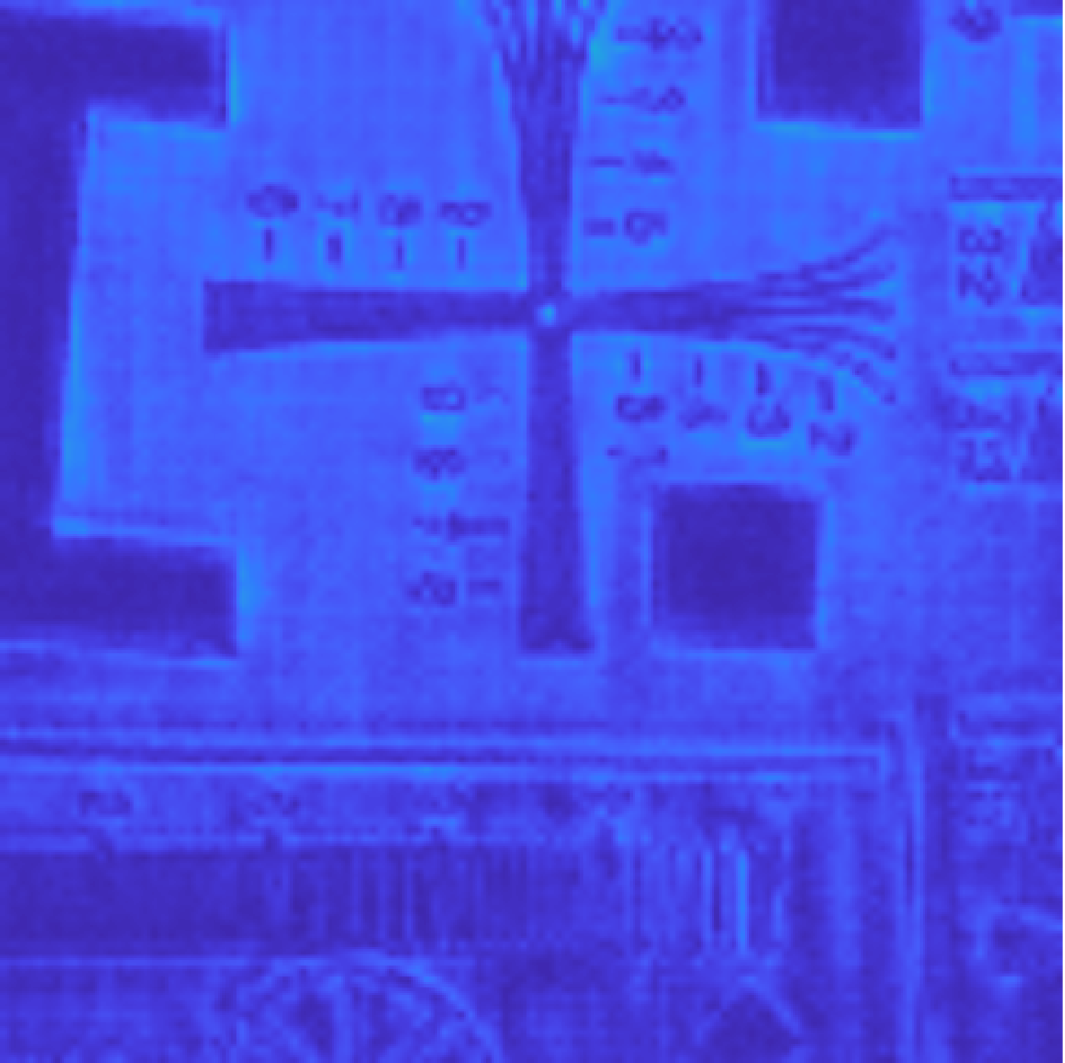}\vspace{0.4mm} \\
			\includegraphics[width=0.655in,height=0.657in]{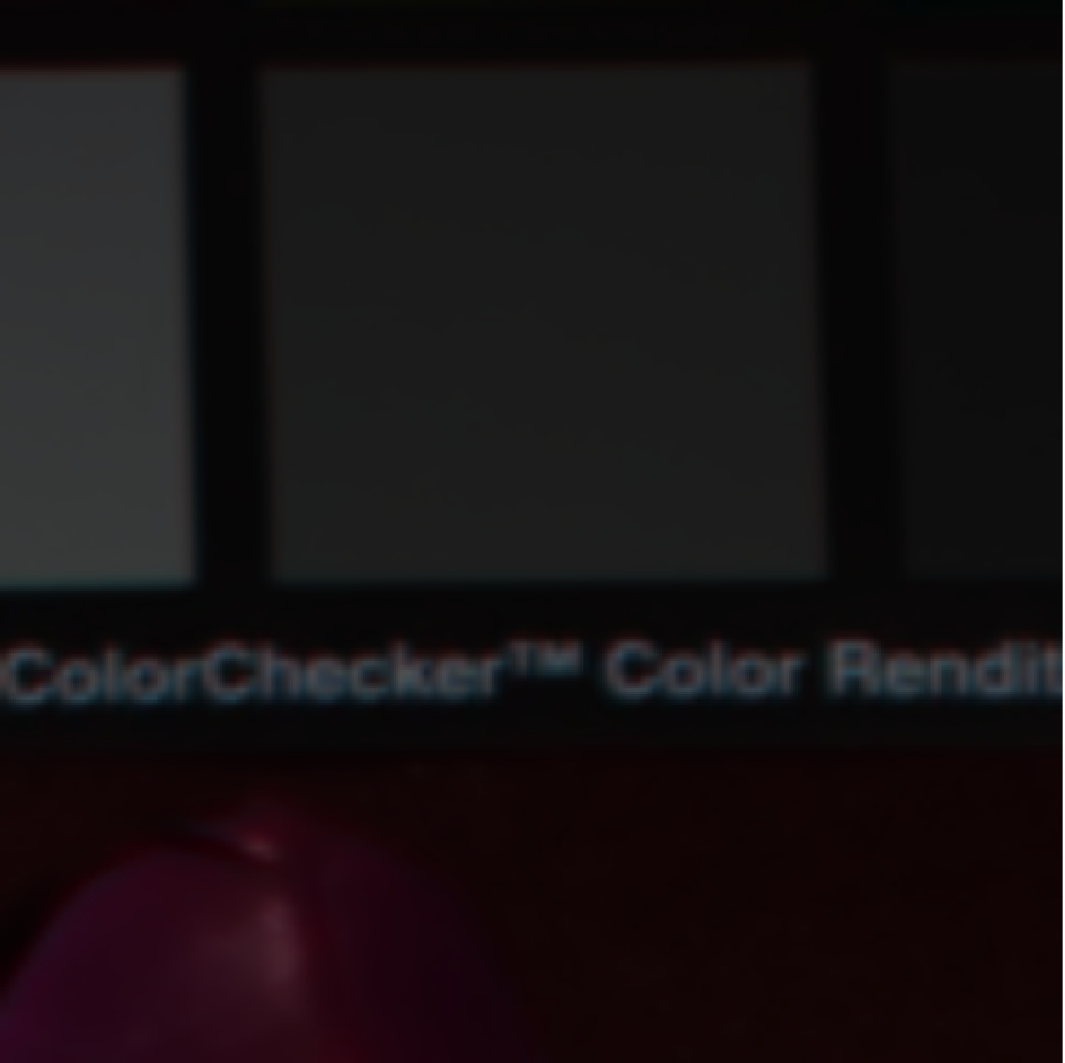}\vspace{0.4mm} \\
			\includegraphics[width=0.655in,height=0.657in]{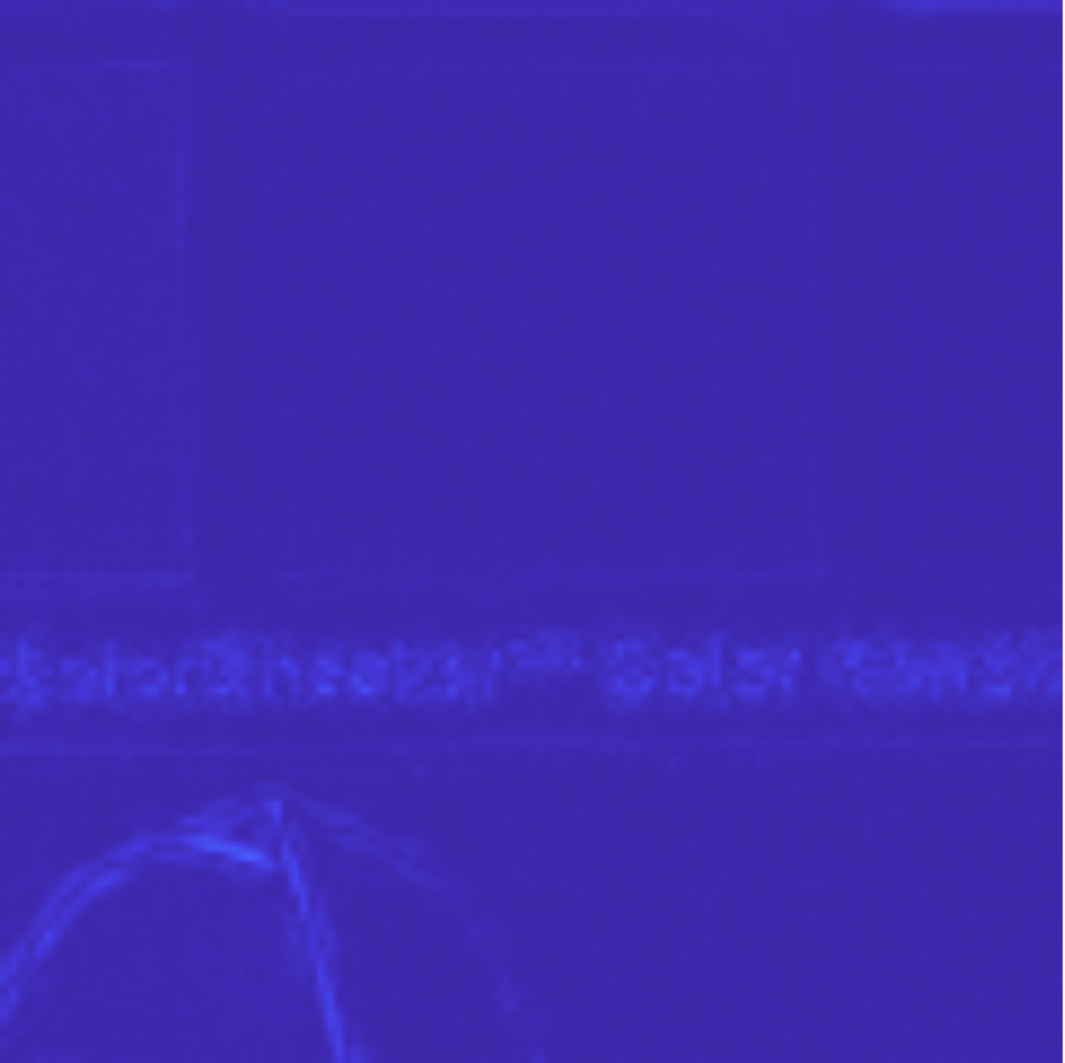}
		\end{minipage}\hspace{-0.3mm}}
	\subfloat{
		\begin{minipage}[b]{0.095\linewidth}
			\includegraphics[width=0.28in,height=2.5in]{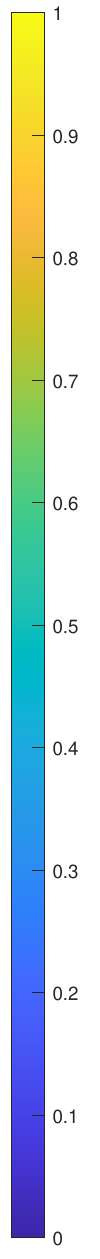}\vspace{0.5mm} \\
	\end{minipage}}
	
	\caption{\footnotesize{The first and third rows show the results using the pseudo-color representation on ``\emph{chart and stuffed toy}'' and ``\emph{feathers}'', respectively, from the CAVE dataset. Some close-ups are depicted in the blue rectangles. The second and fourth rows show the residuals between the GT and the fused products. (a) GT, (b) Ours, (c) DHIF~\cite{huang2022deep}, (d) Fusformer~\cite{hu2022fusformer}, (e) MoG-DCN~\cite{dong2021model}, (f) HSRNet~\cite{hu2021hyperspectral}, (g) SSRNet~\cite{zhang2020ssr}, (h) DBIN~\cite{Wang_2019_ICCV} and (i) ResTFNet~\cite{LIU20201}. }}
	\label{cave_comparison}
\end{figure*}
\subsection{Implicit Neural Feature Fusion Function}
\label{section:3.3}
The objective of the MHIF task is to fuse the different modal inputs of LR-HSI and HR-MSI, resulting in the generation of HR-HSI with high spectral and spatial resolution. Previous fusion techniques usually construct two separate CNN branches for the LR-HSI and HR-MSI~\cite{zhang2020ssr}, and then extract sets of CNN features~\cite{LIU20201,dong2021model,huang2022deep}. However, the CNN-based fusion methods are significantly dependent on stacking convolution structures and lack interpretability. Inspired by the recent developments in INR~\cite{chen2021learning,tang2021joint}, we propose a multimodal and multiscale fusion function based on the INR framework, and use parameter-free weight generation method to facilitate the mining of high-frequency information in the fusion process. In summary, we innovatively create the Implicit Neural Feature Fusion Function (INF³) to guide the fusion process. Unlike the LIIF~\cite{chen2021learning} representation, which directly generates the predicted signal, our INF³ is designed to generate the fused feature map $\mathcal{E} \in \mathbb{R}^{H\times W\times C}$, and then use a decoder structure to work out our final output $\tilde{\mathcal{X}}$. Specifically, the fused feature map $\mathcal{E}$ at position $C_q$ can be represented as follows:
\begin{equation}
	\mathcal{E}_q=\sum_{i\in\mathcal{N}_q}w_{q,i}\mathcal{F}_{q,i},\label{rqa}
\end{equation}
where $\mathcal{N}_q$ indicates the set of the four nearest query coordinates around $C_q$ in the normalized HR domain, $w_{q,i}$ and $\mathcal{F}_{q,i}$ are the weights and multimodal fusion information of query coordinate $C_q$ at position $C_i$, respectively. Typically, we regard $\mathcal{F}_{q,i}\in\mathbb{R}^{1\times 1\times C}$ as the fused feature vector at position $C_i$ when querying coordinate $C_q$. In the following passage, we will introduce how to generate $\mathcal{F}_{q,i}$ and $w_{q,i}$. 

\begin{figure}[h]
	\centering
	\includegraphics[width=8.4cm, height=3.5cm]{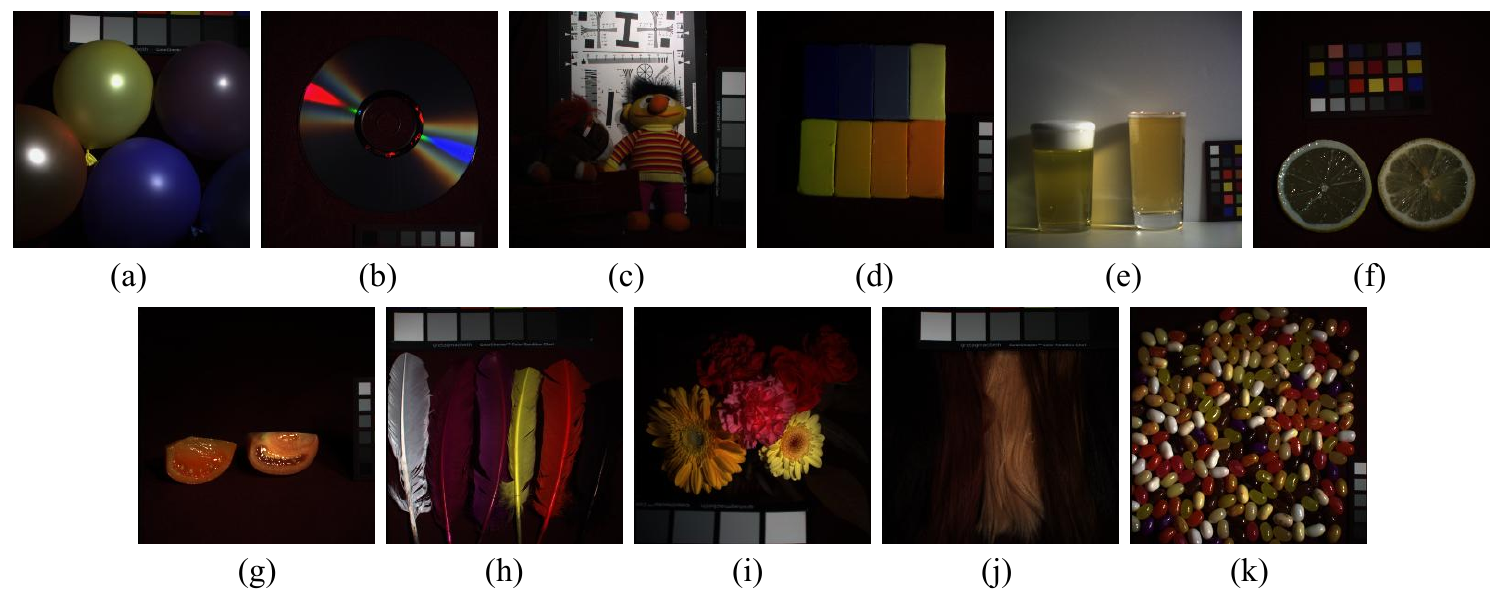}
	\caption{The testing images from the CAVE dataset: (a) \emph{balloons}, (b) \emph{cd}, (c) \emph{chart and stuffed toy}, (d) \emph{clay}, (e) \emph{fake and real beers}, (f) \emph{fake and real lemon slices}, (g) \emph{fake and real tomatoes}, (h) \emph{feathers}, (i) \emph{flowers}, (j) \emph{hairs}, and (k) \emph{jelly beans}. An RGB color representation is used to depict the images.}
	\label{fig_CAVE}
	\vspace{-5mm}
\end{figure}

\noindent$\textbf{Dual high-frequency fusion:}$
\label{section:3.3.1}
We observe that different resolution information plays an important role in MHIF tasks. To this end, we design a structure, \emph{i.e.,} dual high-frequency fusion (DHFF), which combines high-frequency spatial information at different resolutions. Firstly, we concatenate the LR domain spatial information $\mathcal{S}_{pa}^D \in \mathbb{R}^{h\times w\times D_2}$ and spectral information $\mathcal{S}_{pe}\in \mathbb{R}^{h\times w\times D_1}$ vectors at position $C_i$, which can be formulated as follows: 
\begin{equation}
	\mathcal{F}_{q,i}^1=\mathrm{Concat}(\mathcal{S}_{pe}(C_i),\mathcal{S}_{pa}^D(C_i)), 
	\label{fusion_1}
\end{equation}
where $\mathcal{F}_{q,i}^1 \in \mathbb{R}^{1\times 1\times (D_1+D_2)}$ is the fusion of spectral information and spatial information at the same resolution. Specifically, we generate LR domain high-frequency spatial information, \emph{i.e.,} $\mathcal{S}_{pa}^D$ by the following formula:
\begin{equation}
	\mathcal{S}_{pa}^D=\mathrm{Mean}(\mathcal{S}_{pa}).
	\label{mean}
\end{equation}
Given the up-sampling ratio is $r$, we operate mean operation on the $r \times r$ region of $\mathcal{S}_{pa} \in \mathbb{R}^{H\times W\times D_2}$ and then get the $\mathcal{S}_{pa}^D \in \mathbb{R}^{h\times w\times D_2}$. This design is to incorporate the LR domain high-frequency information, to better serve the MLP layer. In Sec.~\ref{section4.1}, we will design relevant ablation study and verify the effectiveness of this design. Secondly, we combine HR domain information $\mathcal{S}_{pa}\in\mathbb{R}^{H\times W\times D_2}$ at position $C_q$ with LR domain fusion information $\mathcal{F}_{q,i}^1$. The process above can be expressed as:
\begin{equation}
	\mathcal{F}_{q,i}^2=\mathrm{Concat}(\mathcal{F}_{q,i}^1,\mathcal{S}_{pa}(C_q)),
	\label{f_ab_1}
\end{equation}
where $\mathcal{F}_{q,i}^2 \in \mathbb{R}^{1\times 1\times (D_1+D_2+D_2)}$ serves as the result of DHFF. In general, our DHFF naturally combines feature vectors of different modalities and different scales with the aim of making the MLP acquire both spatial and spectral information. Similar to the previous INR-based work, we obtain the coordinate modal information by adding the relative positions of $C_q$ and $C_i$ to the fusion process, which can be represented as:
\begin{equation}
	\mathcal{F}_{q,i}^3=\mathrm{Concat}(\mathcal{F}_{q,i}^2,C_q-C_i),
\end{equation}
where $\mathcal{F}_{q,i}^3 \in \mathbb{R}^{1\times 1\times (D_1+D_2+D_2+2)}$ is the result after adding the relative distance information of interpolation on the basis of $\mathcal{F}_{q,i}^2$. Finally, we utilize an MLP layer to learn the information in ${F}_{q,i}^3$ and get the following expression:
\begin{equation}
	\mathcal{F}_{q,i}=\mathrm{MLP}_{\Theta}(\mathcal{F}_{q,i}^3),
\end{equation}
where $\mathcal{F}_{q,i} \in \mathbb{R}^{1\times 1\times C}$ is multimodal fusion information of query coordinate $C_i$ when querying coordinate $C_q$, $\mathrm{MLP}_{\Theta}(\cdot)$ is a fully connected layer, and $\Theta$ serves as its learnable parameters.
\begin{table*}[!t] 
	\setlength{\tabcolsep}{2pt}
	\renewcommand\arraystretch{1.05}
	\centering
	\label{table_reduced}
	\caption{Average quantitative comparisons on 11 CAVE examples and 10 Harvard examples simulating a scaling factor of 4. The best values are highlighted in red, and the second best values are signed in blue. M refers to millions.}
	\resizebox{\linewidth}{!}{
		\begin{tabular}{@{}ccccccccccc@{}}
			\toprule
			\multirow{2}{*}{Methods} & \multicolumn{5}{c}{CAVE} & \multicolumn{5}{c}{Harvard} \\ \cmidrule(l){2-11} 
			&PSNR &SAM &ERGAS &SSIM &$\#$params &PSNR &SAM &ERGAS &SSIM &$\#$params  \\ \midrule
			
			Bicubic                  &34.33$\pm$3.88 &4.45$\pm$1.62 &7.21$\pm$4.90 &0.944$\pm$0.0291 & \multicolumn{1}{c|}{\makecell[c]{$-$}} &38.71$\pm$4.33 &2.53$\pm$0.67 &4.45$\pm$41.81 &0.948$\pm$0.0268 &\makecell[c]{$-$} \\
			
			MTF-GLP-HS~\cite{selva2015hyper}&37.69$\pm$3.85 &5.33$\pm$1.91 &4.57$\pm$2.66 &0.973$\pm$0.0158 &\multicolumn{1}{c|}{\makecell[c]{$-$}}  &33.81$\pm$3.50&6.25$\pm$2.42&3.47$\pm$1.82&0.952$\pm$0.0321&  \makecell[c]{$-$} \\
			
			CSTF-FUS~\cite{li2018fusing}&34.46$\pm$4.28 &14.37$\pm$5.30 &8.29$\pm$5.29 &0.866$\pm$0.0747&\multicolumn{1}{c|}{\makecell[c]{$-$}}
			&39.13$\pm$3.50&6.91$\pm$2.66&4.64$\pm$1.80&0.913$\pm$0.0487&\makecell[c]{$-$} \\
			
			LTTR~\cite{dian2019learning}&35.85$\pm$3.49 &6.99$\pm$2.55 &5.99$\pm$2.92 &0.956$\pm$0.0288&\multicolumn{1}{c|}{\makecell[c]{$-$}}
			&37.91$\pm$3.58&5.35$\pm$1.94&2.44$\pm$1.06&0.972$\pm$0.0183&\makecell[c]{$-$} \\
			
			LTMR~\cite{dian2019hyperspectral}&36.54$\pm$3.30 &6.71$\pm$2.19 &5.39$\pm$2.53&0.963$\pm$0.0208&\multicolumn{1}{c|}{\makecell[c]{$-$}}
			&38.41$\pm$3.58&5.05$\pm$1.70&2.24$\pm$0.97&0.970$\pm$0.0166&\makecell[c]{$-$} \\
			
			IR-TenSR~\cite{xutgrs2022}&35.61$\pm$3.45 &12.30$\pm$4.68 &5.90$\pm$3.05 &0.945$\pm$0.0267 &\multicolumn{1}{c|}{\makecell[c]{$-$}}
			&40.47$\pm$3.04&4.36$\pm$1.52&5.57$\pm$1.57&0.962$\pm$0.0140 &\makecell[c]{$-$} \\
			
			\hline
			DBIN~\cite{Wang_2019_ICCV}&50.83$\pm$4.29 &2.21$\pm$0.63 &1.24$\pm$1.06 &0.996$\pm$0.0026 &\multicolumn{1}{c|}{\makecell[c]{\textcolor{blue}{\textbf{0.469M}}}} &47.88$\pm$3.87 &2.31$\pm$0.46 &1.95$\pm$0.81 &0.988$\pm$0.0066 &0.469M\\
			
			ResTFNet~\cite{LIU20201}&45.58$\pm$5.47 &2.82$\pm$0.70 &2.36$\pm$2.59 &0.993$\pm$0.0056 &
			\multicolumn{1}{c|}{\makecell[c]{2.387M}} &45.93$\pm$4.35 &2.61$\pm$0.69 &2.56$\pm$1.32 &0.985$\pm$0.0082 &2.387M\\
			
			SSRNet~\cite{zhang2020ssr}&48.62$\pm$3.92 &2.54$\pm$0.84 &1.63$\pm$1.21 &0.995$\pm$0.0023&
			\multicolumn{1}{c|}{\makecell[c]{\textcolor{red}{\textbf{0.027M}}}}  &47.95$\pm$3.37 &2.31$\pm$0.60 &2.30$\pm$1.42 &0.987$\pm$0.0070 &\textcolor{red}{\textbf{0.027M}}\\
			
			HSRNet~\cite{hu2021hyperspectral}&50.38$\pm$3.38 &2.23$\pm$0.66 &1.20$\pm$0.75 &0.996$\pm$0.0014 &
			\multicolumn{1}{c|}{\makecell[c]{0.633M}} &\textcolor{blue}{\textbf{48.29$\pm$3.03}} &2.26$\pm$0.56 &\textcolor{blue}{\textbf{1.87$\pm$0.81}} &\textcolor{blue}{\textbf{0.988$\pm$0.0064}} &0.633M\\
			
			MoG-DCN~\cite{dong2021model} & \textcolor{blue}{\textbf{51.63$\pm$4.10}} &2.03$\pm$0.62 &\textcolor{blue}{\textbf{1.11$\pm$0.82}} & 0.997$\pm$0.0018 &
			\multicolumn{1}{c|}{\makecell[c]{6.840M}}  &47.89$\pm$4.09 &\textcolor{red}{\textbf{2.11$\pm$0.52}} &1.89$\pm$0.82 & 0.988$\pm$0.0073 &6.840M\\
			
			Fusformer~\cite{hu2022fusformer}&49.98$\pm$8.10 &2.20$\pm$0.85 &2.50$\pm$5.21 &0.994$\pm$0.0111 &\multicolumn{1}{c|}{\makecell[c]{0.504M}}  &47.87$\pm$5.13 &2.84$\pm$2.07 &2.04$\pm$0.99 &0.986$\pm$0.0101 &\textcolor{blue}{\textbf{0.467M}}\\
			
			DHIF~\cite{huang2022deep} & 51.07$\pm$4.17 &\textcolor{blue}{\textbf{2.01$\pm$0.63}} & 1.22$\pm$0.97& \textcolor{blue}{\textbf{0.997$\pm$0.0016}} &\multicolumn{1}{c|}{\makecell[c]{22.462M}}   & 47.68$\pm$3.85 &2.32$\pm$0.53 & 1.95$\pm$0.92& 0.988$\pm$0.0074 &22.462M\\
			
			INF³ (ours) &\textcolor{red}{\textbf{52.36$\pm$3.93}} &\textcolor{red}{\textbf{1.99$\pm$0.60}} &\textcolor{red}{\textbf{0.99$\pm$0.73}} 
			&\textcolor{red}{\textbf{0.997$\pm$0.0013}} &\multicolumn{1}{c|}{\makecell[c]{2.902 M}}  &\textcolor{red}{\textbf{48.46$\pm$3.43}} &\textcolor{blue}{\textbf{2.14$\pm$0.52}} &\textcolor{red}{\textbf{1.83$\pm$0.76}} &\textcolor{red}{\textbf{0.989$\pm$0.0064}} &2.902 M\\ 
			\midrule
			\textbf{Ideal value} &$\mathbf{\infty}$ &\textbf{0} &\textbf{0} &\textbf{1} & -  &$\mathbf{\infty}$ &\textbf{0} &\textbf{0} &\textbf{1} & -\\ 
			\bottomrule				
	\end{tabular}}
	\label{table_reduced}
\end{table*}

\noindent$\textbf{Cosine similarity:}$ The proposed INR with cosine similarity (INR-CS) method generates weights based on cosine similarity. In Eq.~\eqref{rqa}, $w_{q,i}\in\mathbb{R}$ is the weight at the position $C_i$ when querying the coordinate $C_i$. Part of the previous work viewed the generation of this weight simply as a solution to the interpolation problem, using area-based method to generate the target weights~\cite{chen2021learning}, which ignores local texture and information about the data itself. The other part of the work proposes to learn the weights by network parameters, \emph{i.e.,} learning similar weights by graph attention mechanisms~\cite{tang2021joint} which lacks of interpretability. In order to utilize information about $\mathcal{F}_{q,i}$ while keeping interpretability, we propose a parameter-free approach named INR-CS as follows:
\begin{equation}
	w_{q,i}=\frac{\exp(\Vert{\mathcal{F}_{q,\widehat{q}}^1\Vert\cdot\Vert \mathcal{F}_{q,i}^1\Vert
			\langle\mathcal{F}_{q,\widehat{q}}^1, \mathcal{F}_{q,i}^1\rangle})}{w_q}, \\
\end{equation}
where
\begin{equation}
	w_q = \sum_{i\in\mathcal{N}_q}\exp(\Vert{\mathcal{F}_{q,\widehat{q}}^1\Vert\cdot\Vert\mathcal{F}_{q,i}^1\Vert
		\langle\mathcal{F}_{q,\widehat{q}}^1, \mathcal{F}_{q,i}^1\rangle}).
\end{equation}
$\mathcal{F}_{q,i}^1$ is given by Eq.~\eqref{fusion_1}, where the $\widehat{q}$ is the closest point to $C_q$ in the LR domain. In detail, $\mathcal{F}_{q,\widehat{q}}^1=\mathrm{Concat}(\mathcal{S}_{pa}^D(\widehat{q}),\mathcal{S}_{pe}(\widehat{q}))$, and $\langle \cdot,\cdot \rangle$ represents the cosine operation between vectors. The similarity between feature vectors is normalized by the softmax function.

\section{Experiment}
\noindent$\textbf{Datasets:}$
Following the previous studies, we conduct experiments to evaluate our model on the CAVE \footnote{\url{https://www.cs.columbia.edu/CAVE/databases/multispectral/}}and Harvard\footnote{\url{http://vision.seas.harvard.edu/hyperspec/index.html}} datasets. In detail, the CAVE dataset contains 32 HSIs with 31 spectral bands ranging in wavelengths from 400 nm to 700 nm in increments of 10 nm. We randomly select 20 images for training, and the remaining 11 images make up the testing dataset. In addition, the Harvard dataset includes 77 HSIs of both indoor and outdoor scenes, with each HSI having a size of $1392 \times 1040 \times 31$ and spanning the 420 nm to 720 nm spectral range. We chose 20 of them and crop the upper left portion ($1000 \times 1000$), with the 10 images being utilized for testing and the remaining 10 were used for training.

\noindent$\textbf{Data Simulation:}$
We input LR-HSI and HR-MSI $(\mathcal{X, Y})$ pairs into the end-to-end network, and use HR-HSI $\bar{\mathcal{X}}$ for training. Due to ground-truth (GT) $\bar{\mathcal{X}}$ is not available in real life, a simulation process is thus required. As for CAVE dataset, we crop the 20 selected training images to generate 3920 overlapping patches with the dimension $64\times 64\times 31$, and this patches will serve as GT $\bar{\mathcal{X}}$. In order to generate the proper LR-HSIs, we use a $3\times 3$ Gaussian kernel with a standard deviation of 0.5 to blur the initial HR-HSIs and downsample the blurred patches with a scaling factor of 4. Additionally, we utilize the common spectral response function of the Nikon D700\footnote{\url{https://www.maxmax.com/nikon_d700_study.htm}} camera and HR-HSIs to create the HR-MSI patches. Thus, we generate 3920 LR-HSIs with a size of  $16 \times 16 \times 31$ and HR-MSIs with a size of  $64 \times 64 \times 3$ form the input pairs $(\mathcal{X, Y})$. Following that, the inputs pairs and associated GTs are divided at random into training data (80\%) and testing data (20\%). To create the input LR-HSI and HR-MSI products as well as the GTs, this method is also applied to the Harvard dataset.
\begin{figure*}[!h]
	\scriptsize
	\setlength{\tabcolsep}{0.9pt}
	\centering
	\subfloat[GT]{
		\begin{minipage}[b]{0.19\linewidth}
			\includegraphics[width=1.35in,height=1.35in]{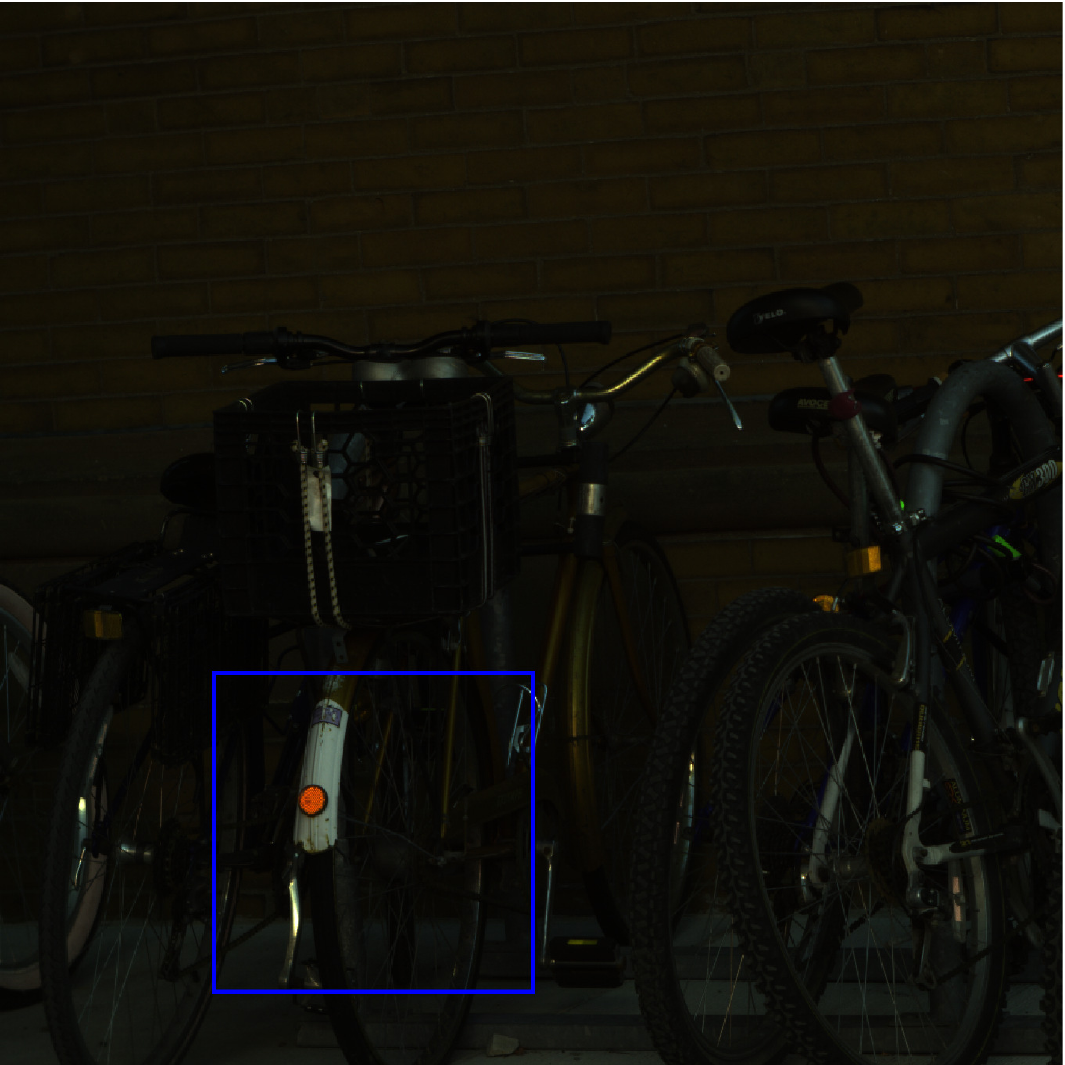}\vspace{0.2mm} \\
			\includegraphics[width=1.35in,height=1.35in]{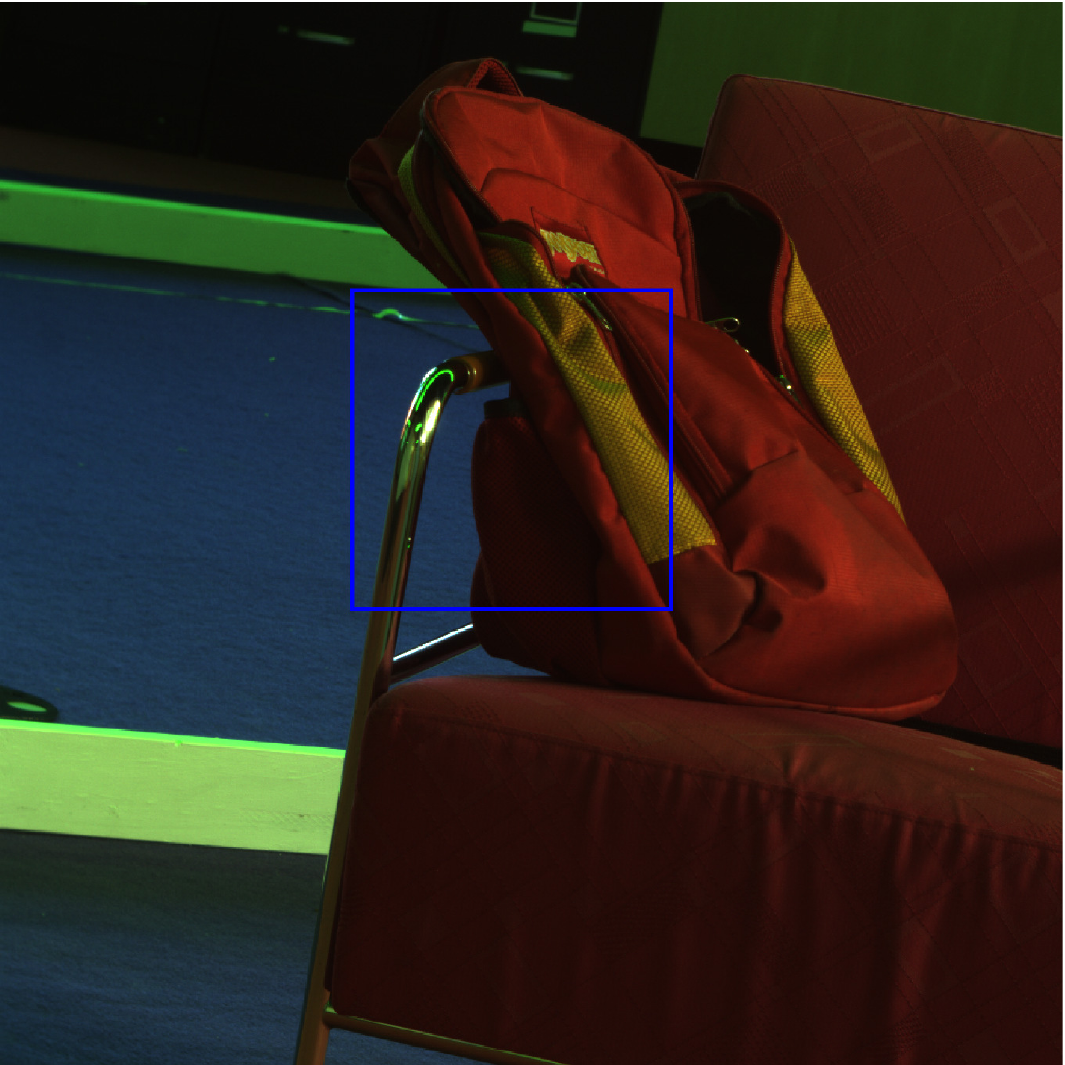}
		\end{minipage}\hspace{0.4mm}}
	\subfloat[Ours]{
		\begin{minipage}[b]{0.095\linewidth}
			\includegraphics[width=0.655in,height=0.657in]{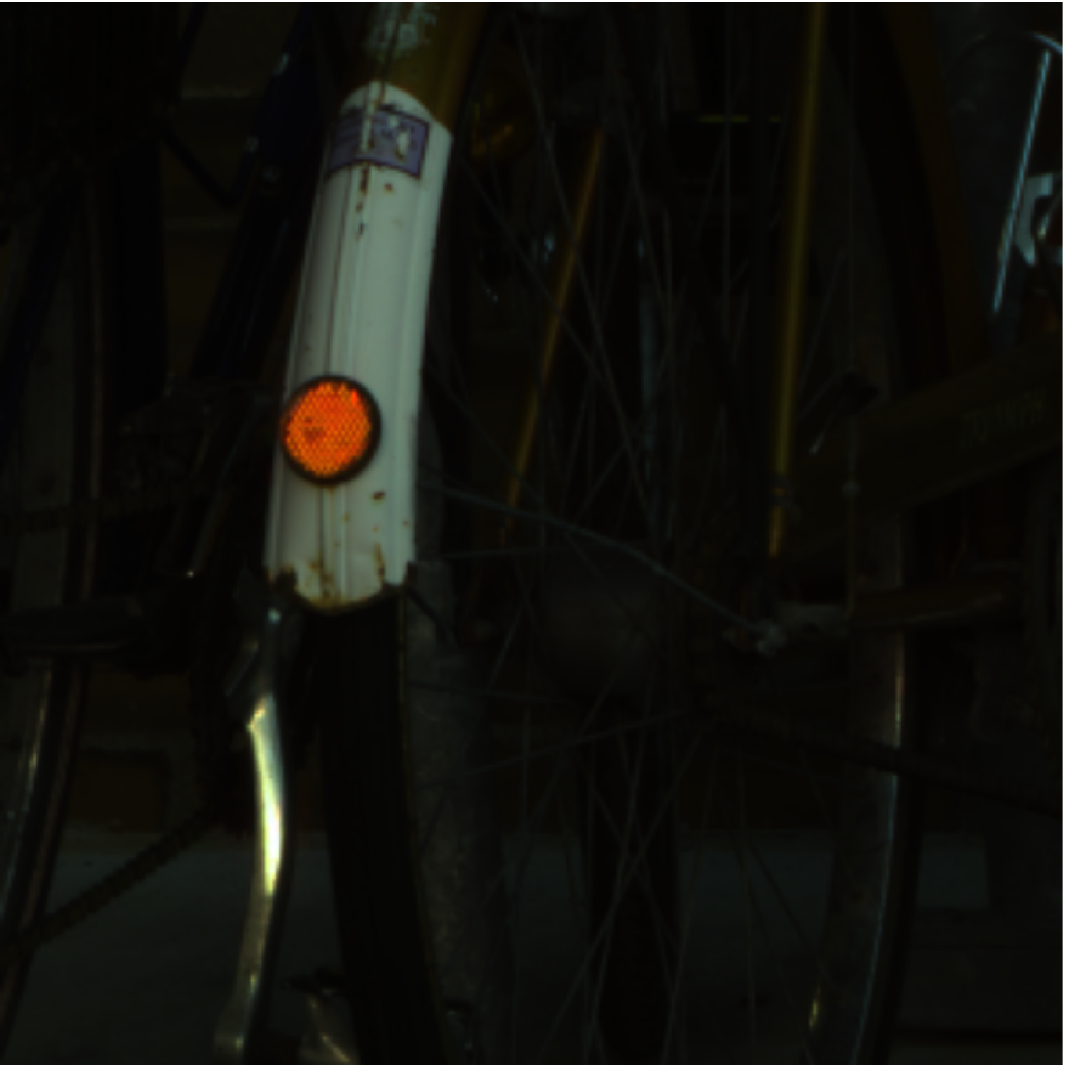}\vspace{0.4mm} \\
			\includegraphics[width=0.655in,height=0.657in]{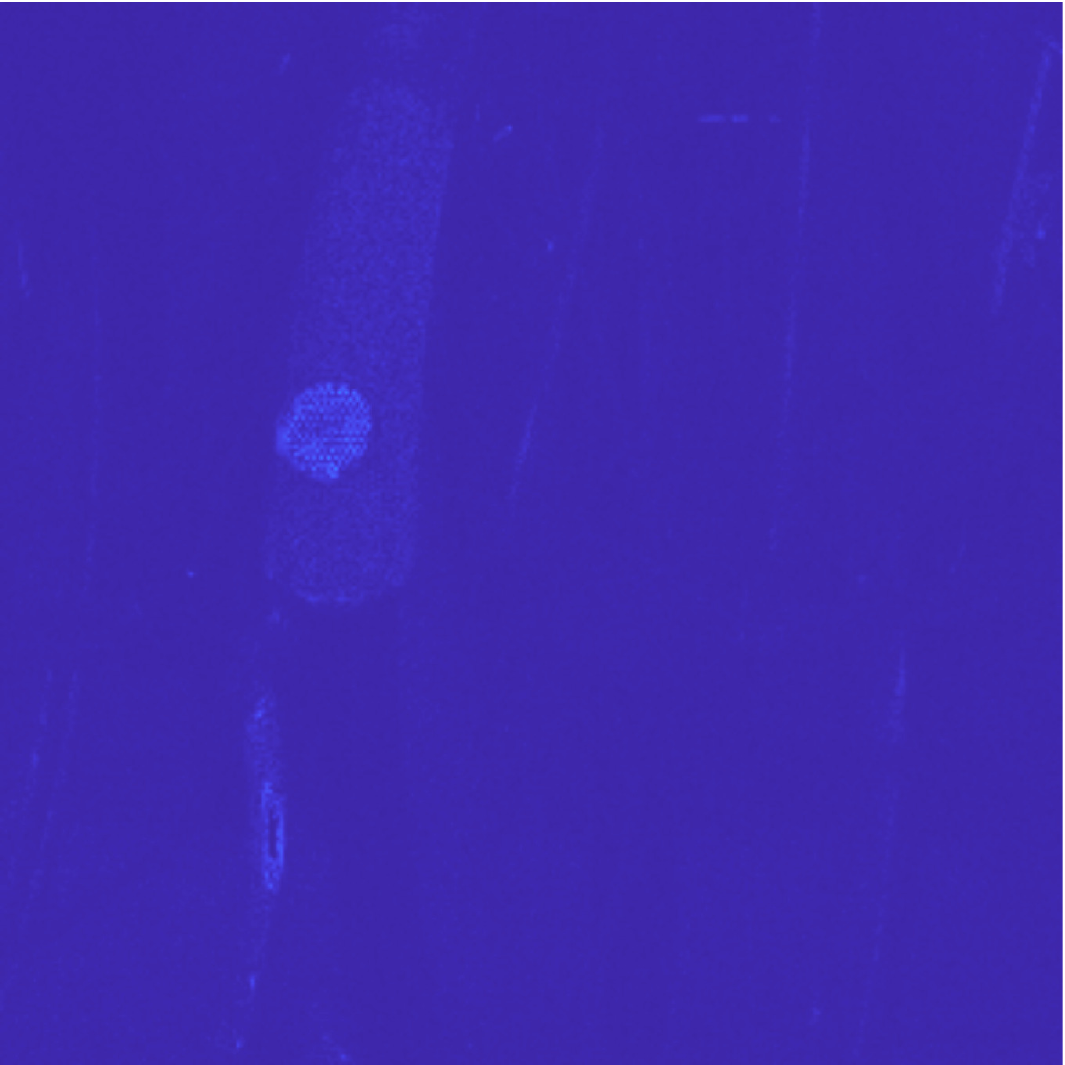}\vspace{0.4mm} \\
			\includegraphics[width=0.655in,height=0.657in]{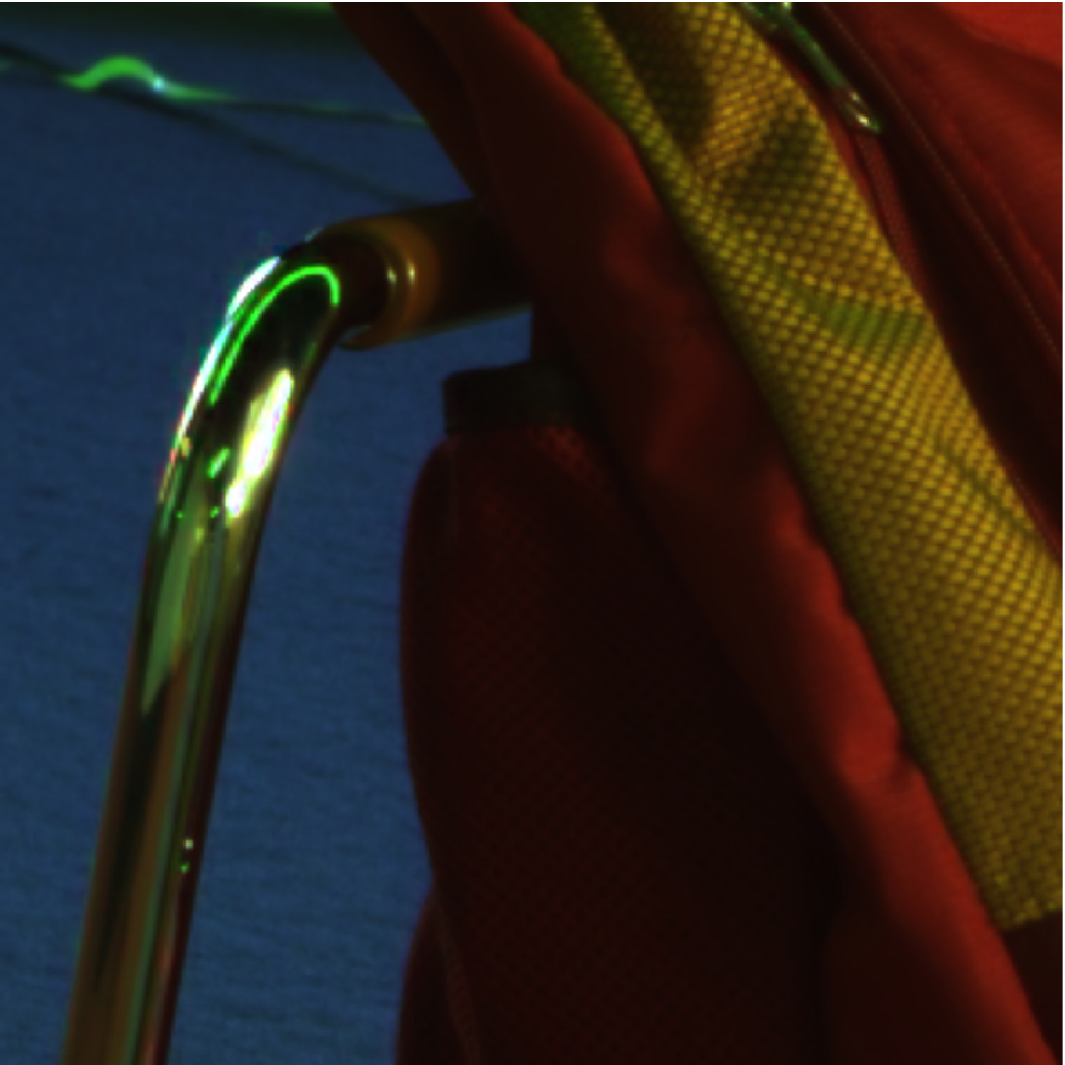}\vspace{0.4mm} \\
			\includegraphics[width=0.655in,height=0.657in]{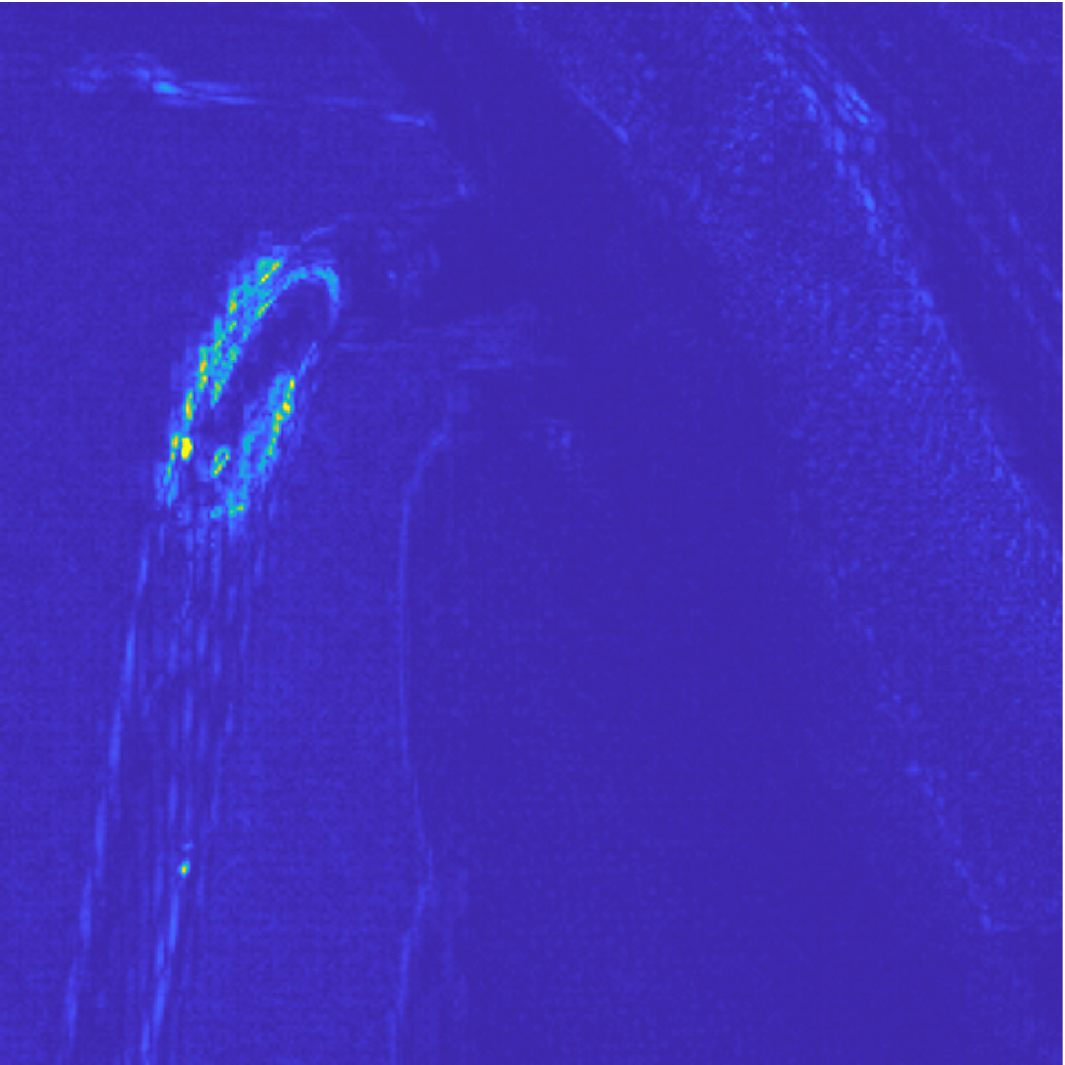} 
		\end{minipage}\hspace{-0.3mm}}
	\subfloat[DHIF]{
		\begin{minipage}[b]{0.095\linewidth}
			\includegraphics[width=0.655in,height=0.657in]{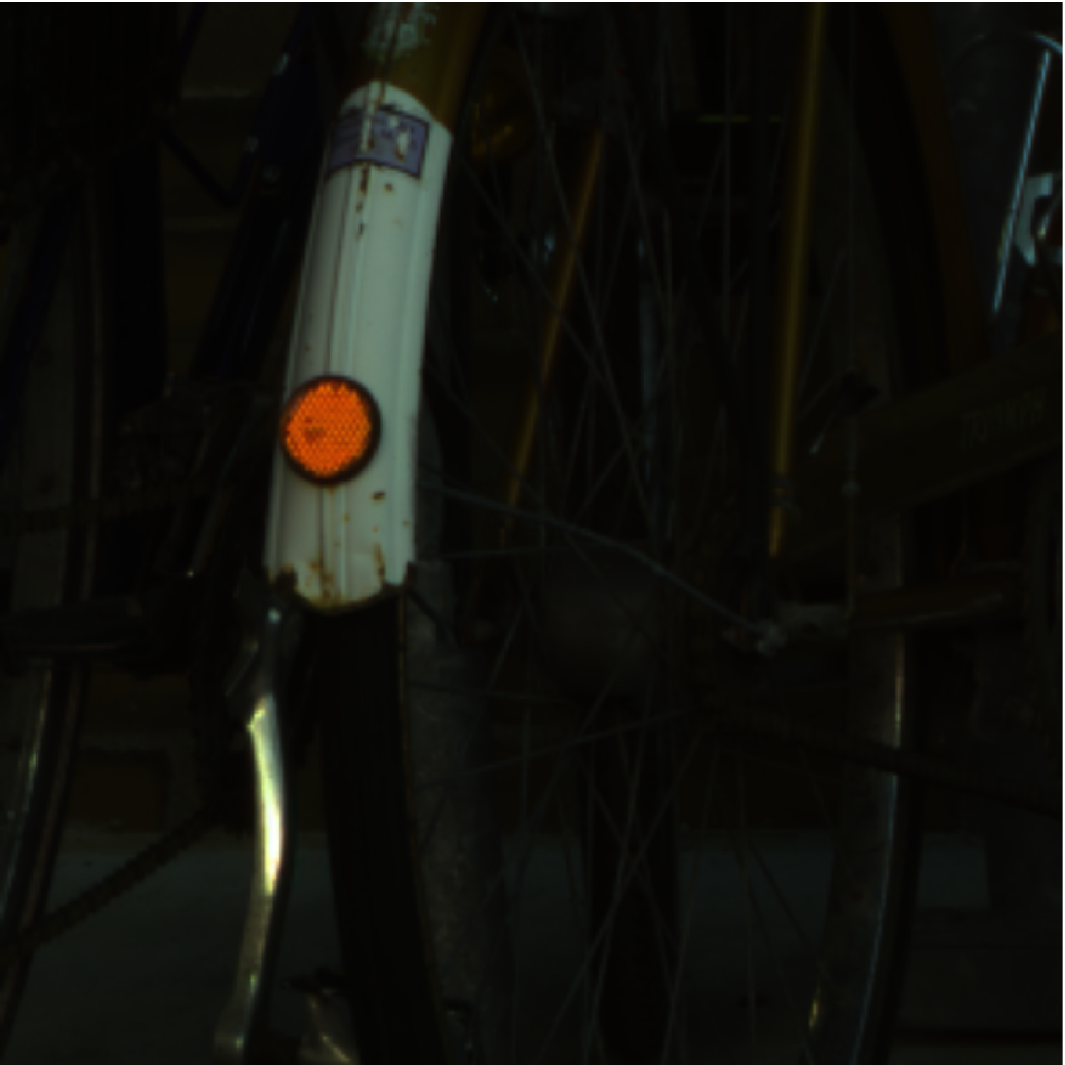}\vspace{0.4mm} \\
			\includegraphics[width=0.655in,height=0.657in]{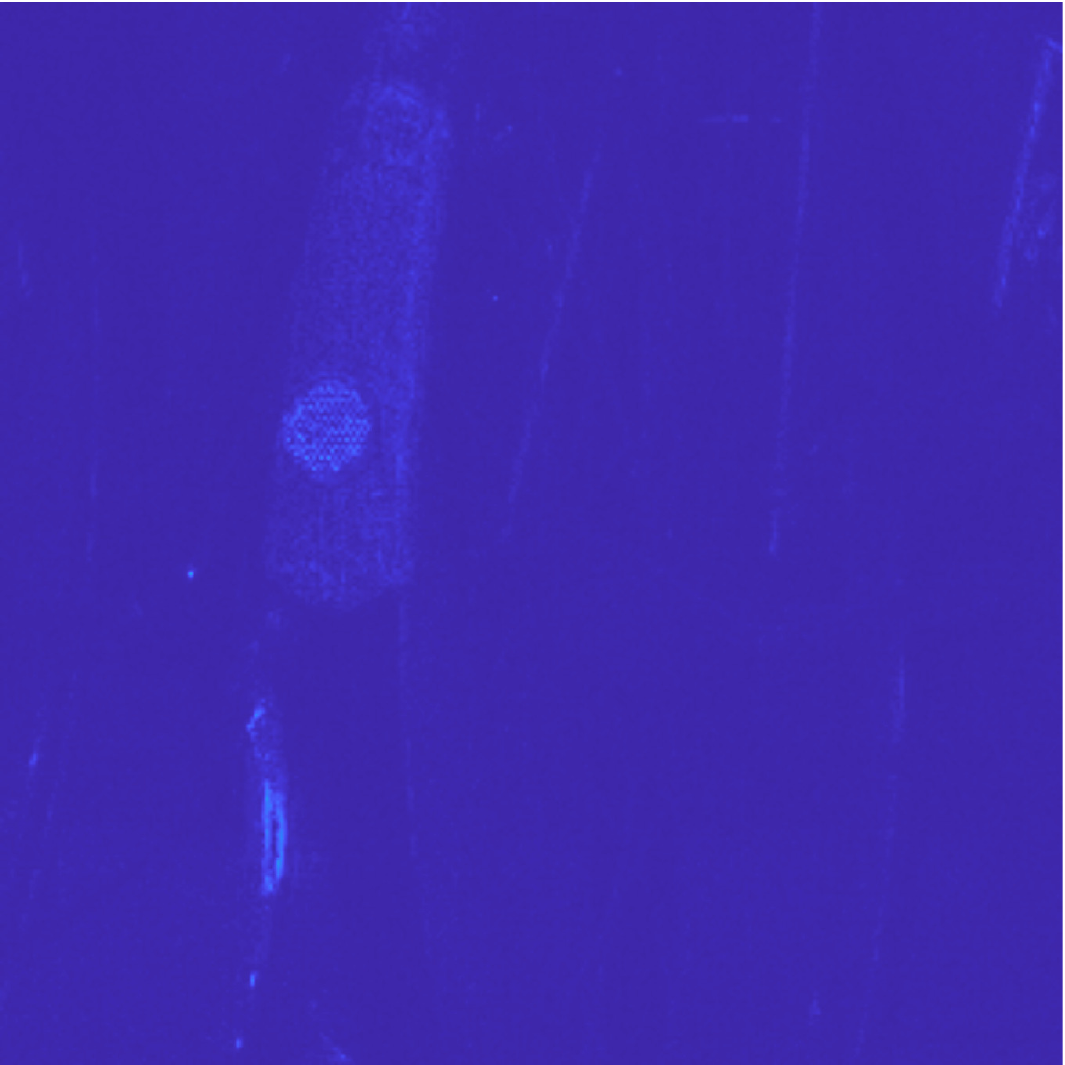}\vspace{0.4mm} \\
			\includegraphics[width=0.655in,height=0.657in]{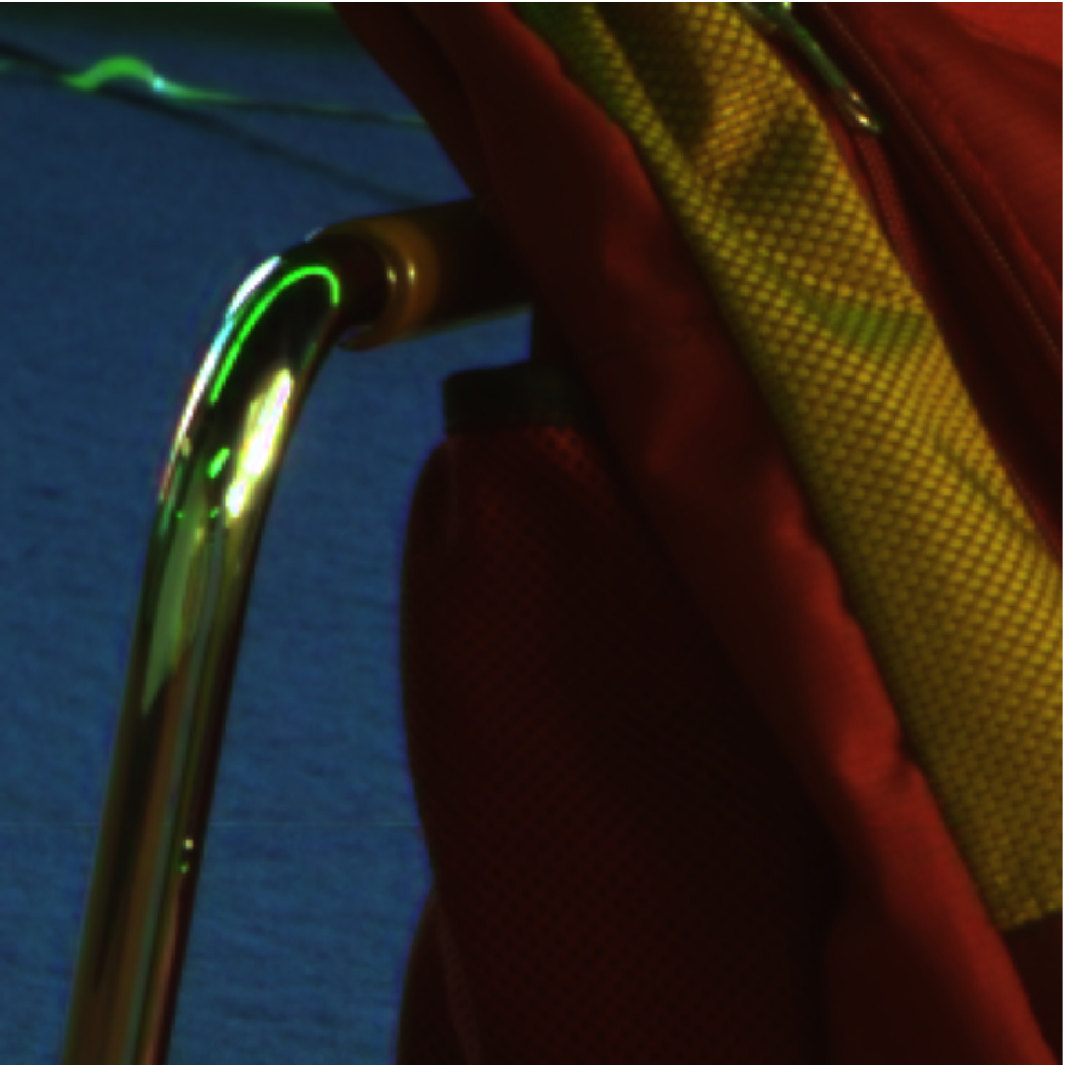}\vspace{0.4mm} \\
			\includegraphics[width=0.655in,height=0.657in]{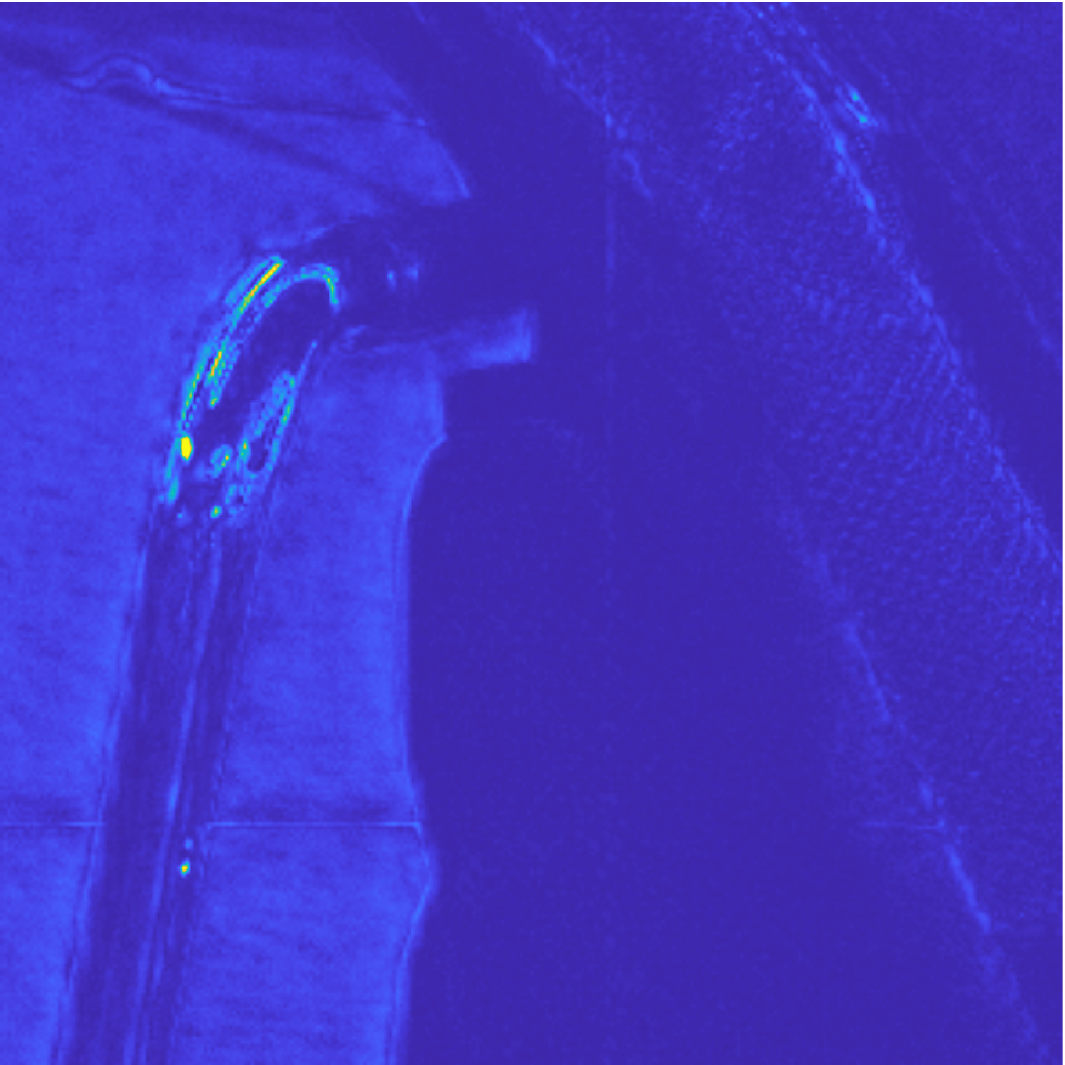}
		\end{minipage}\hspace{-0.3mm}}
	\subfloat[Fusformer]{
		\begin{minipage}[b]{0.095\linewidth}
			\includegraphics[width=0.655in,height=0.657in]{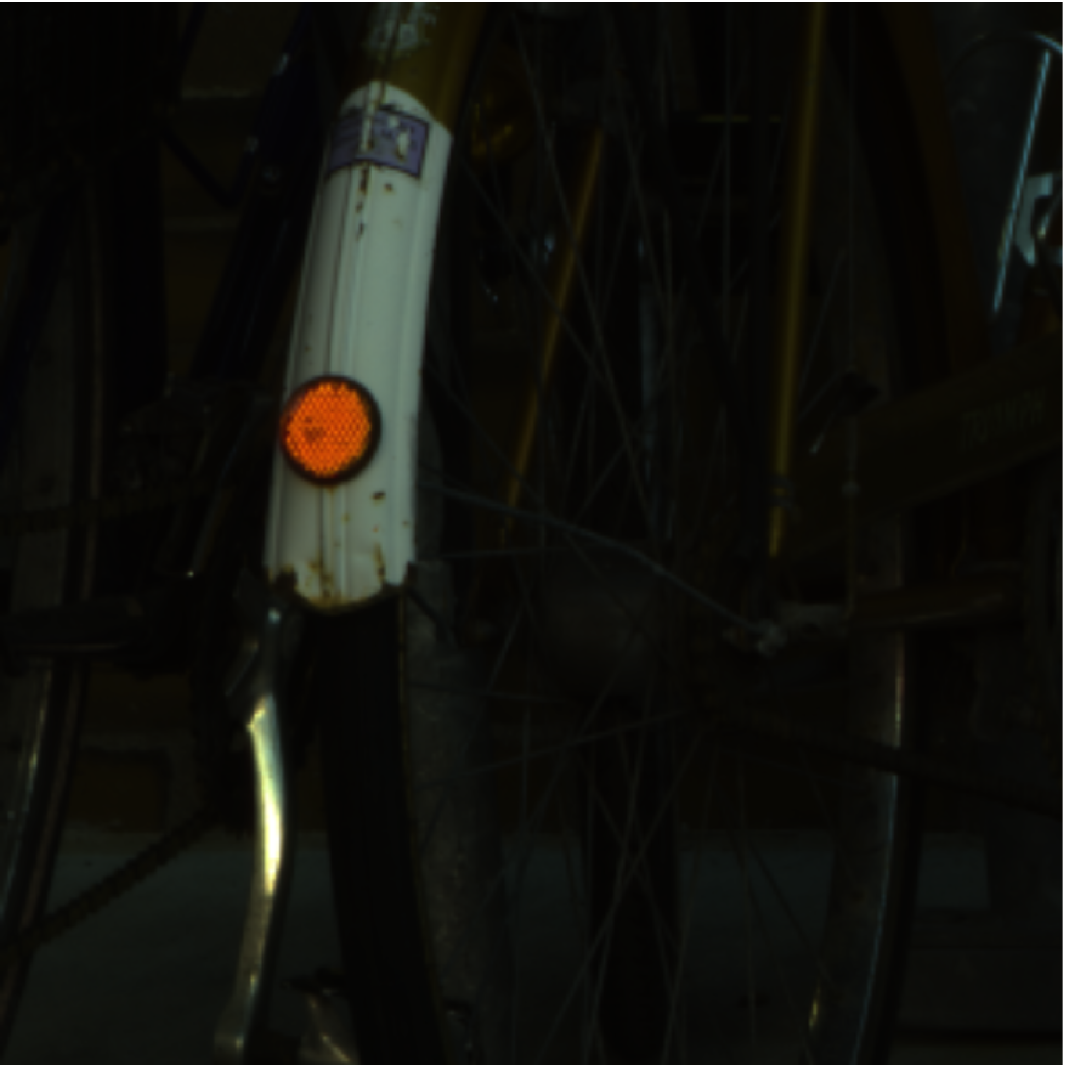}\vspace{0.4mm} \\
			\includegraphics[width=0.655in,height=0.657in]{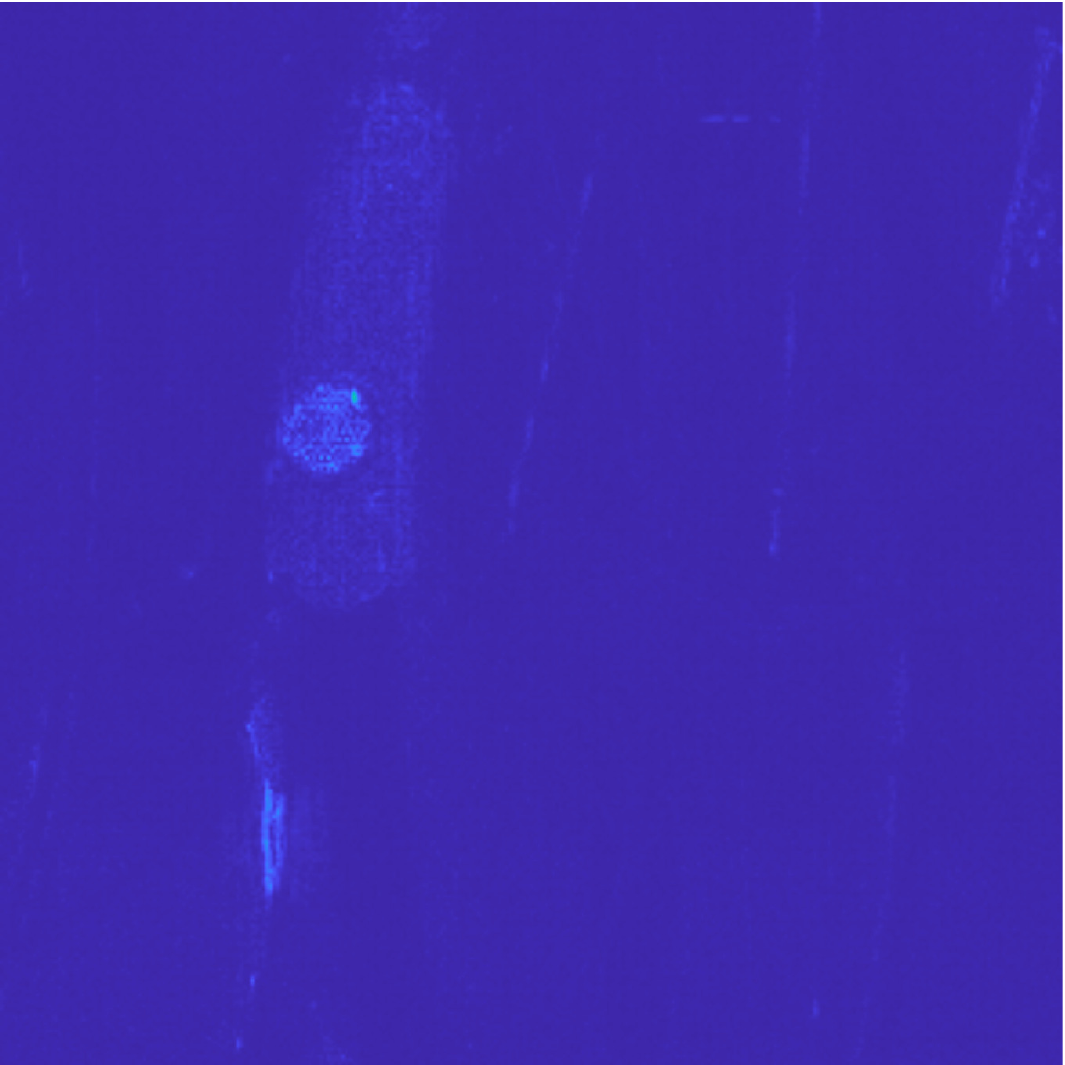}\vspace{0.4mm} \\
			\includegraphics[width=0.655in,height=0.657in]{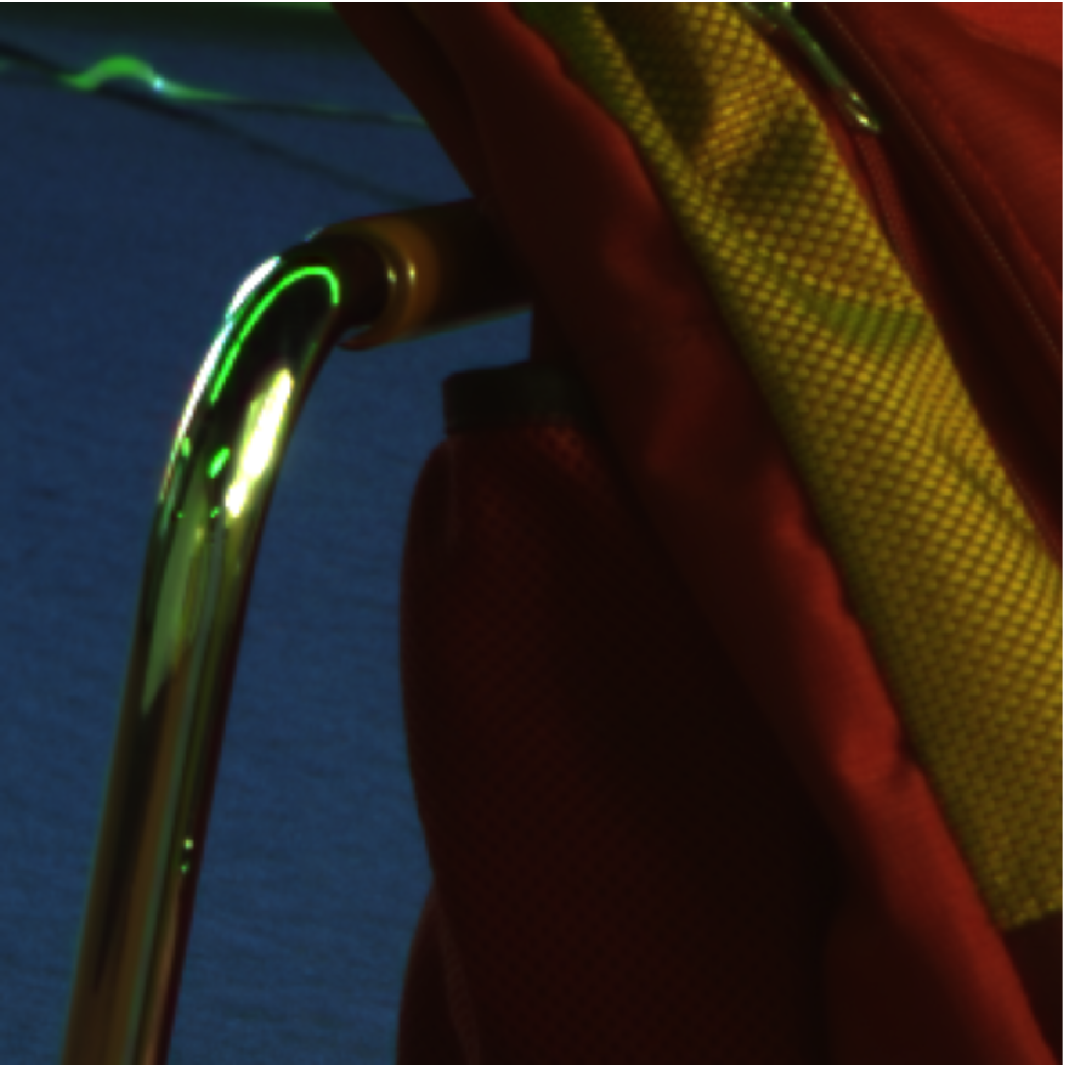}\vspace{0.4mm} \\
			\includegraphics[width=0.655in,height=0.657in]{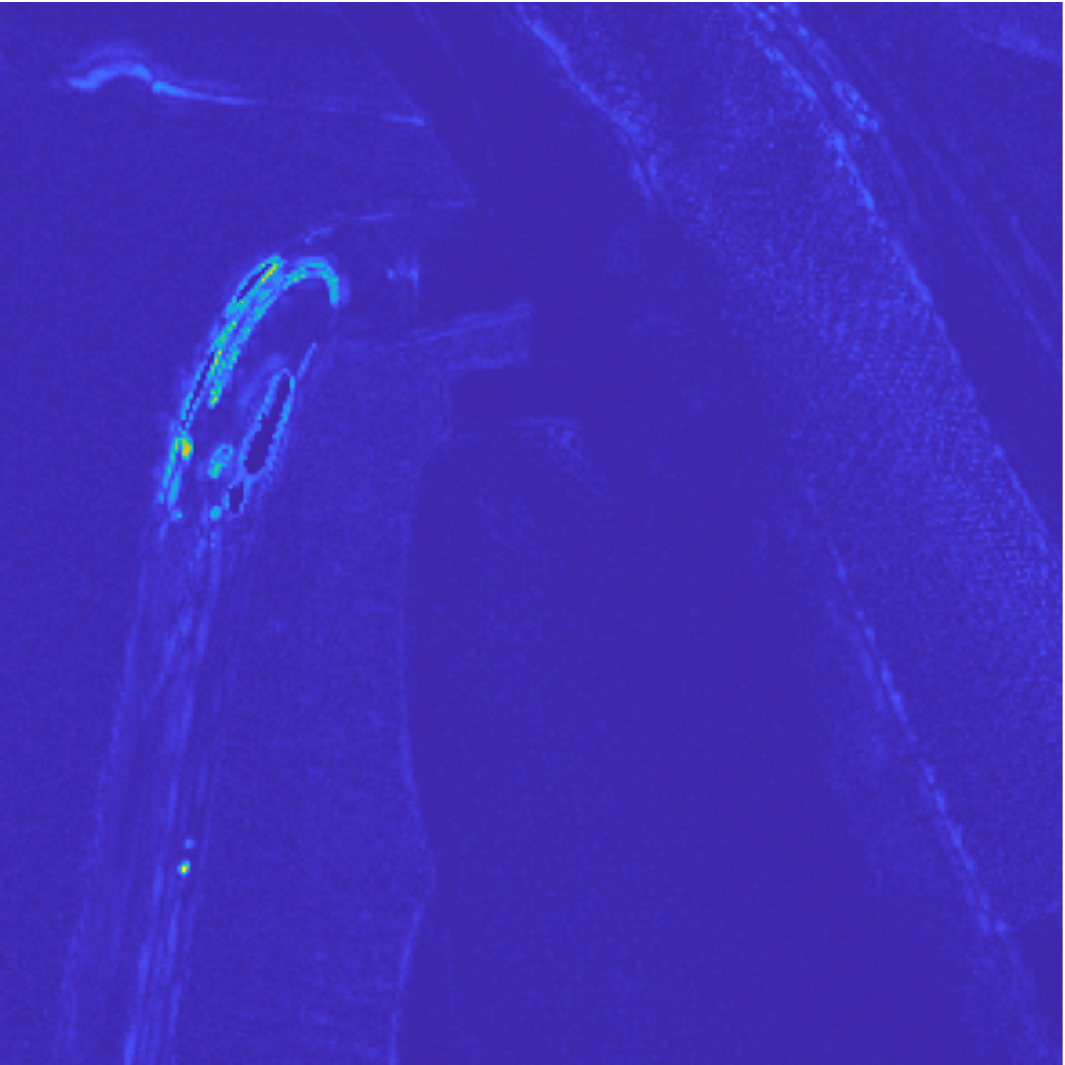}
		\end{minipage}\hspace{-0.3mm}}
	\subfloat[MoG-DCN]{
		\begin{minipage}[b]{0.095\linewidth}
			\includegraphics[width=0.655in,height=0.657in]{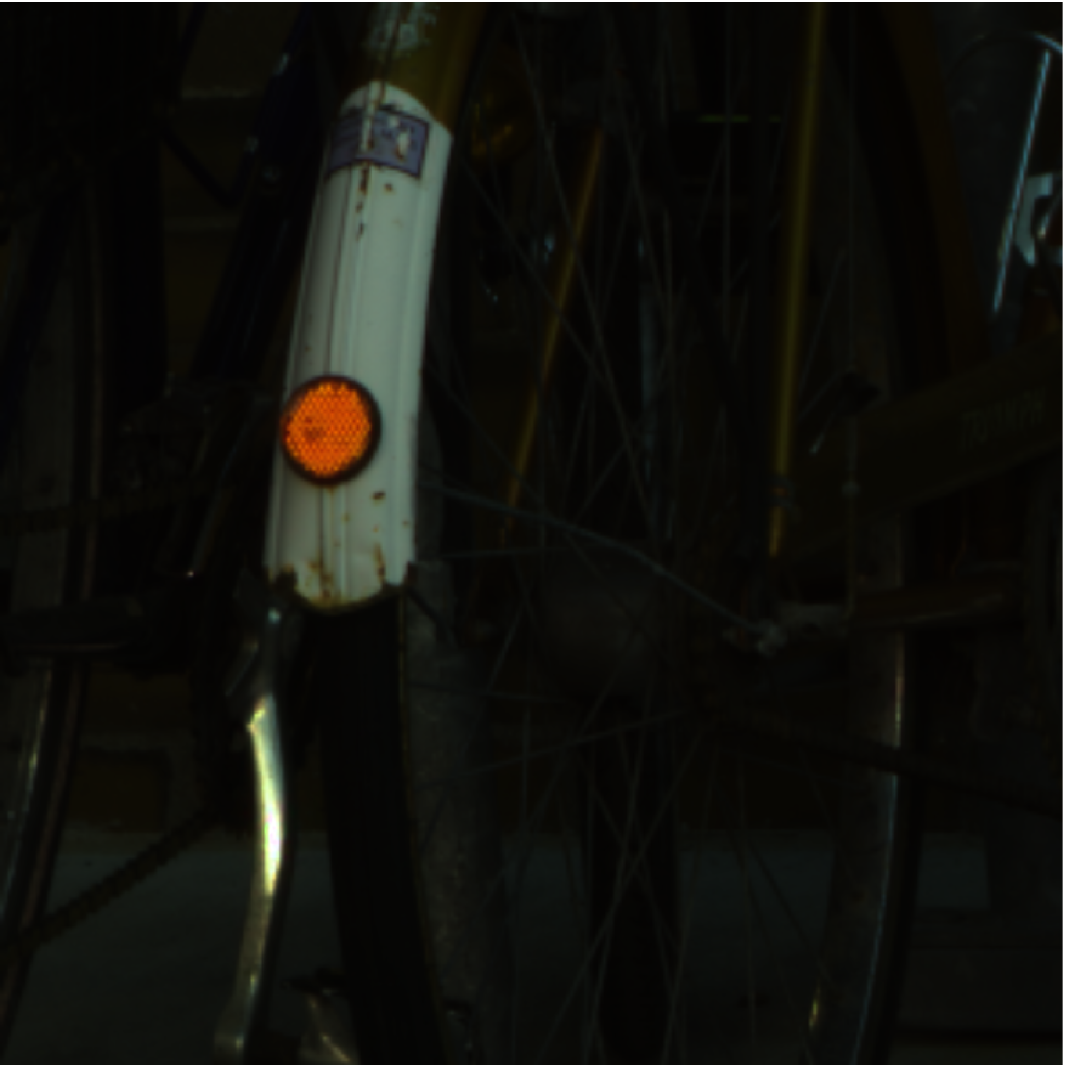}\vspace{0.4mm} \\
			\includegraphics[width=0.655in,height=0.657in]{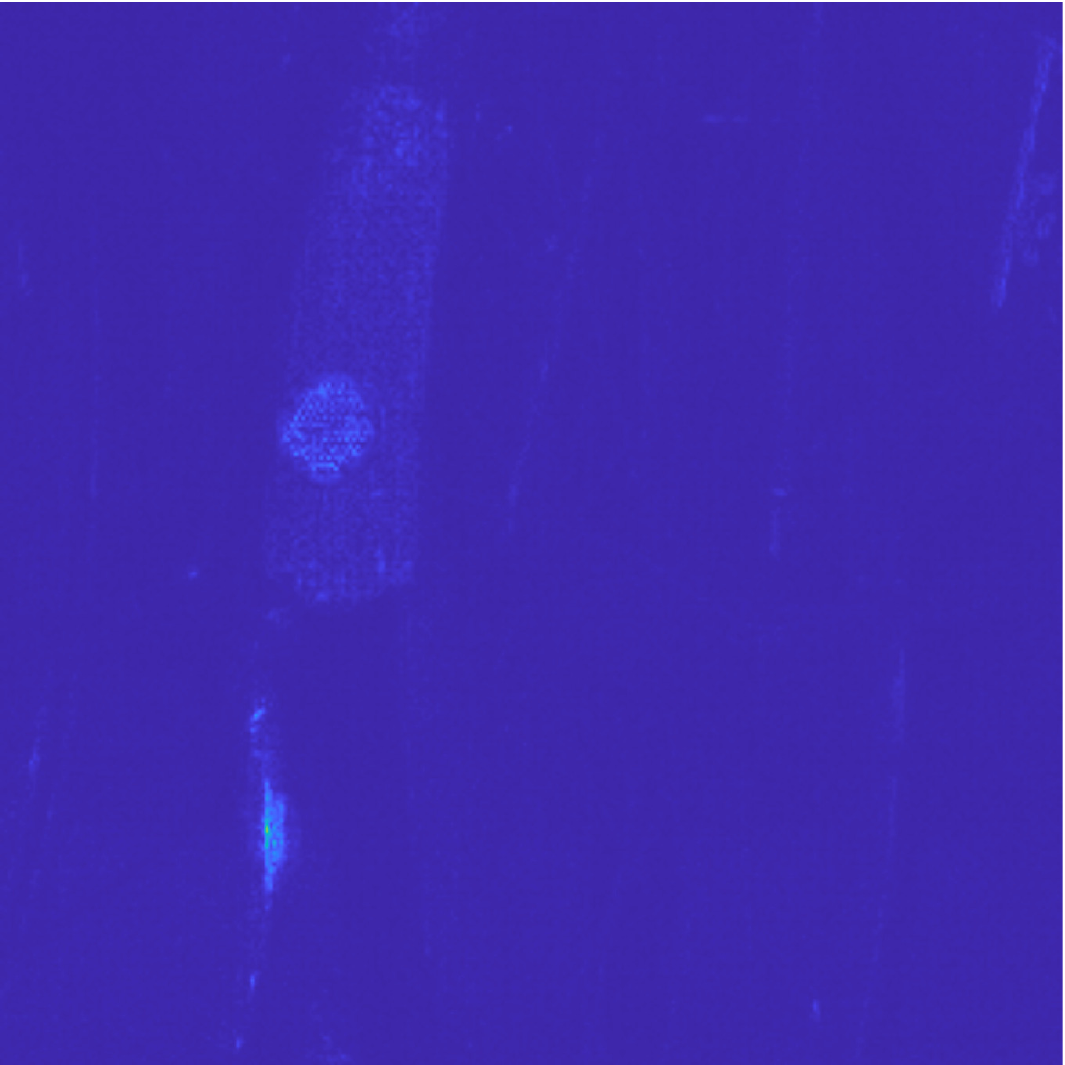}\vspace{0.4mm} \\
			\includegraphics[width=0.655in,height=0.657in]{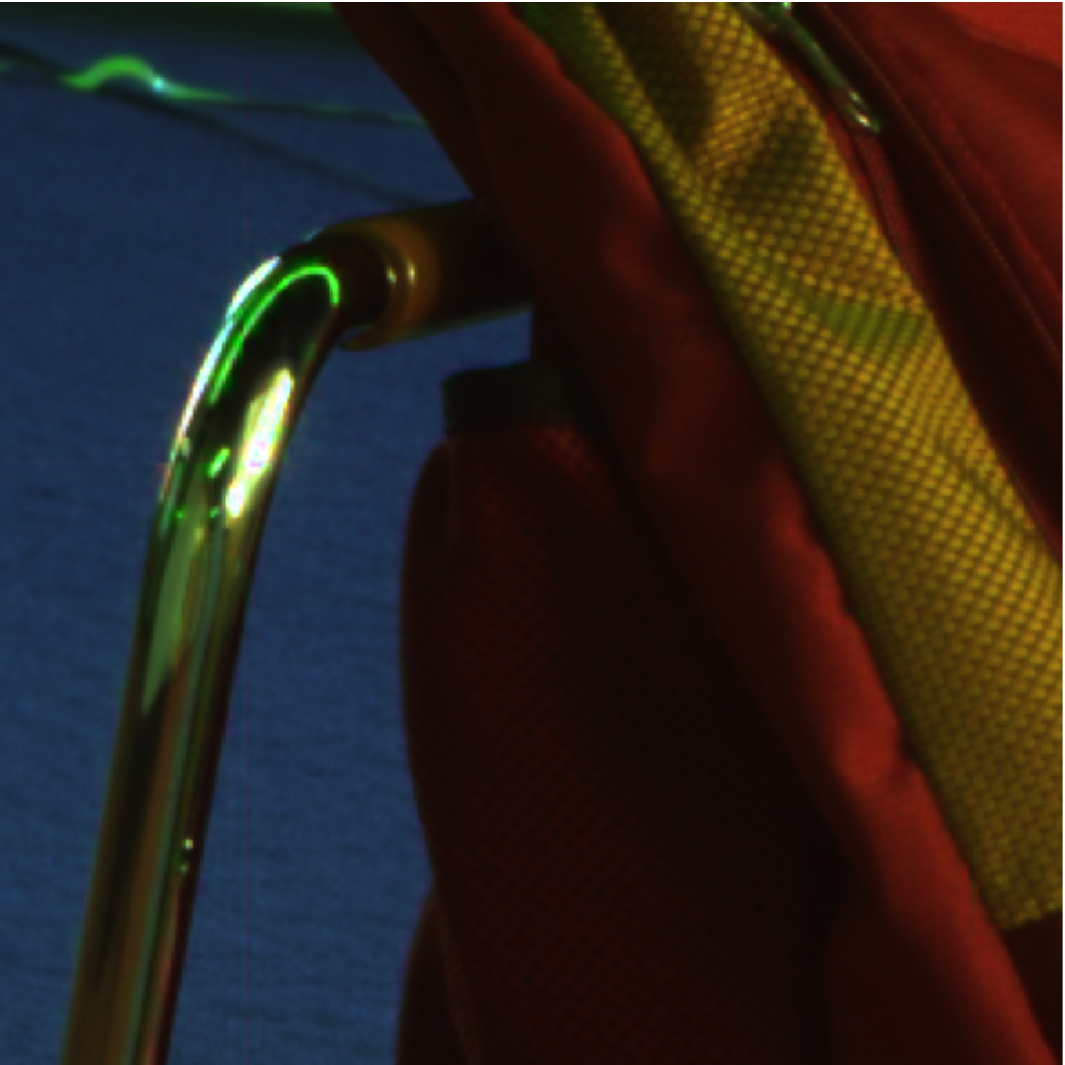}\vspace{0.4mm} \\
			\includegraphics[width=0.655in,height=0.657in]{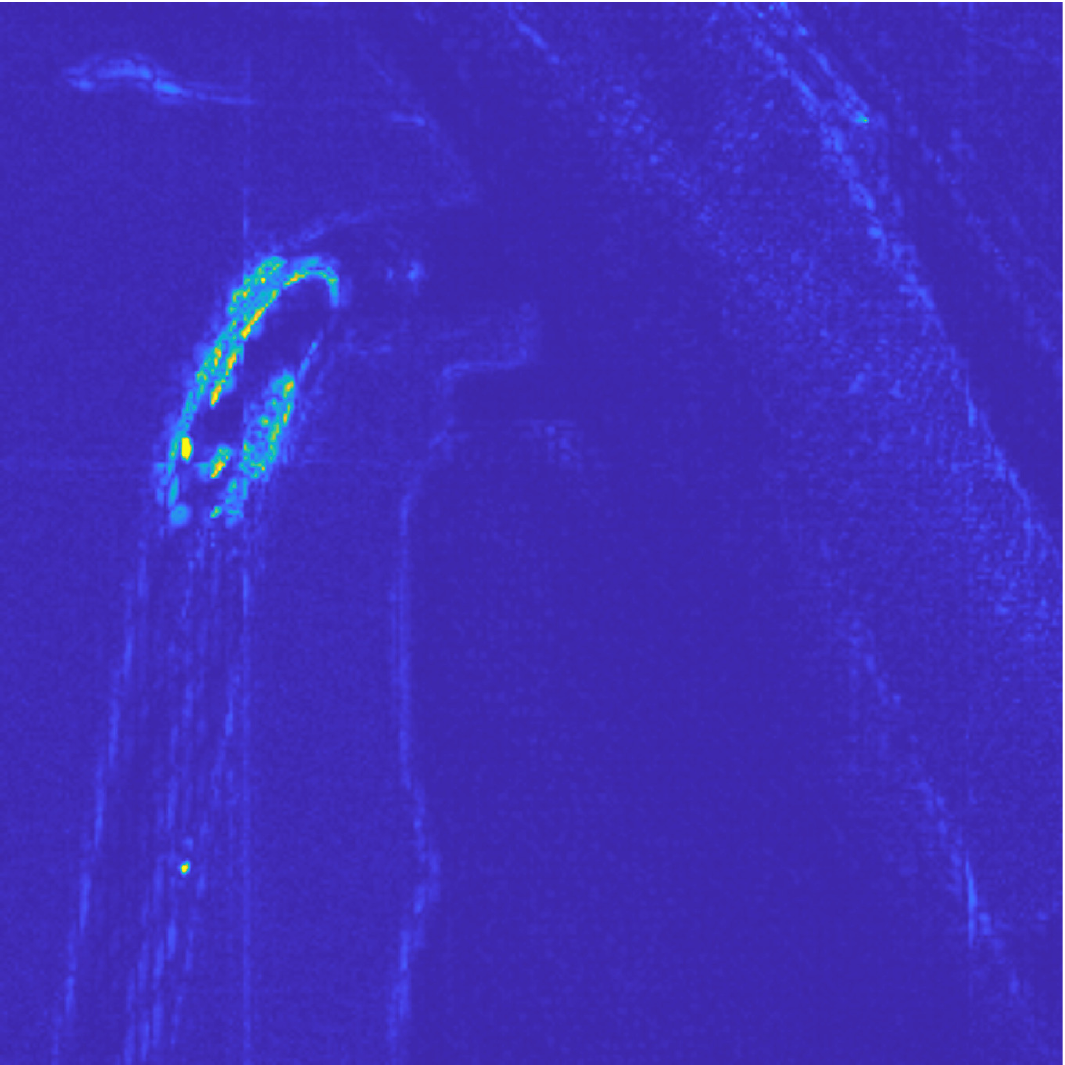}
		\end{minipage}\hspace{-0.3mm}}  
	\subfloat[HSRNet]{
		\begin{minipage}[b]{0.095\linewidth}
			\includegraphics[width=0.655in,height=0.657in]{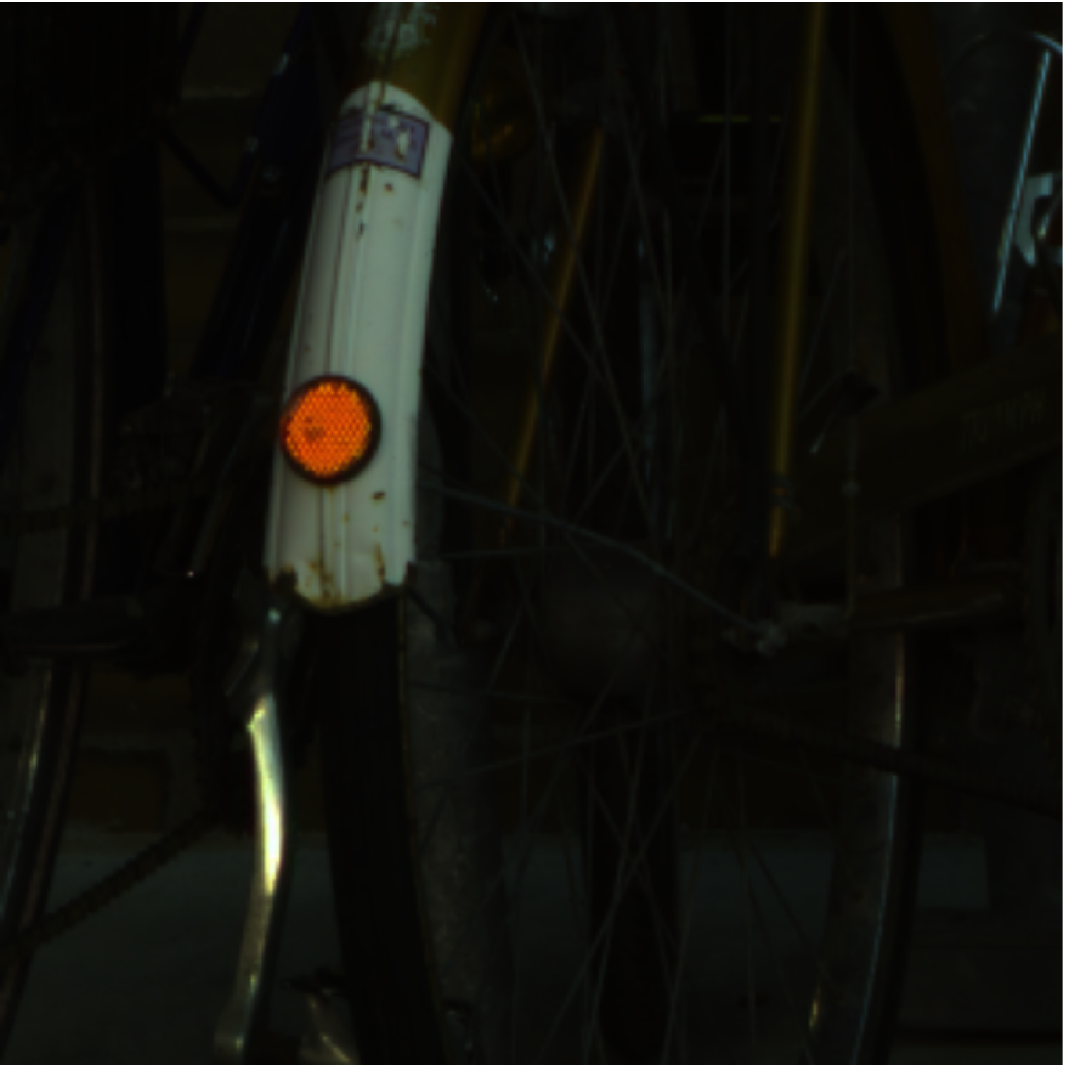}\vspace{0.4mm} \\
			\includegraphics[width=0.655in,height=0.657in]{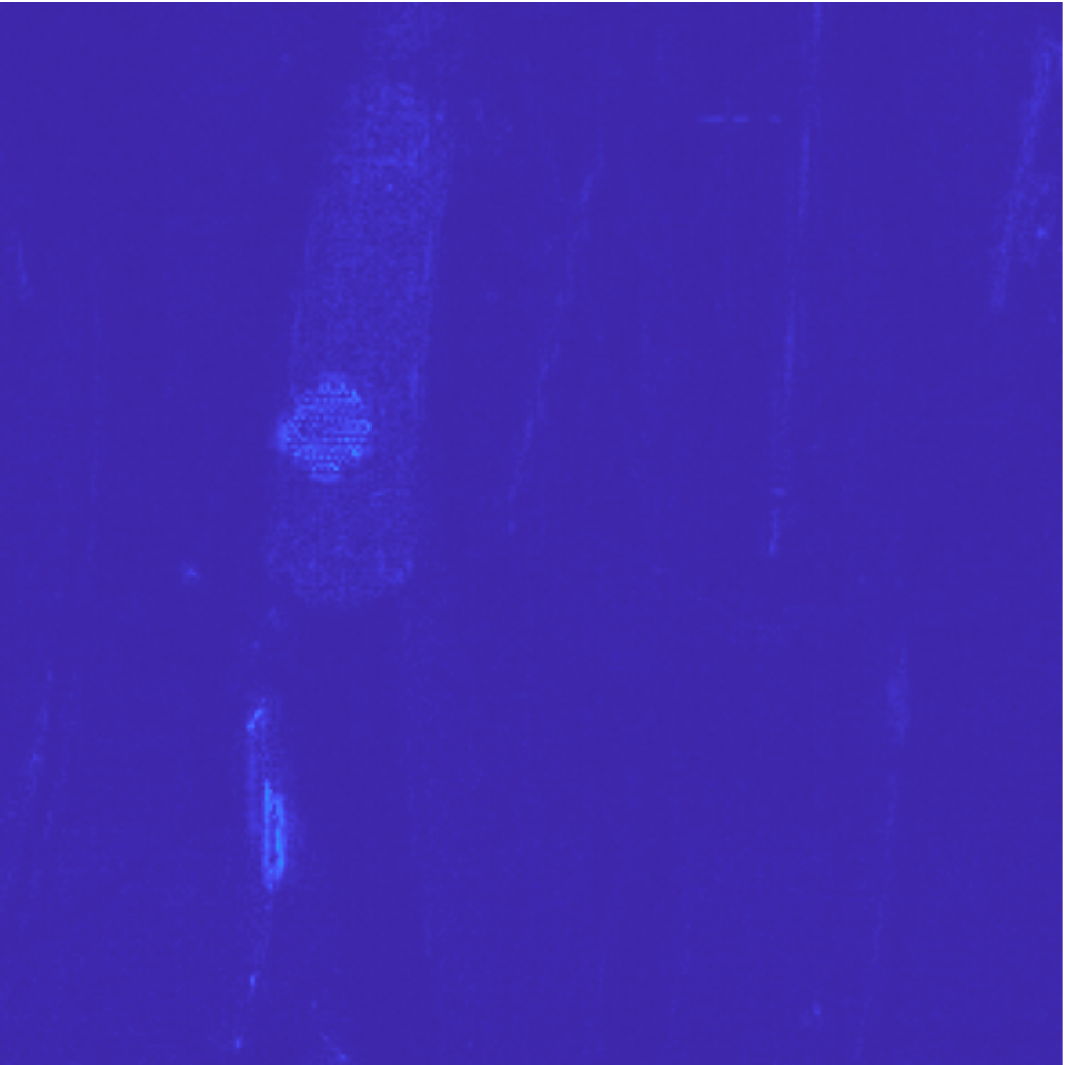}\vspace{0.4mm} \\
			\includegraphics[width=0.655in,height=0.657in]{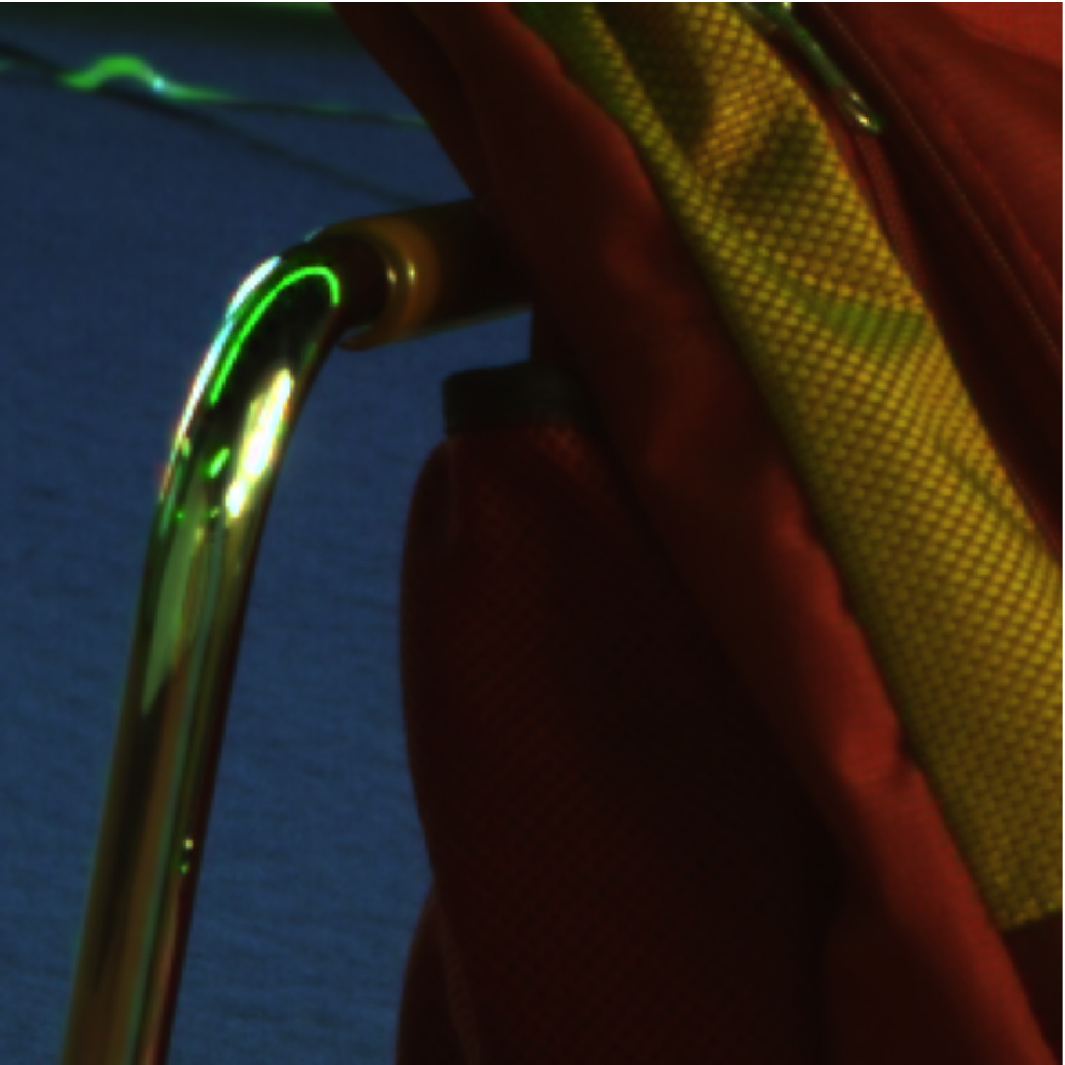}\vspace{0.4mm} \\
			\includegraphics[width=0.655in,height=0.657in]{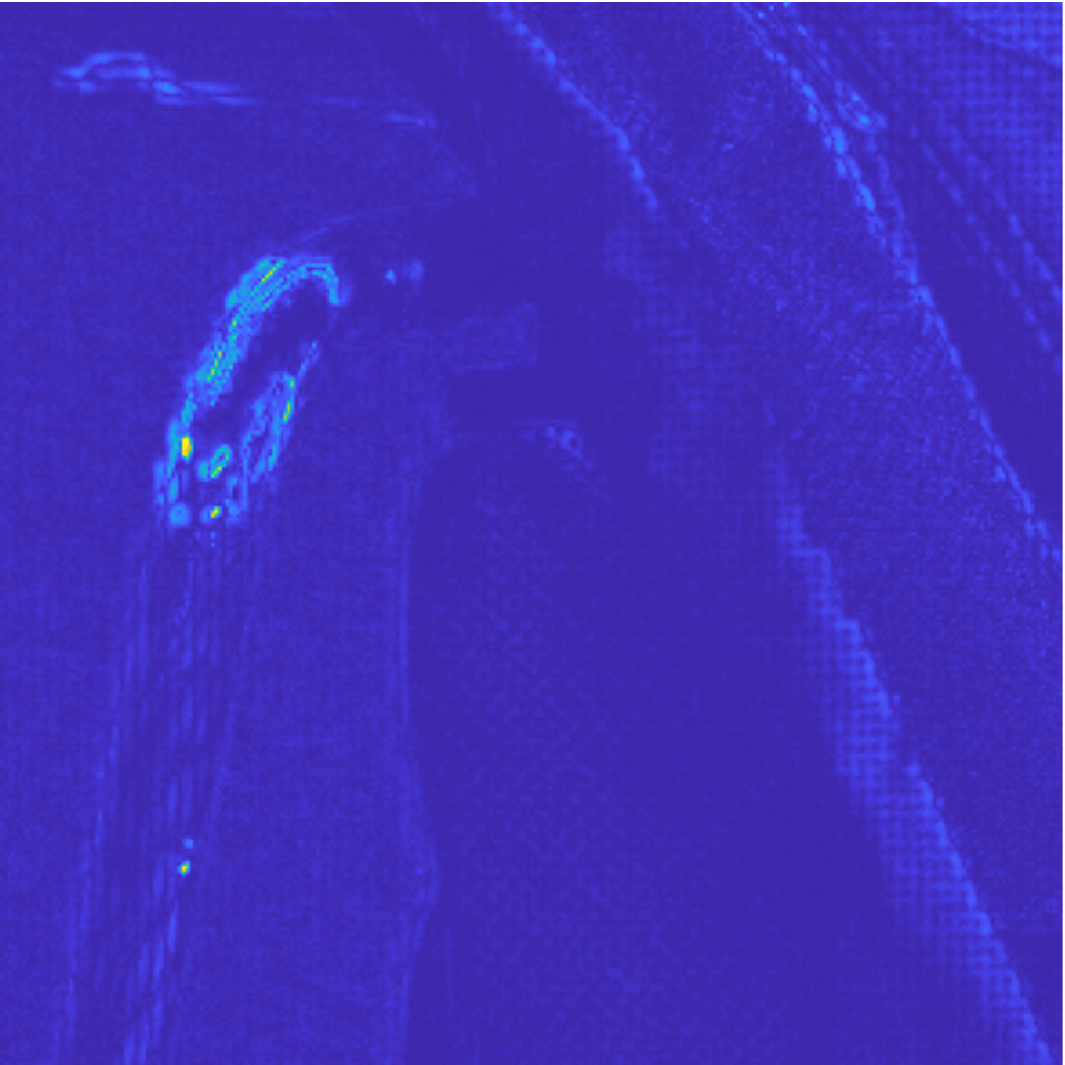}
		\end{minipage}\hspace{-0.3mm}}
	\subfloat[SSRNet]{
		\begin{minipage}[b]{0.095\linewidth}
			\includegraphics[width=0.655in,height=0.657in]{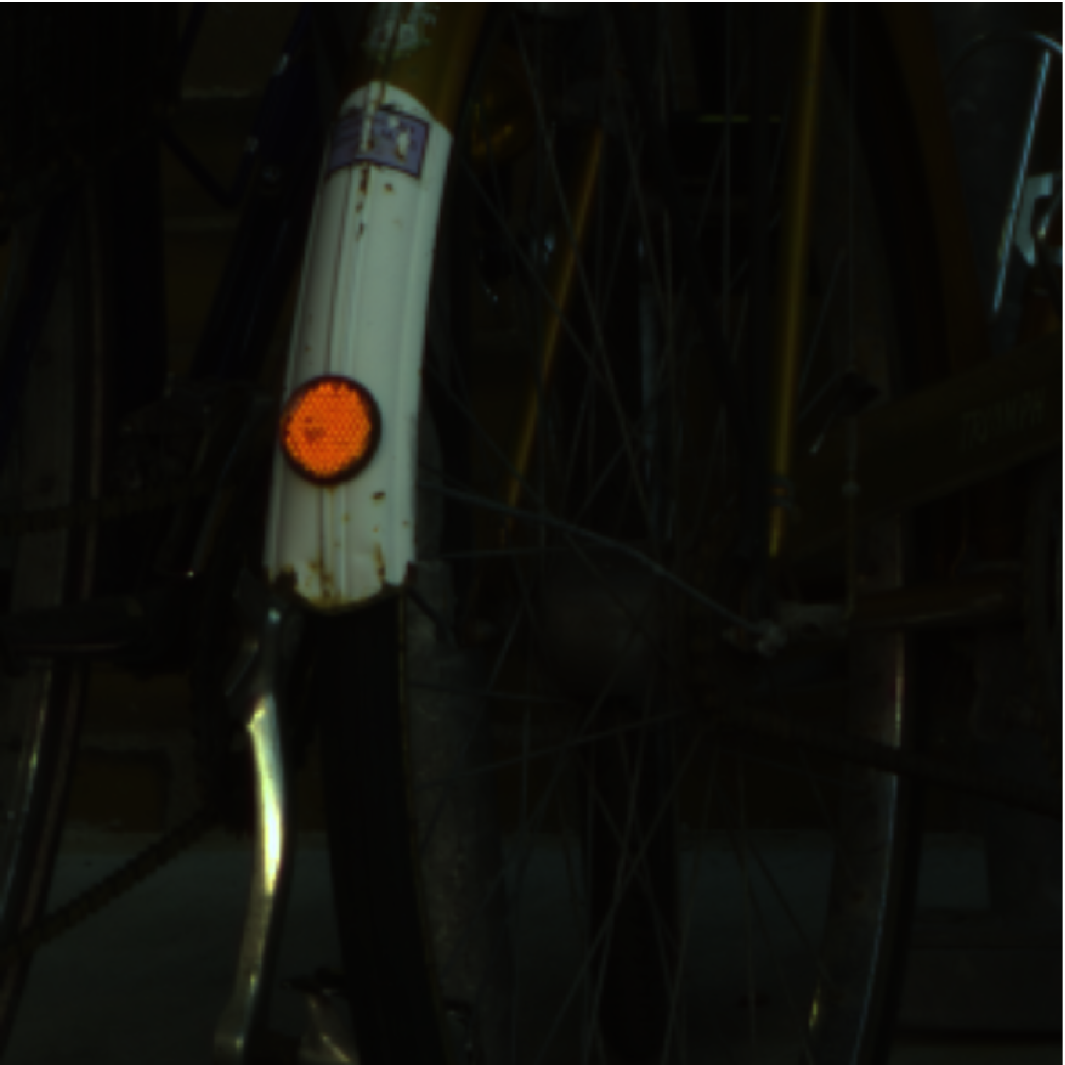}\vspace{0.4mm} \\
			\includegraphics[width=0.655in,height=0.657in]{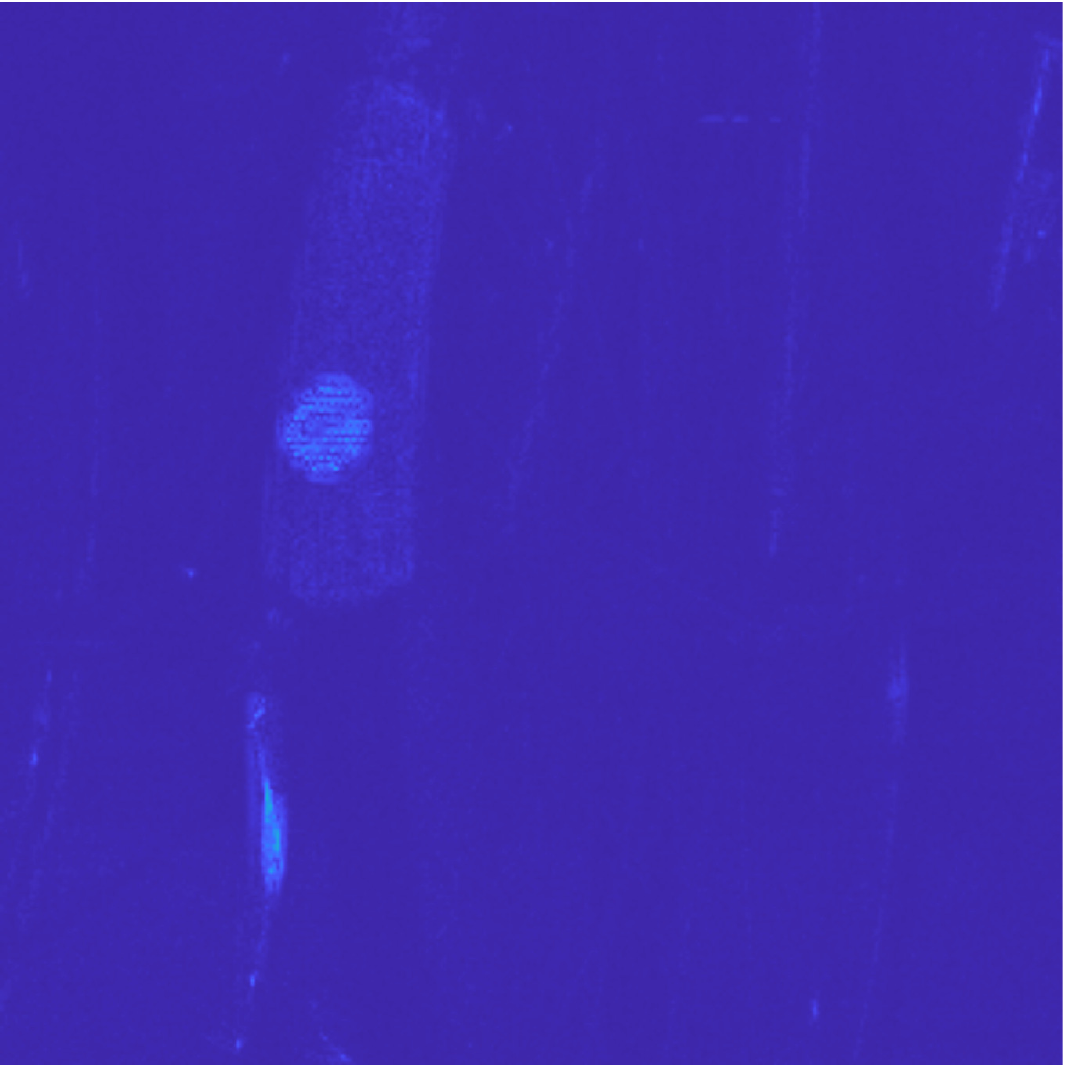}\vspace{0.4mm} \\
			\includegraphics[width=0.655in,height=0.657in]{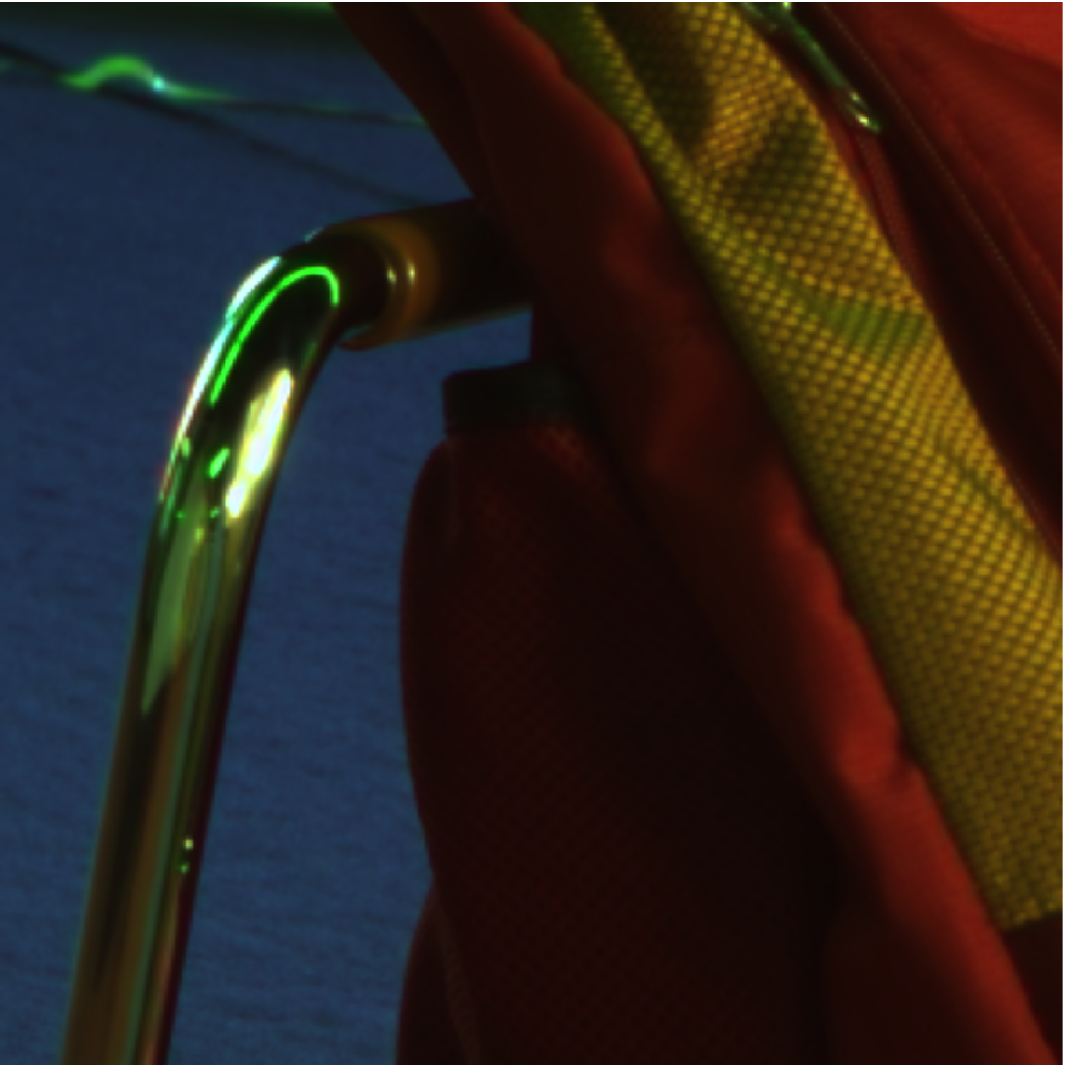}\vspace{0.4mm} \\
			\includegraphics[width=0.655in,height=0.657in]{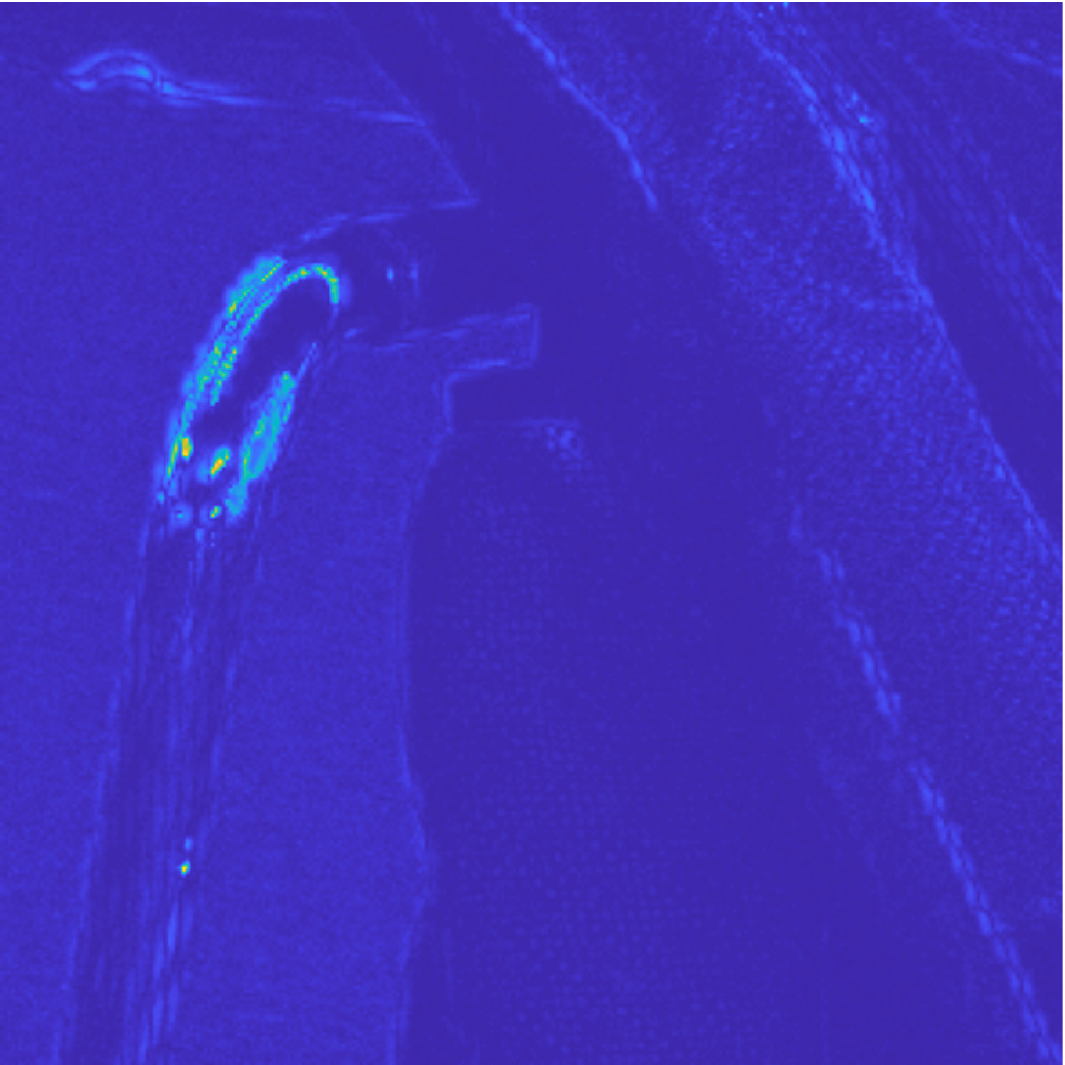}
		\end{minipage}\hspace{-0.3mm}} 
	\subfloat[DBIN]{
		\begin{minipage}[b]{0.095\linewidth}
			\includegraphics[width=0.655in,height=0.657in]{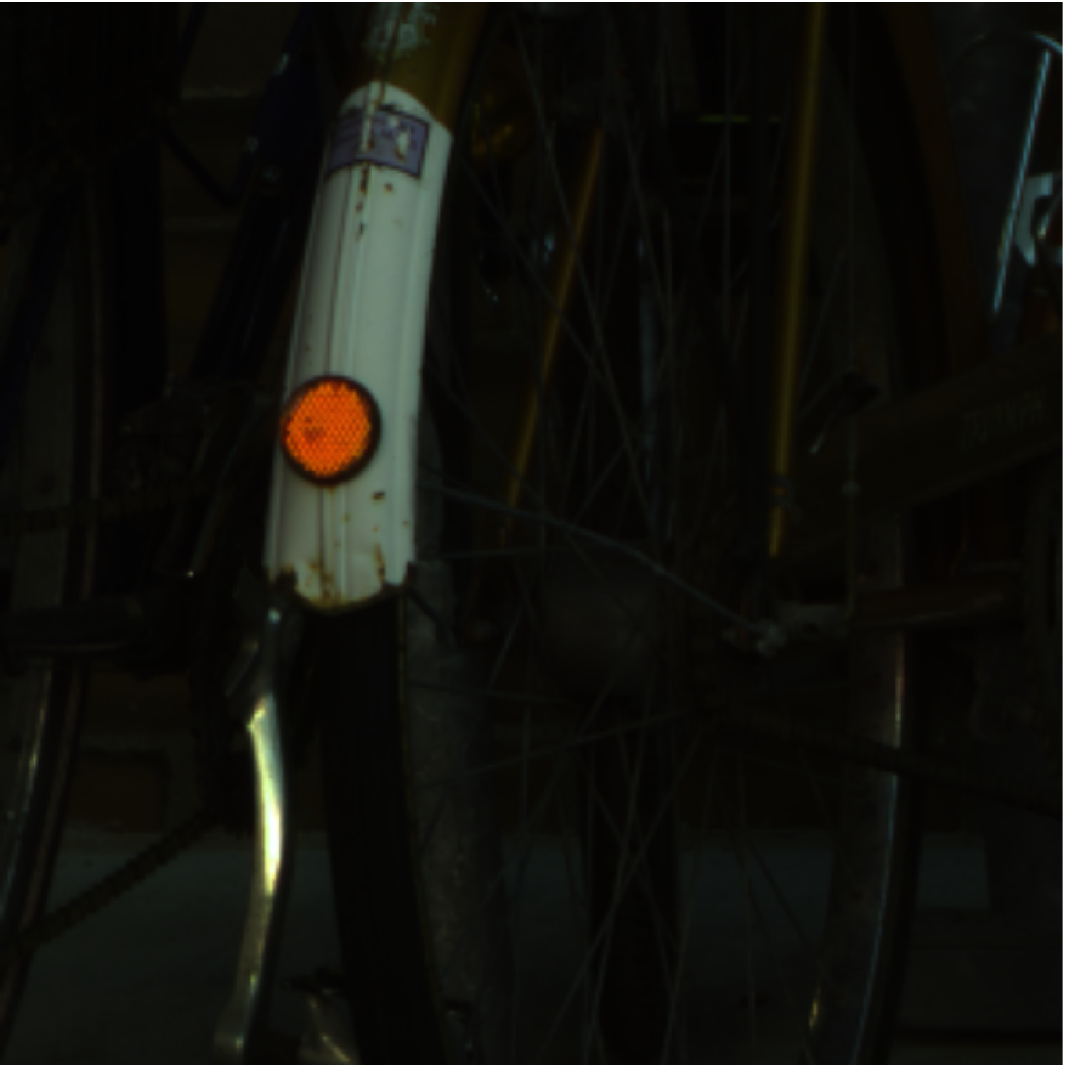}\vspace{0.4mm} \\
			\includegraphics[width=0.655in,height=0.657in]{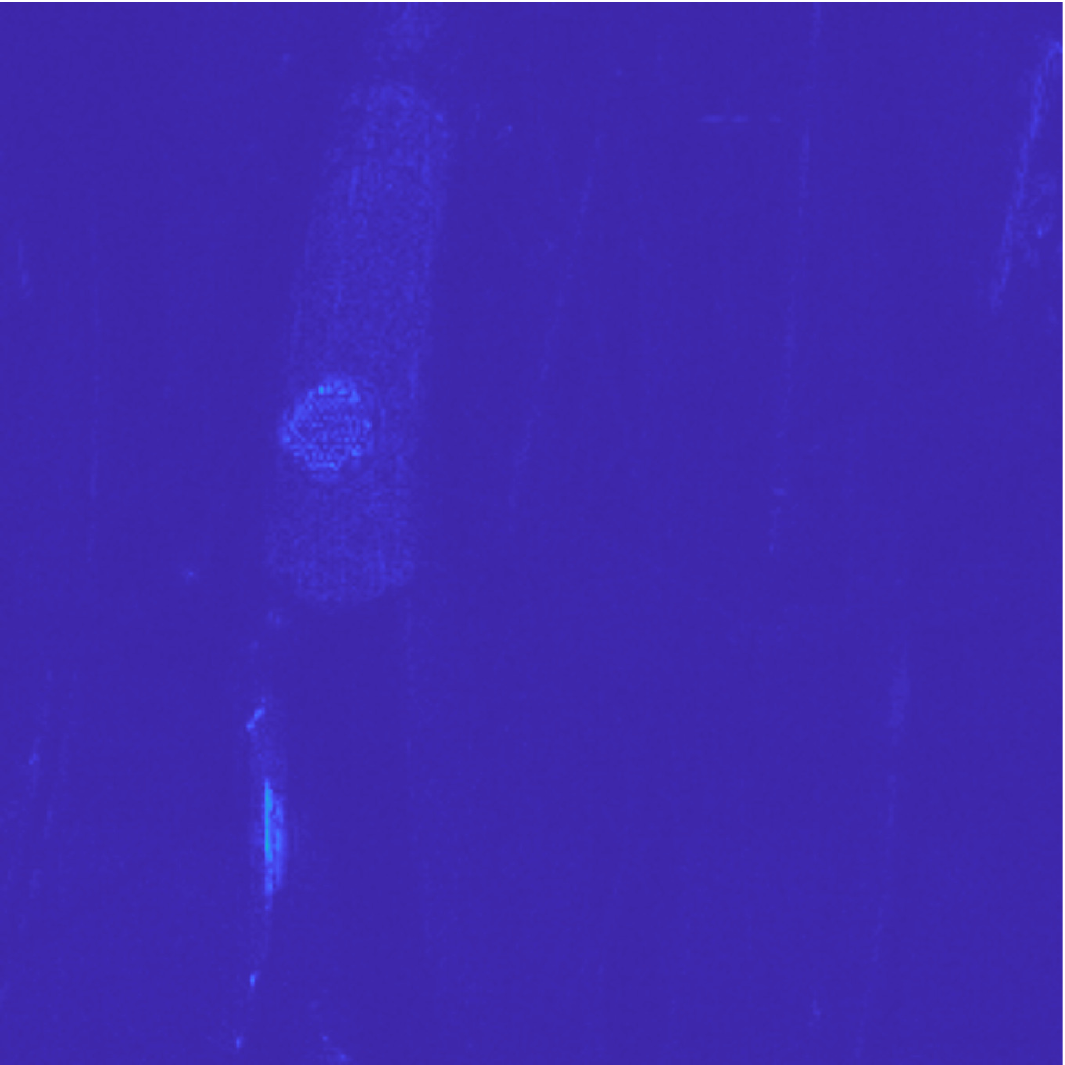}\vspace{0.4mm} \\
			\includegraphics[width=0.655in,height=0.657in]{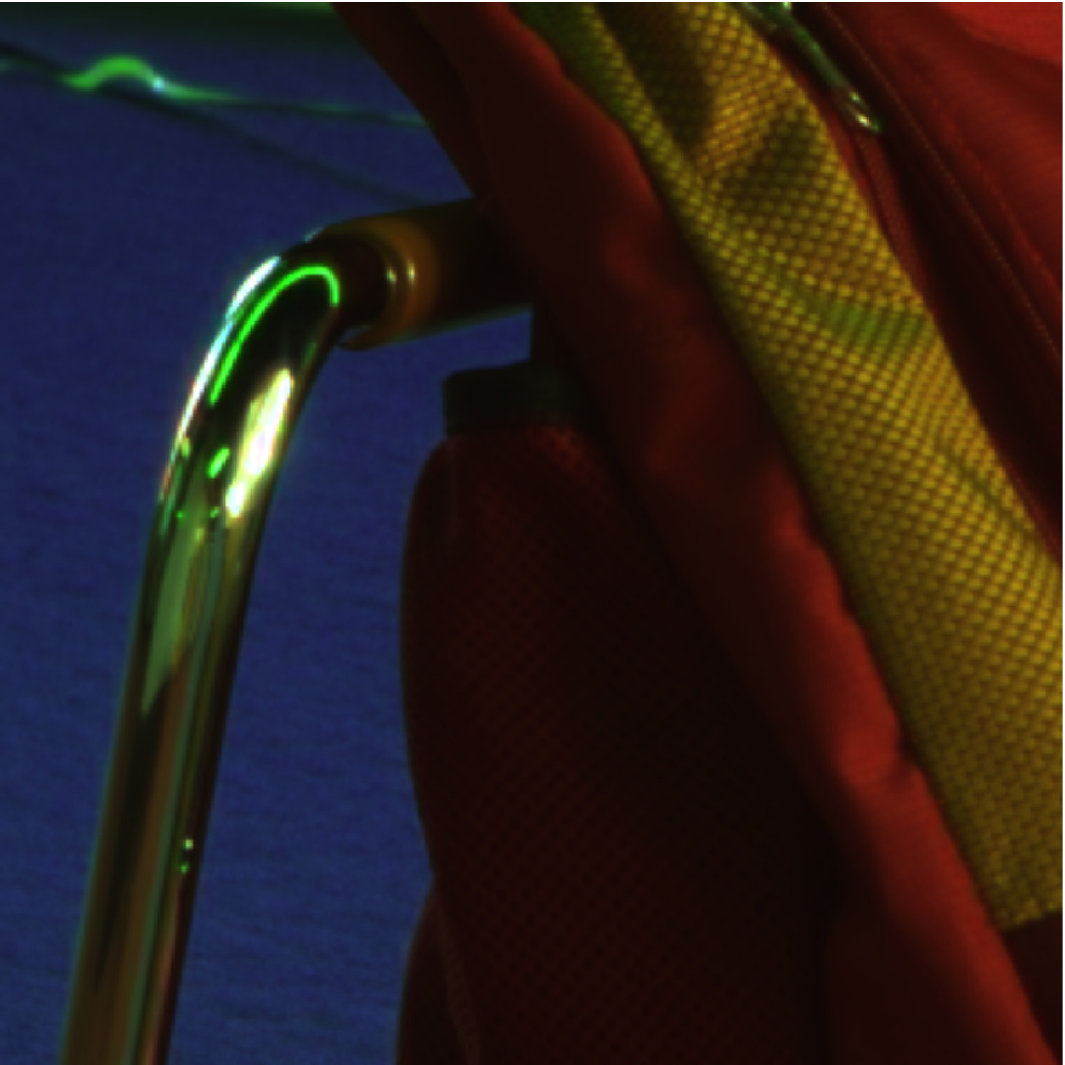}\vspace{0.4mm} \\
			\includegraphics[width=0.655in,height=0.657in]{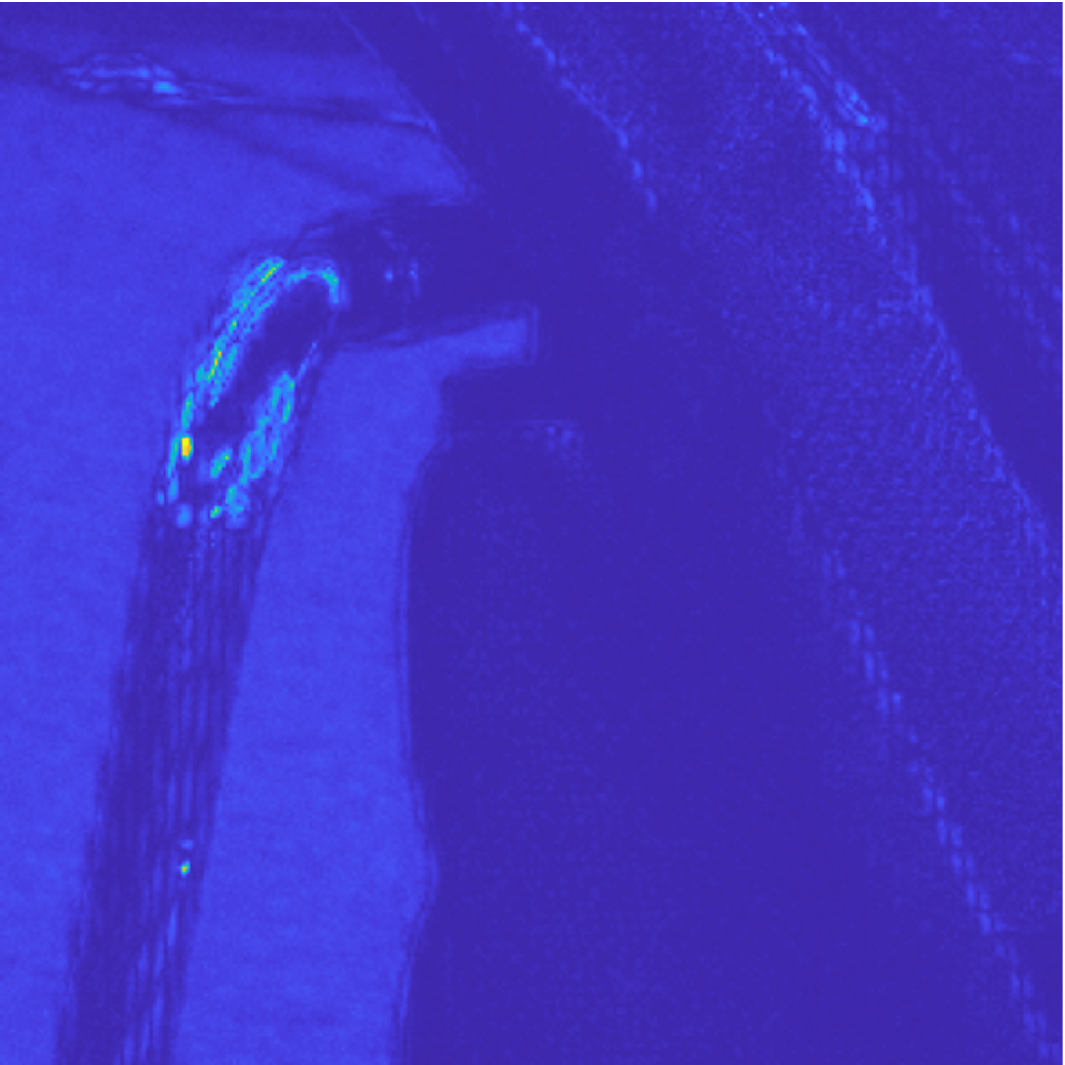}
		\end{minipage}\hspace{-0.3mm}}
	\subfloat[ResTFNet]{
		\begin{minipage}[b]{0.095\linewidth}
			\includegraphics[width=0.655in,height=0.657in]{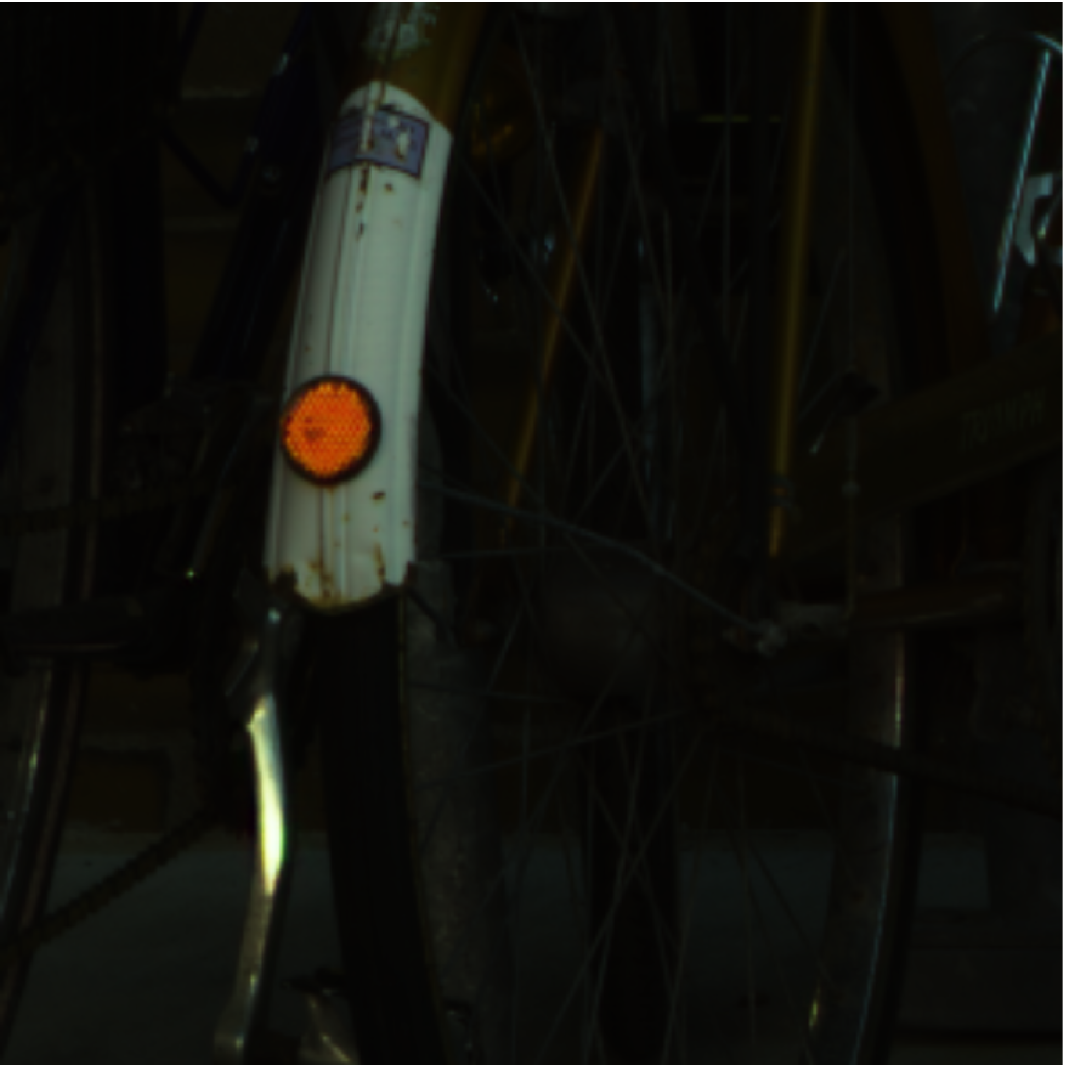}\vspace{0.4mm} \\
			\includegraphics[width=0.655in,height=0.657in]{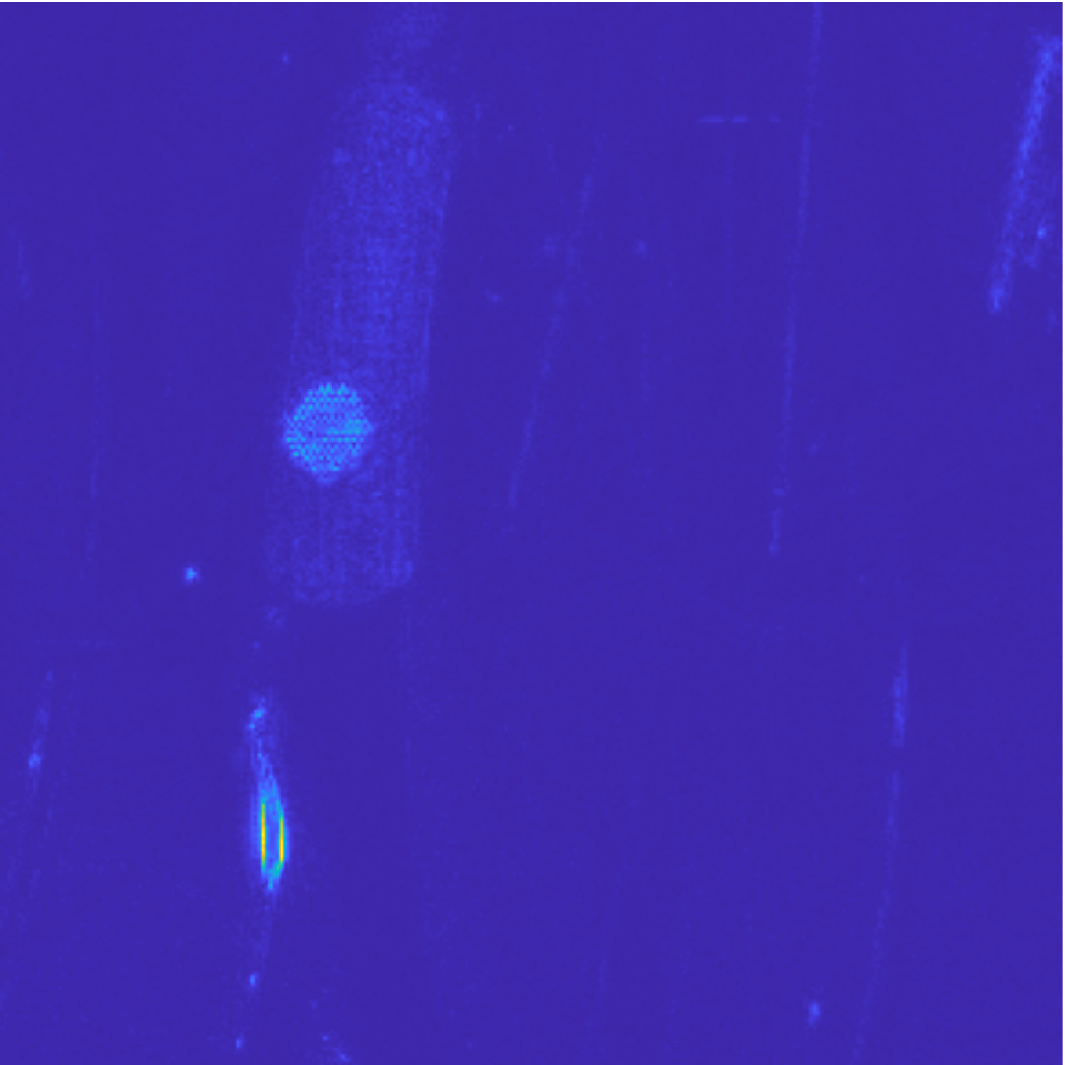}\vspace{0.4mm} \\
			\includegraphics[width=0.655in,height=0.657in]{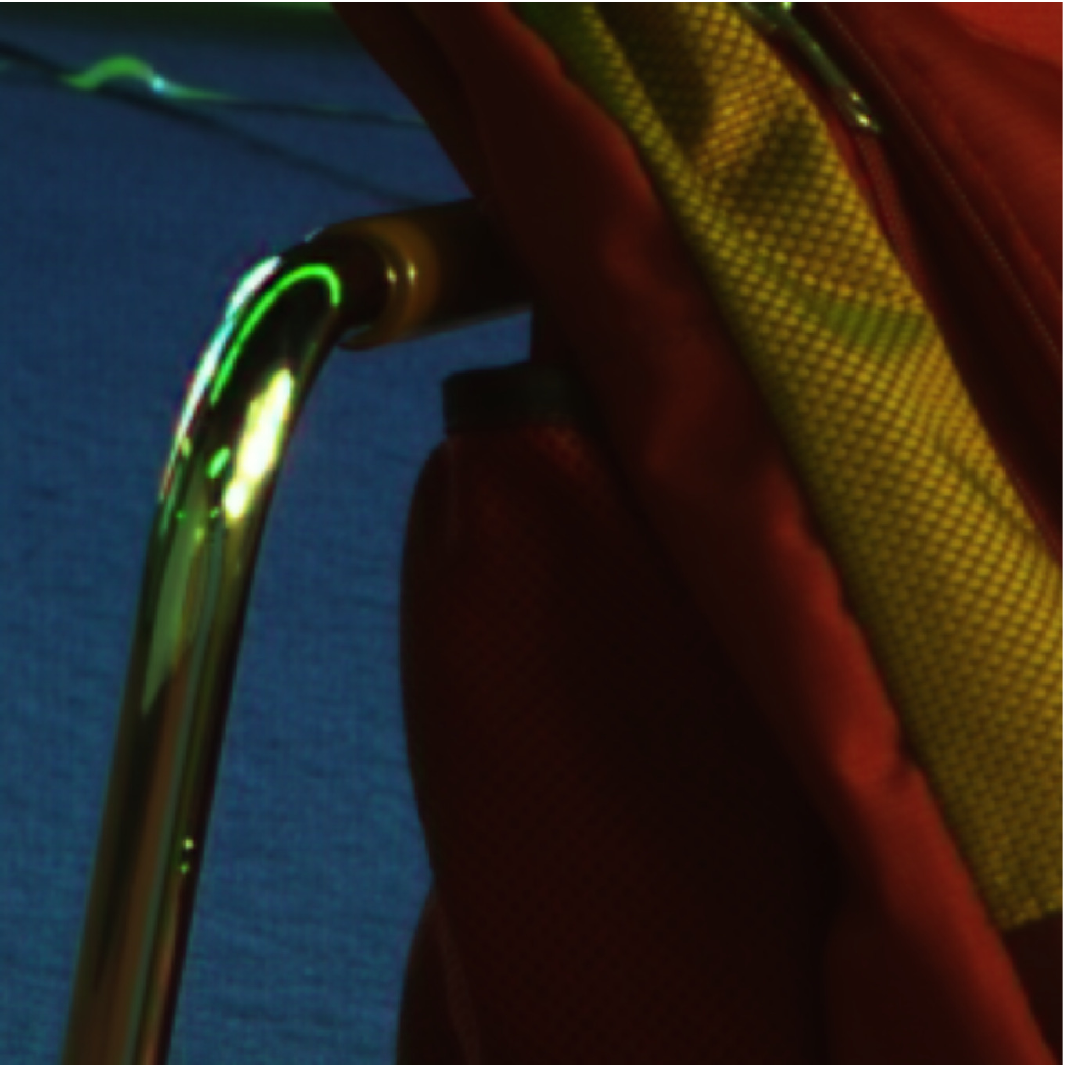}\vspace{0.4mm} \\
			\includegraphics[width=0.655in,height=0.657in]{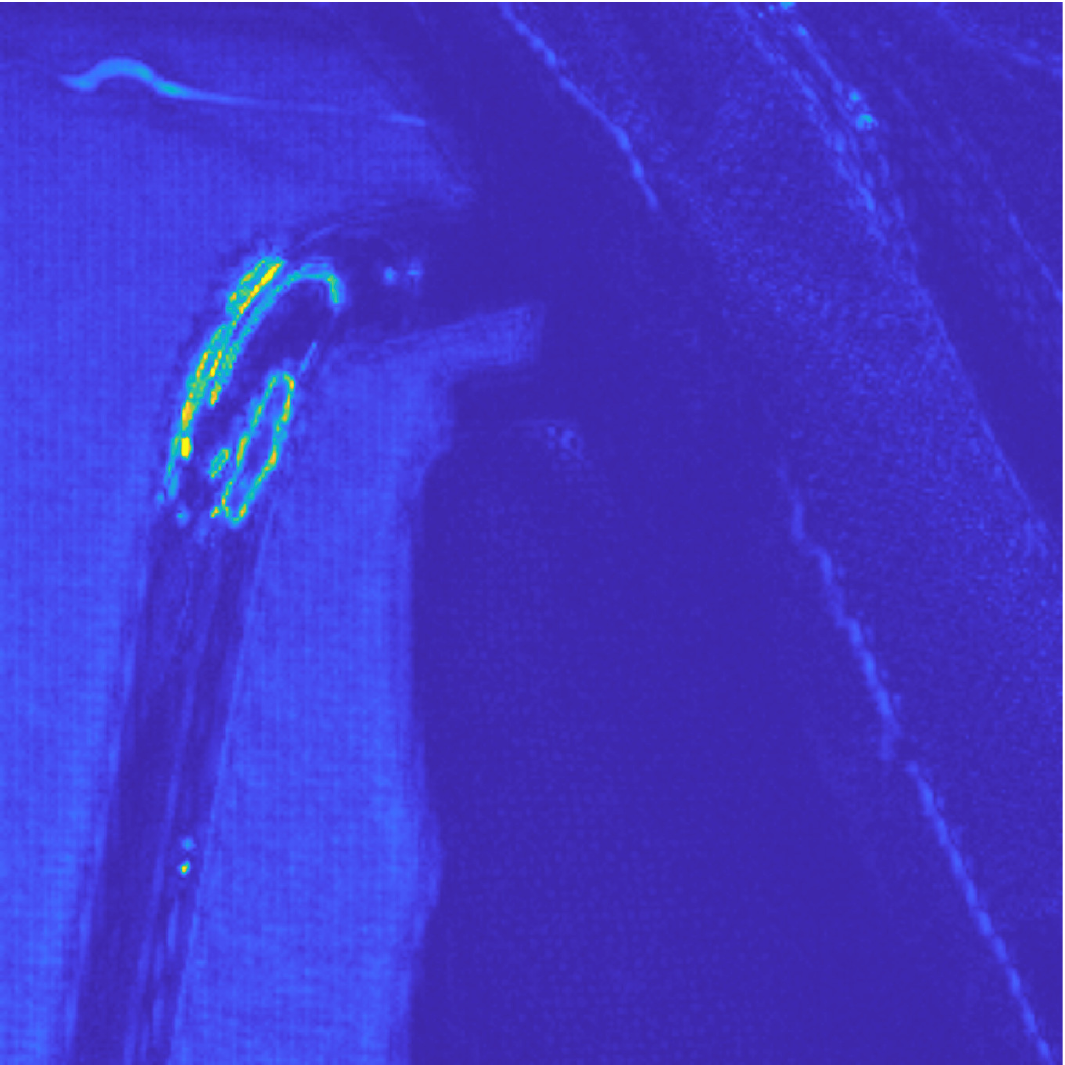}
		\end{minipage}\hspace{-0.3mm}}
	\subfloat{
		\begin{minipage}[b]{0.095\linewidth}
			\includegraphics[width=0.28in,height=2.5in]{Fig/colorbar.pdf}\vspace{0.5mm} \\
	\end{minipage}}
	
	\caption{\footnotesize{The first and third rows show the results using the pseudo-color representation from the Harvard dataset. Noting that we select three bands (31-20-10) from HSIs as the red, green and blue channels. We zoomed in on the blue rectangles to show more detail. The second and fourth rows show the residuals between the GT and the fused products. (a) GT, (b) Ours, (c) DHIF~\cite{huang2022deep}, (d) Fusformer~\cite{hu2022fusformer}, (e) MoG-DCN~\cite{dong2021model}, (f) HSRNet~\cite{hu2021hyperspectral}, (g) SSRNet~\cite{zhang2020ssr}, (h) DBIN~\cite{Wang_2019_ICCV} and (i) ResTFNet~\cite{LIU20201}. }}
	\label{harvard_comparison}
\end{figure*}

\noindent$\textbf{Benchmark:}$ To verify the superiority of the proposed INF³, we compare it with various state-of-the-art methods including MTF-GLP-HS~\cite{selva2015hyper}, CSTF-FUS~\cite{li2018fusing}, LTTR\cite{dian2019learning}, LTMR\cite{dian2019hyperspectral}, IR-TenSR\cite{xutgrs2022}, DBIN~\cite{Wang_2019_ICCV}, SSRNet~\cite{zhang2020ssr}, ResTFNet~\cite{LIU20201}, HSRNet~\cite{hu2021hyperspectral}, MoG-DCN~\cite{dong2021model}, Fusformer~\cite{hu2022fusformer} and the DHIF~\cite{huang2022deep} network. In specific, the upsampled LR-HSI in Fig.~\ref{network} is the bicubic-interpolated result, which is added to the experiment as a baseline. By the way, all the deep learning approaches are trained with the same input pairs for a fair comparison. Moreover, the related hyperparameters are selected consistent with the original papers.

\noindent$\textbf{Implementation Details:}$
The proposed network implements in PyTorch 1.11.0 and Python 3.8.0 using Adam optimizer\cite{kingma2014adam} with a learning rate of 0.0001 to minimize sum of absolute difference $\mathcal{L}_{1}$ by 1000 epochs and Linux operating system with a NVIDIA RTX3080 GPU (12GB). 

\begin{figure}[h]
	\centering
	\includegraphics[width=8.5cm, height=4.2cm]{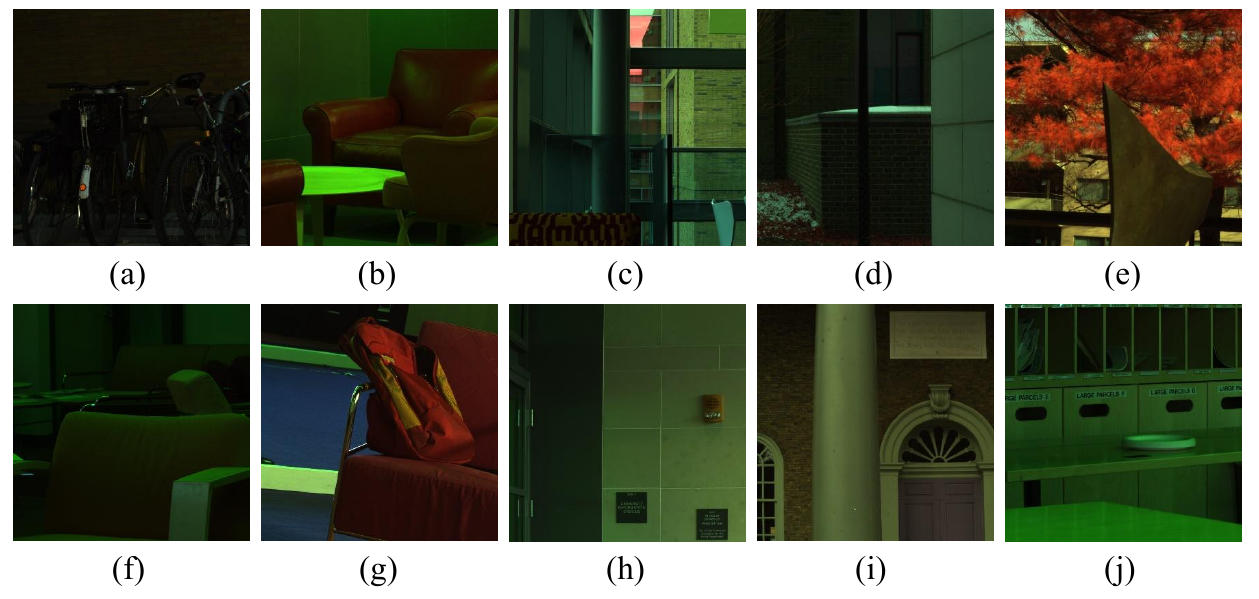}
	\caption{The 10 images tested on the Harvard dataset are (a) \emph{bikes}, (b) \emph{sofa1}, (c) \emph{window}, (d) \emph{fence}, (e) \emph{tree}, (f) \emph{sofa2}, (g) \emph{backpack}, (h) \emph{wall}, (i) \emph{door} and (j) \emph{parcels}. }
	\label{fig_HARVARD}
	\vspace{-5mm}
\end{figure}

\noindent{\bf Results on CAVE Dataset:} In this section, we evaluate the effectiveness of our proposed INF³ method on the CAVE dataset (scaling factor of 4) and compare it with existing MHIF methods. As shown in the left part of Tab.~\ref{table_reduced}, our INF³ outperforms other state-of-the-art deep learning models by a large margin. For instance, our INF³ improves PSNR by 1.31 dB, 2.40 dB, 0.75 dB, and 2.00 dB compared with DHIF~\cite{huang2022deep}, Fusformer~\cite{hu2022fusformer}, MoG-DCN~\cite{dong2021model}, and HSRNet~\cite{hu2021hyperspectral}, respectively. The proposed INF³ achieves significant improvements in two QIs, i.e., SAM and ERGAS. In particular, our INF³ improves ERGAS by $11.71\%$ and $18.33\%$, compared with the second and third best models. In addition, our INF³ outperforms MoG-DCN~\cite{dong2021model} and DHIF~\cite{huang2022deep} on SAM and has only two-fifths and one-seventh of their parameters. Moreover, to aid in visual verification, we provide pseudo-color depictions of the fused products and some error maps in Fig.\ref{cave_comparison}. It can be observed that the generated results of our INF³ are very close to the ground truth and maintain better reconstruction quality with more accurate textures. Regarding the absolute error maps in Fig.\ref{cave_comparison}, the closer the reconstruction impact is to the original picture, the more blue the error map's color is. It is evident that INF³ restores texture details better than the other techniques under comparison, which is consistent with the analysis in Tab.~\ref{table_reduced}.

\noindent{\bf Results on Harvard Dataset:} Fig.\ref{fig_HARVARD} displays 10 test images from the Harvard dataset. Moreover, the right-hand portion of Tab.\ref{table_reduced} presents the comparison results of five indices obtained by all compared methods on another hyperspectral image dataset, namely Harvard, for a scaling factor of 4. It is evident that the average PSNR value of our proposed INF³ is higher by 0.17 dB and 0.51 dB compared to the second-best and third-best methods, respectively. Although our model is slightly inferior to the second-best MoG-DCN~\cite{dong2021model} in terms of SAM, our model's parameters are only two-fifths of MoG-DCN's. Moreover, our model achieves the best results on ERGAS and SSIM, indicating the best structural recovery. Furthermore, Fig.~\ref{harvard_comparison} illustrates that our proposed INF³ is capable of reconstructing the detailed structure of the original image. Notably, our method restores the finest details of the bike, the metallic sheen, and the texture of the backpack. These error maps also demonstrate that our proposed INF³ achieves the best fidelity in terms of texture details. Additionally, the fact that our residuals are closer to blue indicates that our recovery is better than other methods. 

\subsection{Ablation Study}
\label{section4.1}
In this section, we profoundly discuss the effectiveness of dual high-frequency fusion (DHIF), which combines LR and HR domain in the INF³. Our primary concern is whether injecting relative location information can aid the network in image recovery. Therefore, we conducted an ablation study to assess this. Furthermore, we included the proposed weight generation method in the ablation study. To maintain brevity and generality, the analysis is conducted on the CAVE dataset.

\begin{table}[!ht] 
	\setlength\tabcolsep{2pt} 
	\tiny
	\caption{The average four QIs and the corresponding parameters on the CAVE dataset simulating a scaling factor of 4. LR and HR mean low-resolution and high-resolution domain high-frequency information injection, respectively.}\label{table_lh}
	\resizebox{\linewidth}{!}{
		\begin{tabular}{cc|ccccc}
			\hline
			LR & HR & PSNR &SAM &ERGAS &SSIM &\\ \hline
			\usym{2713} &\usym{2717} &42.55$\pm$2.58 &2.91$\pm$0.93 &2.82$\pm$1.74 &0.990$\pm$0.0020\\
			\usym{2717}&\usym{2713} &52.17$\pm$4.02 &2.01$\pm$0.61 &1.02$\pm$0.77 &0.997$\pm$0.0014 \\
			\usym{2713}&\usym{2713} &\textcolor{red}{\textbf{52.36$\pm$3.93}} &\textcolor{red}{\textbf{1.99$\pm$0.60}} &\textcolor{red}{\textbf{0.99$\pm$0.73}} &\textcolor{red}{\textbf{0.997$\pm$0.0013}} \\
			\hline
	\end{tabular}}
	\hspace{2pt}
\end{table}
\emph {1) Dual high-frequency fusion:} To evaluate the effectiveness of dual-high-frequency information injection, we conducted several experiments. As shown in Tab.~\ref{table_lh}, we found that the removal of high-frequency information injection in HR domain resulted in a significant decline in the performance of INF³. This indicates that high-resolution and high-frequency information provides more detailed information during the fusion process of INF³. Moreover, the performance of INF³ slightly decreased when LR domain high-frequency information injection was removed, suggesting that high-frequency information of LR domain plays a supportive role in the fusion process. The utilization of different resolution information resulted in the best performance for our INF³. The importance of information at various resolutions for MHIF tasks inspired us to design this structure, and the experiments supported the rationality behind this design.

\begin{table}[!ht] 
	\setlength\tabcolsep{2pt} 
	\tiny
	\caption{The average four QIs and the corresponding parameters on the CAVE dataset simulating a scaling factor of 4. $\mathbf{\delta_c}$ means the relative coordinate $C_q-C_i$.}\label{table_p}
	\resizebox{\linewidth}{!}{
		\begin{tabular}{c|ccccc}
			\hline
			$\delta_c$ & PSNR &SAM &ERGAS &SSIM &\\ \hline
			\usym{2717} &52.22$\pm$3.92 &\textcolor{red}{\textbf{1.98$\pm$0.58}} &1.00$\pm$0.74 &0.997$\pm$0.0013\\
			\usym{2713} &\textcolor{red}{\textbf{52.36$\pm$3.93}} &1.99$\pm$0.60 &\textcolor{red}{\textbf{0.99$\pm$0.73}} 
			&\textcolor{red}{\textbf{0.997$\pm$0.0013}} \\
			\hline
	\end{tabular}}
	\hspace{0.1pt}
\end{table}

\begin{figure}[h]
	\centering
	\includegraphics[width=8.5cm, height=8.1cm]{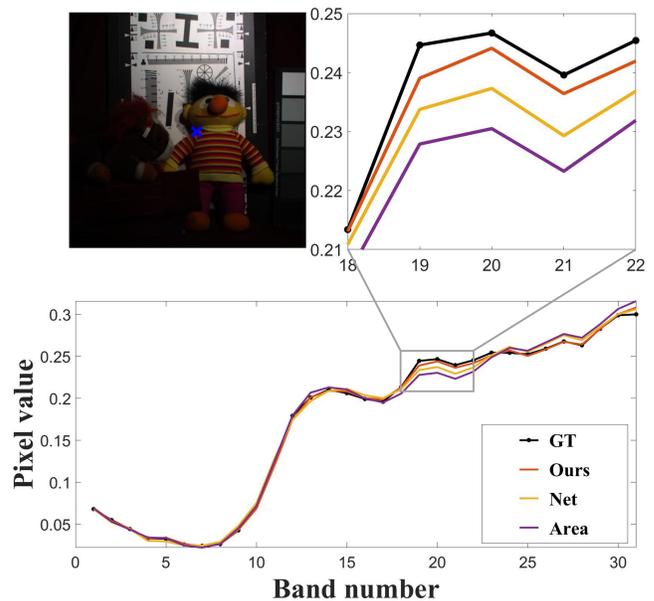}
	\caption{The blue cross on the top left image `chart and stuffed toy' shows where the spectral vector is located, while the top right picture and the bottom picture represent the spectral vector at position (276, 260). The bottom picture displays the output of the network-based and area-based weight generation methods, denoted by `Net' and `Area', respectively.}
	\label{CAVE_p1}
	\vspace{-5mm}
\end{figure}

\emph {2) Relative coordinate :} In this section, we will analyze the effectiveness of the relative coordinate $C_q-C_i$ in INF³. The relative coordinate and pixels belong to different modalities, where the former represents the distance of interpolation, and the latter represents the value of interpolated value. We are curious whether the information from different modalities can aid the MLP in understanding the fusion and interpolation processes in INF³. To address this, we conducted an ablation experiment to eliminate our confusion. Specifically, we removed the relative coordinate from INF³ while keeping the rest unchanged. Tab.~\ref{table_p} presents our results, showing that the inclusion of the relative coordinate improves the network's understanding of the MHIF task and has a positive impact on its realization. 

\begin{table}[!ht]
	\setlength\tabcolsep{2pt}
	\tiny
	\caption{The four QIs and the corresponding parameters on the image `balloons' simulating a scaling factor of 4. Net-based and Area-based represent the method of generating weights based on network and area in JIIF~\cite{tang2021joint} and LIIF~\cite{chen2021learning}, respectively. }\label{table_wg}
	\resizebox{\linewidth}{!}{
		\begin{tabular}{c|ccccc}
			\hline
			Methods \qquad&\qquad PSNR \qquad&\qquad SAM\qquad &\qquad ERGAS \qquad&\qquad SSIM \qquad&\\ 
			\hline	
			Area-based \qquad&\qquad 54.741 \qquad&\qquad 1.294 \qquad&\qquad 0.335 \qquad&\qquad 0.9978\qquad \\
			Net-based \qquad &\qquad 54.392 \qquad&\qquad 1.281 \qquad&\qquad 0.350 \qquad&\qquad 0.9977 \qquad\\
			Ours \qquad&\qquad \textcolor{red}{\textbf{54.813}} \qquad&\qquad \textcolor{red}{\textbf{1.283}}\qquad &\qquad\textcolor{red}{\textbf{0.331}} \qquad
			&\qquad \textcolor{red}{\textbf{0.9978}} \qquad\\	
			\hline
	\end{tabular}}
	\hspace{0.1pt}
	\vspace{-3mm}
\end{table}

\emph {3) Weight generation method:} To assess the superiority of our cosine similarity method, we conducted a comparison with area-based and network-based weight generation methods on the CAVE dataset, with INF³ serving as the backbone. As shown in Tab.~\ref{table_wg}, our approach significantly outperforms the other methods on certain images, such as `chart and stuffed toy'. To further illustrate the possible spectral distortions in the fused products, we visualized the spectral vectors. Fig.\ref{CAVE_p1} shows the spectral vectors for the 31 bands at position (276, 260) in the `chart and stuffed toy' image. For the purpose of clarity, we have zoomed in on the spectral vectors of the 18th-22th bands, as indicated by the rectangular boxes in Fig.\ref{CAVE_p1}. In both the figures, it is evident that the spectral vectors of the proposed method (the red lines) are the closest to the ground truth (GT).

\begin{table}[!ht] 
	\setlength\tabcolsep{2pt} 
	\tiny
	\caption{The average four QIs and the corresponding parameters on the CAVE dataset simulating a scaling factor of 4. M means a million. }\label{table_I}
	\resizebox{\linewidth}{!}{
		\begin{tabular}{c|cccccc}
			\hline
			Methods  &   PSNR    &   SAM     &   ERGAS   &   SSIM    &$\#$params &\\ 
			\hline	
			Bilinear &51.93$\pm$3.99 &2.05$\pm$0.62 &1.04$\pm$0.78 &0.997$\pm$0.0017 &3.003 M\\
			Bicubic &51.98$\pm$4.06 &2.04$\pm$0.61 &1.04$\pm$0.79 &0.997$\pm$0.0018 &3.003 M\\
			Pixel shuffle &52.15$\pm$4.27 &\textcolor{red}{\textbf{1.98$\pm$0.58}} &1.04$\pm$0.83 &0.997$\pm$0.0020 &7.722 M\\
			Ours &\textcolor{red}{\textbf{52.36$\pm$3.93}} &1.99$\pm$0.60 &\textcolor{red}{\textbf{0.99$\pm$0.73}} 
			&\textcolor{red}{\textbf{0.997$\pm$0.0013}} &\textcolor{red}{\textbf{2.902 M}}\\	
			\hline
	\end{tabular}}
	\hspace{2pt}
\end{table}
\emph{4) Upsampling methods:} In this section, we present experiments that compare INF³ with other upsampling methods. Intuitively, INF³ can be regarded as an interpolation algorithm. Unlike traditional interpolation algorithms, it provides each interpolated point with additional relative position information via the MLP layer, which incorporates multi-modal information. Specifically, we compared INF³ with pixel-shuffle~\cite{shi2016real} and traditional interpolation methods that are commonly used in convolutional neural networks. As shown in Tab.~\ref{table_I}, our INF³ outperforms other methods in terms of MHIF tasks with fewer parameters.

\section{Conclusion}
In this paper, we propose the Implicit Neural Feature Fusion Function (INF³) and design an Implicit Neural Fusion Network (INFN) based on it for multispectral and hyperspectral image fusion task. Unlike previous CNN-based approaches, we novelly fuse multimodal information including coordinate, spatial and spectral data for multiple times, and accordingly modify the previous Implicit Neural Representation of upsampling interpolation to make better use of high-frequency information. By training two different branches of the encoder, the input information is fused in two stages and entered within the INR framework, whose effectiveness in utilizing high-frequency information has also been verified. The INF³-based process also provides a generalized paradigm for other multimodal fusion tasks. Experimental results demonstrate that our method can achieve state-of-the-art performance on two different datasets. Moving forward, we will persist in exploring dependable network-based interpolation fusion methods and stable weight generation techniques.


\bibliographystyle{ACM-Reference-Format}
\bibliography{ref}

\end{document}